\documentclass[wcp]{jmlr}

\usepackage{longtable}

\usepackage{booktabs}

\usepackage{amsfonts}
\usepackage{epstopdf,epsf,psfrag}
\usepackage{bbm}

\usepackage{tikz}
\usetikzlibrary{arrows}
\tikzset{
  treenode/.style = {align=center, inner sep=0pt, text centered,
    font=\sffamily},
  arn_n/.style = {treenode, circle, black, font=\sffamily\bfseries, draw=black,
    fill=white, text width=2em},
  arn_r/.style = {treenode, circle, red, draw=red, 
    text width=2em, very thick},
}

\usepackage[utf8]{inputenc}
\usepackage{colortbl}

\usepackage{hyperref}
\usepackage{framed}

\newcommand{\train}{\mathcal{S}}

\newcommand{\dict}{\mathcal{D}}

\jmlrvolume{101}
\jmlryear{2019}

\title[Functional Isolation Forest]{Functional Isolation Forest}

  \author{\Name{Guillaume Staerman} \Email{guillaume.staerman@telecom-paris.fr}\\
 \Name{Pavlo Mozharovskyi} \Email{pavlo.mozharovskyi@telecom-paris.fr}\\
 \Name{Stephan Clémençon} \Email{stephan.clemencon@telecom-paris.fr}\\
 \Name{Florence d'Alché-Buc} \Email{florence.dalche@telecom-paris.fr}\\
 \addr LTCI, Télécom Paris, Institut Polytechnique de Paris
}

\begin{document}

\maketitle

\begin{abstract}
For the purpose of monitoring the behavior of complex infrastructures (\textit{e.g.} aircrafts, transport or energy networks), high-rate sensors are deployed to capture multivariate data, generally unlabeled, in quasi continuous-time to detect quickly the occurrence of anomalies that may jeopardize the smooth operation of the system of interest. The statistical analysis of such massive data of functional nature raises many challenging methodological questions. The primary goal of this paper is to extend the popular {\sc Isolation Forest} (IF) approach to Anomaly Detection, originally dedicated to finite dimensional observations, to functional data. The major difficulty lies in the wide variety of topological structures that may equip a space of functions and the great variety of patterns that may characterize abnormal curves. We address the issue of (randomly) splitting the functional space in a flexible manner in order to isolate progressively any trajectory from the others, a key ingredient to the efficiency of the algorithm. Beyond a detailed description of the algorithm, computational complexity and stability issues are investigated at length. From the scoring function measuring the degree of abnormality of an observation provided by the proposed variant of the IF algorithm, a \textit{Functional Statistical Depth} function is defined and discussed, as well as a multivariate functional extension. Numerical experiments  provide strong empirical evidence of the accuracy of the extension proposed.
\end{abstract}


\section{Introduction}\label{sec:intro}

The digital information boom, that goes hand in hand with the recent technological advances in data collection and management (e.g. IoT, distributed platforms), offers new perspectives in many areas of human activity (e.g. transportation, energy, health, commerce, insurance), and confronts these domains with major scientific challenges for exploiting these observations. The ever growing availability of massive data, often collected in quasi-real time, engendered high expectations, in particular the need of increased automation and computational efficiency, with the goal to design more and more `intelligent' systems. In particular, modern high-rate sensors enabling the continuous observation of the behavior of complex systems  pave the way for the design of efficient unsupervised machine-learning approaches to anomaly detection, that may find applications in various domains ranging from fraud surveillance to distributed fleet monitoring through predictive maintenance or health monitoring of complex systems.  However, although many unsupervised learning procedures for anomaly detection (AD in abbreviated form) have been proposed, analyzed and applied in a variety of practical situations (see, \textit{e.g.}, \citep{Chandola}), the case of functional data, though of crucial importance in practice (refer to \citep{ramsey,ferraty} for an account of Functional Data Analysis) has received much less attention in the literature, the vast majority of methods that are documented in the literature being generally model-based. The main barrier to the design of nonparametric anomaly detection techniques tailored to the functional framework lies in the huge diversity of patterns that may carry the information that is relevant to discriminate between abnormal and normal observations, see \citep{rousseeuw}.

It seems indeed far from straightforward to extend machine-learning methods for anomaly detection in the finite-dimensional case such as \citep{ScottNowak06,SPSSW01,SHS05,VerVer06,ParkHuangDing10}, unless preliminary \textit{filtering techniques} are used. The  \textit{filtering approach} consists in projecting the functional data onto an adequate finite dimensional function subspace and using then the coefficients describing the latter to "feed" next some AD algorithm for multivariate data \citep{ramsey}. The basis functions are either selected through Principal Component Analysis (they correspond in this case to elements of the Karhunen-Loeve basis related to the process under study, supposedly of second order), or else are chosen among a dictionary of "time-frequency atoms" according to their capacity to represent efficiently the data. The representation a priori chosen, which can either enhance artificially certain accessory patterns or else make totally disappear some crucial features, critically determines performance of such an approach, the type of anomalies that can be recovered being essentially shaped by this choice.

The angle embraced in the present article is very different, the goal pursued being to extend the popular {\sc Isolation Forest} methodology \citep{LiuTZ08,LiuTZ12} to the functional setup. This \textit{ensemble learning} algorithm builds a collection of \textit{isolation trees} based on a recursive and randomized tree-structured partitioning procedure. An isolation tree is a binary tree, representing a nested collection of partitions of the finite dimensional feature space, grown iteratively in a top-down fashion, where the cuts are axis perpendicular and random (uniformly, w.r.t. the splitting direction and the splitting value both at the same time). Incidentally, a variant referred to as {\sc Extended Isolation Forest} \citep{Hariri}, has recently been proposed in the purpose of bias reduction: rather than randomly selecting a perpendicular split, a splitting direction is randomly chosen in the unit ball. An \textit{anomaly score} is assigned to any observation, depending on the length of the path necessary to isolate it from the rest of the data points, the rationale behind this approach being that anomalies should be easier to isolate in a random manner than normal (in the sense of 'non-abnormal') data. Beyond obvious advantages regarding computational cost, scalability (\textit{e.g.} isolation trees can be built from subsamples) and interpretability, the great flexibility offered by Isolation Forest regarding the splitting procedure called recursively makes it appealing when it comes to isolate (multivariate) functions/curves, possibly exhibiting a wide variety of geometrical shapes. It is precisely the goal of this paper to introduce a new generic algorithm, \textsc{Functional Isolation Forest} (FIF) that generalizes (Extended) Isolation Forest to the infinite dimensional context. Avoiding dimensionality reduction steps, this extension is shown to preserve the assets of the original algorithm concerning computational cost and interpretability. Its efficiency is supported by strong empirical evidence through a variety of numerical results.
 
The paper is organized as follows. Section \ref{sec:back} recalls the principles under the Isolation Forest algorithm for AD in the multivariate case and introduces the framework we consider for AD based on functional data. In Section \ref{sec:fun}, the extension to the functional case is presented and its properties are discussed at length. In Section \ref{sec:NR}, we study the behavior of the new algorithm and compare its performance to alternative methods standing as natural competitors in the functional setup through experiments. In Section \ref{sec:mulext}, extension to multivariate functional data is considered, as well as relation to the data depth function and an application to the supervised classification setting. Eventually, several concluding remarks are collected in Section \ref{sec:conc}.

\section{Background and Preliminaries}\label{sec:back}

Here we briefly recall the Isolation Forest algorithm and its advantages (Section~\ref{subsec:IF}) and next introduce the framework for functional anomaly detection we consider throughout the paper (Section~\ref{subsec:fda}).

\subsection{Isolation Forest }\label{subsec:IF}

As a first go, we describe the Isolation Forest algorithm for AD in the multivariate context in a formalized manner for clarity's sake, as well as the Extended Isolation Forest version, see \citep{LiuTZ08,LiuTZ12} and \citep{Hariri} respectively.  These two unsupervised algorithms can be viewed as \textit{Ensemble Learning} methods insofar as they build a collection of binary trees and an anomaly scoring function based on the aggregation of the latter.  Let $\train_n=\{x_1,\; \ldots,\; x_n \}$ be a training sample composed of  $n$ independent realizations of a generic random variable, $X$, that takes its value in a finite dimensional Euclidian space,
$\mathbb{R}^d$ say, $X=(X^{(1)},\; \ldots,\; X^{(d)})$. 

An \textit{isolation tree} (\textit{i}tree in abbreviated form) $\mathcal{T}$ of depth $J\geq 1$ is a proper binary tree that represents a nested sequence of partitions of the feature space $\mathbb{R}^d$. The root node corresponds to the whole space $\mathcal{C}_{0,0}=\mathbb{R}^d$, while any node of the tree, indexed by the pair $(j,k)$ where $j$ denotes the depth of the node with $0\leq j< J$ and $k$, the node index with $0\leq k\leq 2^{j}-1$, is associated to a subset $\mathcal{C}_{j,k}\subset\mathbb{R}^d$. A non terminal node $(j,k)$ has two children, corresponding to disjoint subsets $\mathcal{C}_{j+1,2k}$ and $\mathcal{C}_{j+1,2k+1}$ such that $\mathcal{C}_{j,k}=\mathcal{C}_{j+1,2k}\cup \mathcal{C}_{j+1,2k+1}$. A node $(j,k)$ is said to be terminal if it has no children.

Each {\it i}tree is obtained by recursively filtering a subsample of  training data of size $\psi$ in a top-down fashion, by means of the following procedure. The dataset composed of the training observations present at a node $(j,k)$ is denoted by $\train_{j,k}$. At iteration $k+2^j$ of the \textit{i}tree growing stage, a direction $m$ in $\{1,\; \ldots,\; d\}$, or equivalently a \textit{split variable} $X^{(m)}$, is selected uniformly at random (and independently from the previous draws) as well as a \textit{split value} $\kappa$ in the interval $[\min_{x\in \train_{j,k}}x^{(m)},\; \max_{x\in \train_{j,k}}x^{(m)}]$ corresponding to the range of the projections of the points in $\train_{j,k}$ onto the $m$-th axis. The children subsets are then defined by $\mathcal{C}_{j+1,2k}=\mathcal{C}_{j,k} \cap \{x\in \mathbb{R}^d:\; x^{(m)}\leq \kappa  \}$ and $\mathcal{C}_{j+1,2k+1}=\mathcal{C}_{j,k} \cap \{x\in \mathbb{R}^d:\; x^{(m)}> \kappa  \}$, the children training datasets being defined as
$\train_{j+1,2k}=\train_{j,k} \cap \mathcal{C}_{j+1,2k}$ and $\train_{j+1,2k+1}=\train_{j,k} \cap \mathcal{C}_{j+1,2k+1}$.

An \textit{i}tree $\mathcal{T}$ is thus built by iterating this procedure until all training data points are isolated (or the depth limit $J$ set by the user is attained). A preliminary subsampling stage can be performed in order to avoid swamping and masking effects, when the size of the dataset is too large.
 When it isolates any training data point, the \textit{i}tree contains exactly $\psi-1$ internal nodes and $\psi$ terminal nodes.  An \textit{i}tree  constructed accordingly to a training subsample allows to assign to each training datapoint $x_i$ a path length $h_{\mathcal{T}}(x_i)$, namely the depth at which it is isolated from the others, \textit{i.e.} the number of edges $x_i$ traverses from the root node to the terminal node that contains the sole training data $x_i$. More generally, it can be used to define an anomaly score for any point $x \in \mathbb{R}^d$.
 \medskip
 
 \noindent{\bf Anomaly Score prediction.}
 As the terminal nodes of the \textit{i}tree $\mathcal{T}$ form a partition of the feature space, one may then define the piecewise constant function $h_{\tau}:\mathbb{R}^d\to \mathbb{N}$ by: $\forall x\in \mathbb{R}^d$,
$$
h_{\tau}(x)= j \text{ if and only if } x\in \mathcal{C}_{j,k} \text{ and } (j,k)~\text{is a  terminal node}.
$$ 
This random path length is viewed as an indication for its degree of abnormality in a natural manner: ideally, the more abnormal the point $x$, the higher the probability that the quantity $h_{\tau}(x)$ is small. Hence, the algorithm above can be repeated $N\geq 1$ times in order to produce a collection of \textit{i}trees $\mathcal{T}_1,\; \ldots,\; \mathcal{T}_N$, referred to as an \textit{i}forest, that defines the scoring function
\begin{equation}\label{equ:cscore}
s_n(x)= 2^{- \frac{1}{N c(\psi)}\sum_{l=1}^{N} h_{\tau_l}(x)},
\end{equation}
where $c(\psi)$ is the average path length of unsuccessful searches in a binary search tree, see \citep{LiuTZ08} for further details.
\medskip

\noindent{\bf  Extended Isolation Forest.} Observing that the geometry of the abnormal regions of the feature space is not necessarily well-described by perpendicular splits (\textit{i.e.} by unions of hypercubes of the cartesian product $\mathbb{R}^d$), a more flexible variant of the procedure recalled above has been proposed in \citep{Hariri}, in the purpose of bias reduction. Rather than selecting a direction in $\{1,\; \ldots,\; d \}$, one may choose a direction $u \in \mathbb{S}_{d-1}$, denoting by $ \mathbb{S}_{d-1}$ the unit sphere of the euclidian space $\mathbb{R}^d$.  A node is then cut by choosing randomly and uniformly a threshold value in the range of the projections onto this direction of the training data points lying in the corresponding region. In the case where $X$'s ditribution has a density $f(x)$ w.r.t. a $\sigma$-finite measure $\lambda$ of reference, the goal of anomaly detection can be formulated as the recovery of sublevel sets $\{x\in \mathbb{R}^d:\;\; f(x)\geq q\}$, $q\geq 0$, (under mild assumptions, they are \textit{minimum volume sets} or \textit{quantile regions}, see \citep{Polonik97,ScottNowak06}, when measuring the volume by $\lambda$), which may be not accurately approximated by unions of hyperrectangles (in the Gaussian situation for instance, such regions are the complementary sets of ellipso\"ids, $\lambda$ being Lebesgue measure on $\mathbb{R}^d$).

\subsection{Functional Data Analysis and Anomaly Detection}\label{subsec:fda}

A functional random variable $X$ is a r.v. that takes its values in a space of functions, see, \textit{e.g.},  \citep{ferraty}. To be more specific, let $I\subset \mathbb{R}_+$ be a time interval and consider a r.v. taking its values in the Hilbert space $L_{2}(I)$ of real valued and square integrable (w.r.t. Lebesgue measure) functions $x:I\rightarrow \mathbb{R}$:
\begin{equation*}\begin{array}{lll}
X& : &\Omega \longrightarrow L_{2}(I) \\
&& \omega \longmapsto X(\omega)=(X_t(\omega))_{t\in I}
\end{array}.
\end{equation*}
Without any loss of generality, we restrict ourselves with functions defined on $[0,1]$ throughout the paper. In practice,  only a finite dimensional marginal $(X_{t_1},\; \ldots,\; X_{t_p})$, $t_1<\ldots<t_p$, $p\geq 1$ and $(t_1,\; \ldots,\; t_p)\in [0,1]^p$ can be observed. However, considering $(X_{t_1},\; \ldots,\; X_{t_p})$ as a discretized curve rather than a simple random vector of dimension $p$ permits to take into account the dependence structure between the measurements over time, especially when the time points $t_i$ are not equispaced. To come back to a function from discrete values, interpolation procedures or approximation schemes based on appropriate dictionaries can be used, combined with a preliminary smoothing step when the observations are noisy. From a statistical perspective, the analysis is based on a functional dataset $\mathcal{S}_n = \{x_1,\; \ldots,\;x_n\}$ composed of $n\geq1$ independent realizations of finite-dimensional marginals of the stochastic process $X$, that may be very heterogeneous in the sense that these marginals may correspond to different time points and be of different dimensionality. One may refer to \cite{ramsey}'s book for a deep view on Functional Data Analysis (FDA in short). For simplicity, the functional data considered throughout the paper correspond to the observations of independent realizations of $X$ at the same points.

In this particular context, {\it functional} anomaly detection aims at detecting the curves that significantly differ from the others among the dataset available. Given the richness of spaces of functions, the major difficulty lies in the huge diversity in the nature of the observed differences, which may not only depend on the locations of the curves. Following in the footsteps of \cite{rousseeuw}, one may distinguish between three types of anomalies: \textit{shift} (the observed curve has the same shape as the majority of the sample except that it is shifted away), \textit{amplitude} or \textit{shape} anomalies. All these three types of anomalies can be \textit{isolated/transient} or \textit{persistent}, depending on their duration with respect to that of the observations. One may easily admit that certain types of anomalies are harder to detect than others: for instance, an isolated anomaly in shape compared to an isolated anomaly in amplitude (\textit{i.e.} change point).
Although FDA has been the subject of much attention in recent years, very few generic and flexible methods tailored to functional anomaly detection are documented in the machine-learning literature to the best of our knowledge, except for specific types of anomalies (\textit{e.g.} change-points).

In Statistics, although its applications are by no means restricted to AD, the concept of \textit{functional depth} that allows to define a notion of centrality in the path space and a center-outward ordering of the curves of the functional dataset, see, \textit{e.g.}, \citep{cuevas,claeskens,rousseeuw}, has been used for this purpose. However, since the vast majority of functional depth functions introduced only describe the relative location properties of the sample curves, they generally fail to detect other types of anomalies.
Another popular approach, usually referred to as \textit{filtering}, consists in bringing the AD problem to the multivariate case by means of an adequate projection using \textit{Functional Principal Component Analysis} (FPCA) \citep{ramsey} or a preliminary selected basis of the function space considered (\textit{e.g.} Fourier, wavelets) and apply next an AD algorithm designed for the finite-dimensional setup to the resulting representation. Such methods have obvious drawbacks. In FPCA, estimation of the Kahrunen-Loève basis can be very challenging and lead to loose approximations, jeopardizing next the AD stage, while the \textit{a priori} representation offered by the 'atoms' of a predefined basis or frame may unsuccessfully capture the patterns carrying the relevant information to distinguish abnormal curves from the others. Another approach is based on the notion of \textit{Minimum Volume sets} (MV-sets in shortened version), originally introduced in \citep{EinmahlMason92} and that generalizes the concept of quantile for multivariate distributions and offers a nice nonparametric framework for anomaly detection in finite dimension, see \cite{ScottNowak06}'s work. Given the fact that no analogue of Lebesgue measure on an infinite-dimensional Banach space exists and since, considering a law $\lambda$ of reference (\textit{e.g.} the Wiener or a Poisson measure) on the function space $\mathcal{H}$ of interest, the volume $\lambda(\mathcal{C})$ of a measurable subset $\mathcal{C}\subset \mathcal{H}$ can be hardly computed in general, it is far from straightforward to extend MV-set estimation to the functional setup.

The angle embraced in this paper is quite different. The direct approach we promote here is free from any preliminary representation stage and can be straightforwardly applied to a functional dataset. Precisely, in the subsequent section, we propose to extend the IF algorithm  to the functional data framework, in a very flexible way, so as to deal with a wide variety of anomaly shapes.

\section{Functional Isolation Forest}\label{sec:fun}

We consider the problem of learning a score function $s : \mathcal{H} \rightarrow \mathbb{R}$ that reflects the degree of anomaly of elements in an infinite dimensional space $\mathcal{H}$ w.r.t. $P$. By $\mathcal{H}$, we denote a functional Hilbert space equipped with a scalar product $ \langle .,.\rangle_{\mathcal{H}}$ such that any $\mathbf{x}\in \mathcal{H}$ is a real function defined on $[0,1]$. In the following, we describe in detail the proposed \textsc{Functional Isolation Forest} (FIF) algorithm and discuss its properties.

\subsection{The FIF algorithm} \label{subsec:FIF}

\begin{figure}[ht]
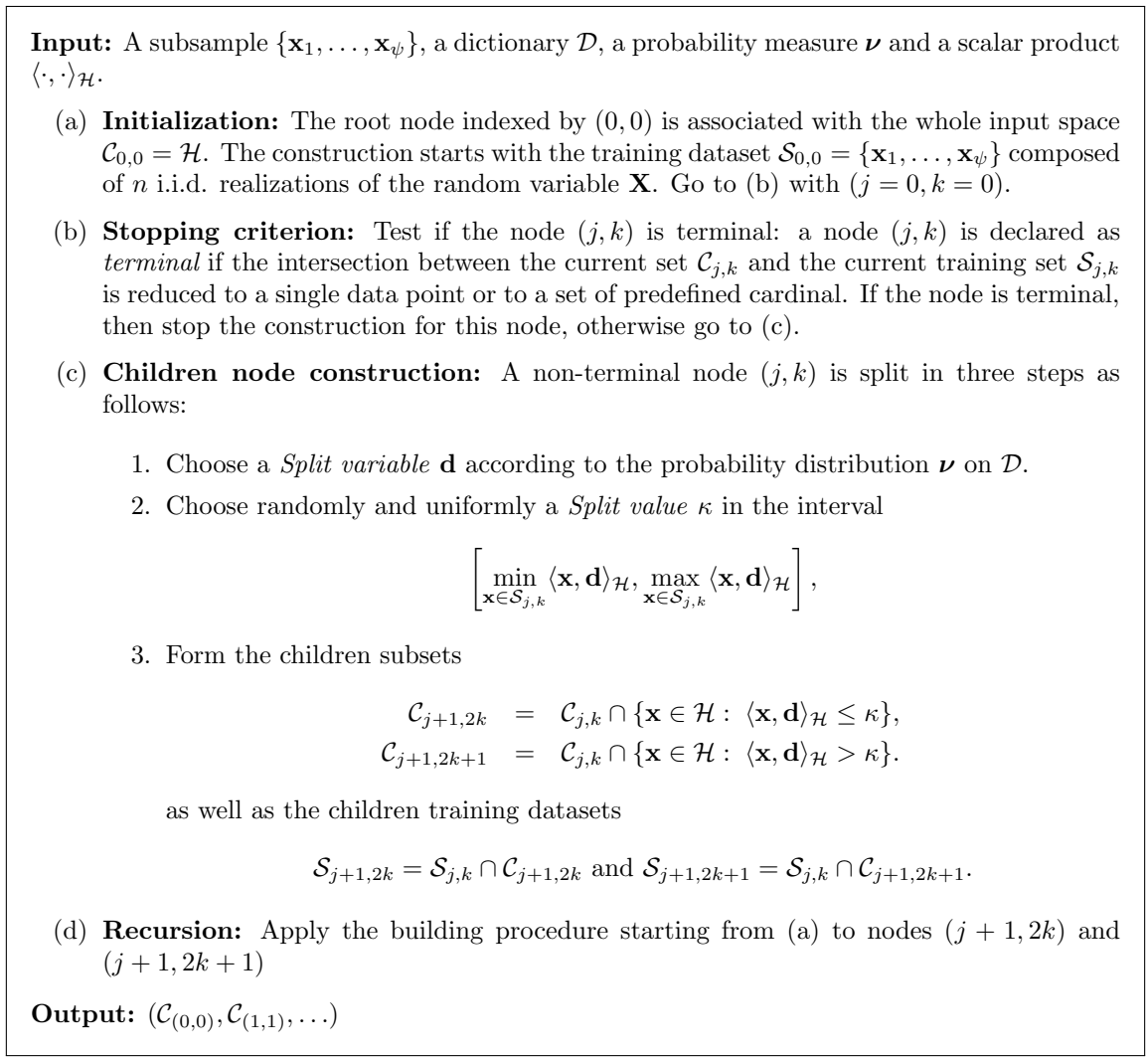

\begin{framed}{\small
{\bf Input:}  A subsample $\{\mathbf{x}_1, \ldots, \mathbf{x}_{\psi} \}$, a dictionary $\mathcal{D}$, a probability measure $\boldsymbol{\nu}$ and a scalar product $\langle \cdot , \cdot \rangle_{\mathcal{H}}$.
\begin{itemize}
\item [(a)] {\bf Initialization:} The root node indexed by $(0,0)$ is associated with the whole input space $\mathcal{C}_{0,0}=\mathcal{H}$. The construction starts with the training dataset $\train_{0,0}= \{\mathbf{x}_1, \ldots, \mathbf{x}_{\psi} \}$ composed of $n$ i.i.d. realizations of the random variable $\mathbf{X}$. Go to (b) with $(j=0, k=0)$.
\item [(b)] {\bf Stopping criterion:} Test if the node $(j,k)$ is terminal: a node $(j,k)$  is declared as {\it terminal} if the intersection between the current set $\mathcal{C}_{j,k}$ and  the current training set $\train_{j,k}$  is reduced to a single data point or to a set of predefined cardinal. If the node is terminal, then stop the construction for this node, otherwise go to (c).
\item [(c)] {\bf Children node construction:} A non-terminal node $(j,k)$ is split in three steps as follows:

\begin{enumerate}
\item Choose a \textit{Split variable} $\mathbf{d}$ according to the probability distribution $\boldsymbol{\nu}$ on $\dict$.

\item Choose randomly and uniformly a \textit{Split value} $\kappa$  in the interval $$\left[ \underset{\mathbf{x} \in \train_{j,k} }{\min}  \langle \mathbf{x},\mathbf{d} \rangle_{\mathcal{H}} ,\underset{ \mathbf{x} \in \train_{j,k}}{\max }  \langle \mathbf{x},\mathbf{d} \rangle_{\mathcal{H}} \right],$$

\item Form the children subsets
\begin{eqnarray*}
\mathcal{C}_{j+1,2k}&=&\mathcal{C}_{j,k} \cap \{\mathbf{x}\in \mathcal{H}:\; \langle \mathbf{x},\mathbf{d}\rangle_{\mathcal{H}}\leq \kappa  \},\\ \mathcal{C}_{j+1,2k+1}&=&\mathcal{C}_{j,k} \cap \{\mathbf{x}\in \mathcal{H}:\; \langle \mathbf{x},\mathbf{d}\rangle_{\mathcal{H}}> \kappa  \}.
\end{eqnarray*}
as well as the children training datasets
\begin{equation*}
\train_{j+1,2k}=\train_{j,k} \cap \mathcal{C}_{j+1,2k} \text{ and } \train_{j+1,2k+1}=\train_{j,k} \cap \mathcal{C}_{j+1,2k+1}.
\end{equation*}
\end{enumerate}
\item [(d)] {\bf Recursion:} Apply the building procedure starting from (a) to nodes $(j+1,2k)$ and $(j+1,2k+1)$

\end{itemize}
{\bf Output:} $(\mathcal{C}_{(0,0)}, \mathcal{C}_{(1,1)}, \ldots )$}
\end{framed}
\caption{\label{fig:functional_construction} Construction procedure of a F-{\it i}tree.}
\end{figure}

A Functional Isolation Forest is a collection of \textit{Functional Isolation Trees} (F-\textit{i}trees) built from $\train=\{\mathbf{x}_1,\; \ldots,\; \mathbf{x}_n \}$, a training sample composed of independent realizations of a functional random variable, $\mathbf{X}$, that takes its values in $\mathcal{H}$. Given a functional observation $\mathbf{x}$, the score returned by FIF is a monotone transformation of the empirical mean of the path lengths $h_{\tau_{l}}(\mathbf{x})$ computed by the F-\textit{i}trees $\mathcal{T}_{l}$, for $l=1, \ldots, N$~as defined in Eq. \ref{equ:cscore} in the multivariate case. While the general construction principle depicted in Section \ref{subsec:IF} remains the same for a F-\textit{i}tree, dealing with functional values raises the issue of finding an adequate feature space to represent various properties of a function. A function may be considered as abnormal according to various criteria of location and shape, and the features should permit to measure  such properties. Therefore four ingredients have been introduced to handle functional data in a general and flexible way: (i) a set of candidate {\it Split variables} and (ii) a scalar product both devoted to function representation, (iii) a probability distribution to sample from this set and select a single {\it Split variable}, (iv) a probability distribution to select a \textit{Split value}. The entire construction procedure of a F-\textit{i}tree is described in Figure ~\ref{fig:functional_construction}. \\

\noindent{\bf Function representation} To define the set of candidate {\it Split variables},  a direct extension of the original IF algorithm \citep{LiuTZ08} would be to randomly draw an argument value (\textit{e.g.} time), and use functional evaluations at this point to split a node, but this boils down to only rely on instantaneous observations of functional data to capture anomalies, which in practice will be usually interpolated. Drawing a direction on a unit sphere as in \citep{Hariri} is no longer possible due to the potentially excessive richness of $\mathcal{H}$. To circumvent these difficulties, we propose to project the observations on elements of a dictionary $\dict \subset\mathcal{H}$ that is chosen to be rich enough to explore different properties of data and well appropriate to be sampled in a representative manner.  More explicitly, given a function $\mathbf{d} \in \dict$, the projection of a function $\mathbf{x} \in \mathcal{H}$ on $\dict$, $\langle \mathbf{x}, \mathbf{d} \rangle_{\mathcal{H}}$ defines a feature that partially describes $\mathbf{x}$. When considering all the functions of dictionary $\dict$, one gets a set of candidate {\it Split variables} that provides a rich representation of function $\mathbf{X}$, depending on the nature of the dictionary. Dictionaries have been throughly studied in the signal processing community to achieve {\it sparse coding} of signals, see \textit{e.g.} \cite{Mallat2}. They also provide a way to incorporate {\it a priori} information about the nature of the data, a property very useful in an industrial context in which functional data often come from the observation of a well known device and thus can benefit from expert knowledge.\\

\noindent{\bf Sampling a} $\boldsymbol{Split}$ $\boldsymbol{variable}$ Once a dictionary is chosen, a probability distribution $\boldsymbol{\nu}$ on $\dict$ is defined to draw a {\it Split variable} $\mathbf{d}$. Note that the choice of the sampling distribution $\boldsymbol{\nu}$ gives an additional flexibility to orientate the algorithm towards the search for specific properties of the functions. \\
\noindent{\bf Sampling a} $\boldsymbol{Split}$ $\boldsymbol{value}$ Given a chosen {\it Split variable}  $\mathbf{d}$ and a current training dataset $\train_{j,k}$, a {\it Split value} is uniformly drawn in the real interval defined by the smallest and largest values of the projections on $\mathbf{d}$ when considering the observations present in the node.\\

\vspace*{0.3cm}

\noindent{\bf Discussion on the dictionary} The choice of a suited dictionary plays a key role in construction of the FIF anomaly score. The dictionary can consist of deterministic functions, incorporate stochastic elements, contain the observations from $\train$, or be a mixture of several mentioned options. In \textit{Computational Harmonic Analysis}, a wide variety of bases or frames, such as wavelets, ridgelets, cosine packets, brushlets and so on, have been developed in the last decades in order to represent efficiently/parsimoniously functions, signals or images exhibiting specific form of singularities (\textit{e.g.} located at isolated points, along hyperplanes) and may provide massive dictionaries.
The following ones will be used throughout the article: \textit{mexican hat wavelet dictionary} (MHW), \textit{Brownian motion dictionary} (B),  \textit{Brownian bridge dictionary} (BB), \textit{cosine dictionary} (Cos), \textit{uniform indicator dictionary} (UI), \textit{dyadic indicator dictionary} (DI), and the \textit{self-data dictionary} (Self) containing the dataset itself. See Section B and C of the Supplementary Materials for detailed definitions of these dictionaries and further discussion on them, respectively.\\

\noindent{\bf Discussion on the  scalar product} Besides the dictionary, the \textit{scalar product} defined on  $\mathcal{H}$ brings some additional flexibility to measure different type of anomaly. While $L_2$ scalar product allows for detection of \textit{location anomalies}, $L_2$ scalar product of derivatives (or slopes) would allow to detect anomalies regarding shape. This last type of anomalies can be challenging; \textit{e.g.} \cite{rousseeuw} mention that \textit{shape anomalies} are more difficult to detect, and \cite{mozha} argue that one should consider both location and slope simultaneously for distinguishing complex curves. Beyond these two, a wide diversity of scalar products can be used, involving a variety of $L_2$-scalar products related to derivatives of certain orders, like in the definition of Banach spaces such as weighted Sobolev spaces, see \cite{Sobolev_book}.
    
\subsection{Ability of FIF to  detect a variety of anomalies}\label{subsec:sfif}

\begin{figure}[h]
\begin{center}
\includegraphics[height=.1375\textheight, trim=0cm 0cm 1cm 1.15cm,clip=true]{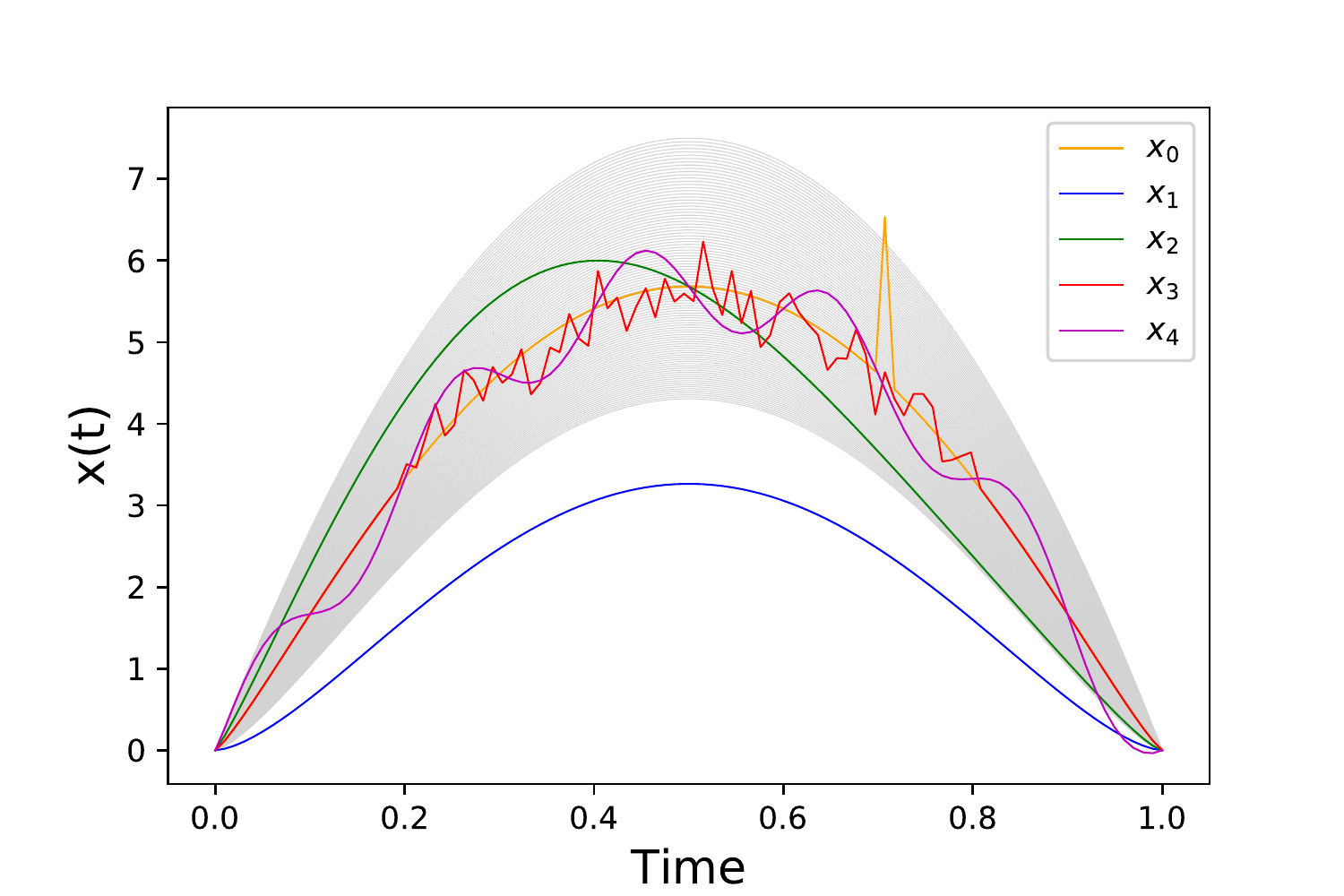}\,\includegraphics[height=.1375\textheight, trim=1.85cm 0cm 1cm 1.15cm,clip=true]{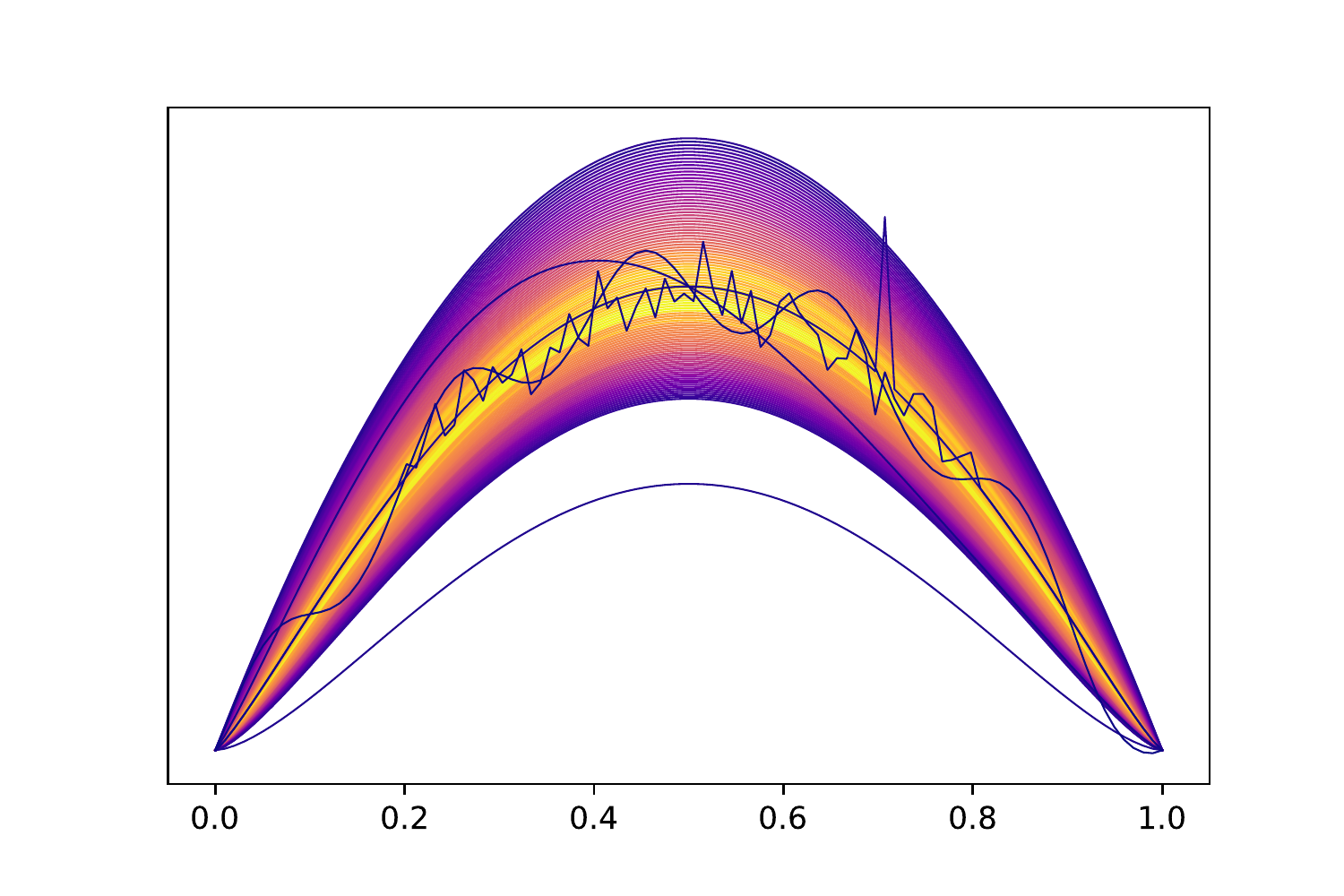}\,\includegraphics[height=.1375\textheight, trim=1.85cm 0cm 0cm 1.15cm,clip=true]{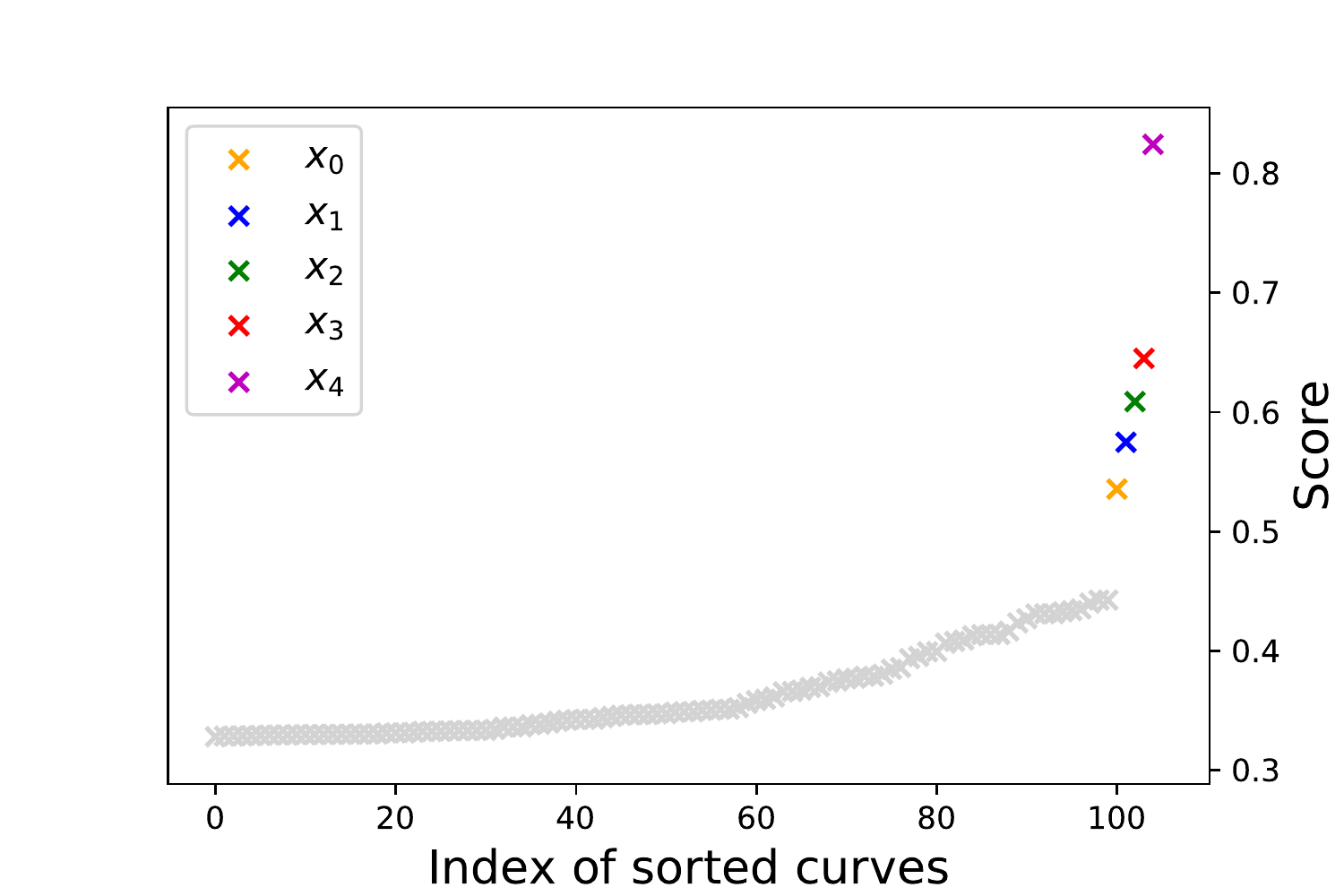}
\end{center}
\caption{The simulated dataset with the five introduced anomalies (left). The scored dataset (middle), the darker the color, the more the curves are considered anomalies. The sorted anomaly score of the dataset (right).  }
\label{fig:anomaly_example}
\end{figure}

As discussed in Section ~\ref{subsec:fda}, most of state-of-the-art methods have a focus on a certain type of anomalies and are unable to detect various deviations from the normal behavior. The flexibility of the FIF algorithm allows for choosing the scope of the detection by selecting both the scalar product and the dictionary. Nevertheless, by choosing  appropriate scalar product and dictionary, FIF is able to detect a great diversity of deviations from normal data. First, to account for both location and shape anomalies, we suggest the following scalar product that provides a compromise between the both
\[
\langle \mathbf{f},\mathbf{g} \rangle := \alpha \times \frac{\langle \mathbf{f},\mathbf{g} \rangle_{L_{2}}}{||\mathbf{f}||\,||\mathbf{g}||}+(1-\alpha) \times \frac{\langle \mathbf{f}',\mathbf{g}' \rangle_{L_{2}}}{||\mathbf{f}'||\,||\mathbf{g}'||}, \quad \alpha \in [0,1]\,,
\]
and illustrate its use right below. Thus, setting $\alpha=1$ yields the classical $L_2$ scalar product, $\alpha=0$ corresponds to the $L_{2}$ scalar product of derivative, and $\alpha=0.5$ is the Sobolev $W_{1,2}$ scalar product. To illustrate the FIF's ability to detect a wide variety of anomalies at a time, we calculate the FIF anomaly scores with the Sobolev scalar product and the \textit{gaussian wavelets dictionary}  for a sample consisting of $105$ curves defined as follows (inspired by \citep{cuevas}, see Fig.~\ref{fig:anomaly_example}):
\begin{itemize}
\item 100 curves defined by $\mathbf{x}(t) = 30 (1-t)^{q}t^{q}$ with $q$ equispaced in $[1,1.4]$,
\item 5 \textit{abnormal} curves composed by one isolated anomaly  $\mathbf{x_0}(t)= 30 (1-t)^{1.2}t^{1.2}$ with a jump in $t=0.7$,  one magnitude anomaly $\mathbf{x_1}(t)= 30 (1-t)^{1.6}t^{1.6}$ and three kind of shape anomalies $\mathbf{x_2}(t)= 30 (1-t)^{1.2}t^{1.2}+\sin (2\pi t)$,
$\mathbf{x_3}(t) = 30 (1-t)^{1.2}t^{1.2}$ noised by $\varepsilon \sim \mathcal{N}(0,0.3^{2})$ on the interval $[0.2,0.8]$ and $\mathbf{x_4}(t)= 30 (1-t)^{1.2}t^{1.2}+\frac{1}{2}\sin (10\pi t)$.
\end{itemize}

One can see that the five anomalies, although very different, are all detected by FIF with a significantly different score.

\section{Numerical Results}\label{sec:NR}

In this section, we provide an empirical study of the proposed algorithm. First, in Section~\ref{subsec:sp} we explore the stability and consistency of the score function w.r.t. the probability distribution of a r.v. $\mathbf{X}$ and the sample size. Furthermore, we examine the influence of proposed dictionaries on the score function and bring performance comparisons with benchmark methods. Second, in Section~\ref{subsec:datasets}, we benchmark the performance of FIF on several real labeled datasets by  measuring its ability to recover an "abnormal" class on the test set. In all experiments, $N$ the number of F-{\it i}trees is fixed to $100$ and the height limit is fixed to $\lceil\log_2(\psi)\rceil$.

\subsection{Impact of the Hyperparameters on Stability}\label{subsec:sp}

Since functional data are more complex than multivariate data, and the dictionary constitutes an additional source of variance, a question of stability of the FIF anomaly score estimates is of high interest. This issue is even more important because of the absence of theoretical developments due to their challenging nature.
\begin{figure}[b]
\begin{center}
\includegraphics[width=.4\textwidth, trim=0cm .6cm 1cm 1.15cm,clip=true]{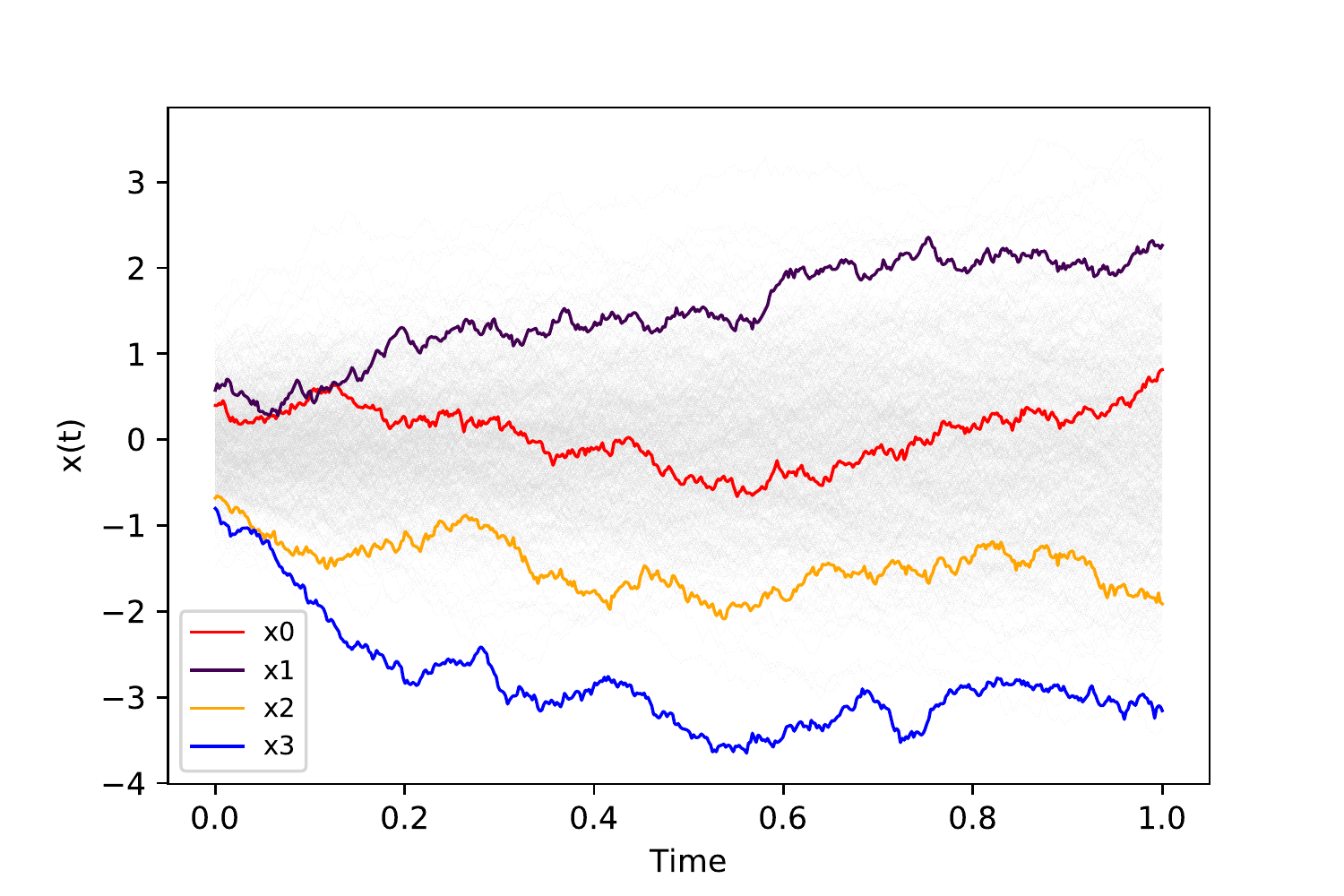}\,\includegraphics[width=.4\textwidth, trim=0cm .6cm 1cm 1.15cm,clip=true]{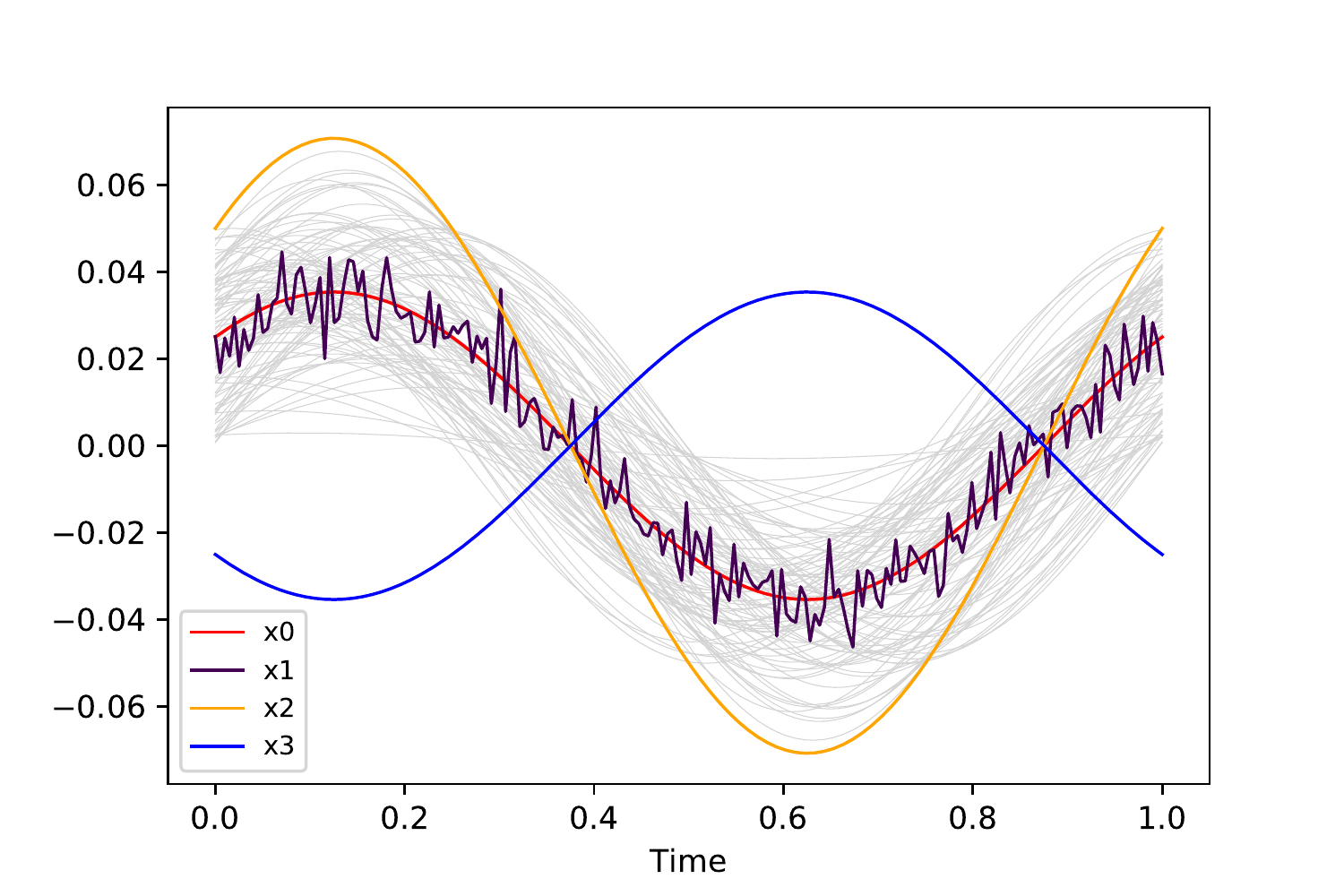}
\caption{Datasets (a) (left) and (b) (right) containing, respectively, $500$ and $200$ functional paths with $4$ selected observations.}\label{fig:varstudy}
\end{center}
\end{figure}
The empirical study is conducted on two simulated functional datasets presented in Fig.~\ref{fig:varstudy}: Dataset (a) is the standard Brownian motion being a classical stochastic process widely used in the literature. Dataset (b) has been used by \cite{claeskens} and has smooth paths. For each dataset, we choose/add four observations for which the FIF anomaly score is computed after training: a normal observation $\mathbf{x}_0$, two anomalies $\mathbf{x}_1$ and $\mathbf{x}_2$, and a more extreme anomaly $\mathbf{x}_3$. We therefore expect the following ranking of the scores: $s_n(\mathbf{x}_0) < s_n(\mathbf{x}_1) \leq s_n(\mathbf{x}_2) < s_n(\mathbf{x}_3)$, for both datasets.

\begin{figure}[!h]
\begin{center}
\begin{tabular}{cc}
\hspace*{0.5cm}$\mathbf{x}_0$ &\hspace*{0.5cm} $\mathbf{x}_1$ \\
\includegraphics[width=.4\textwidth, trim=0cm 0cm 1cm 1.2cm,clip=true]{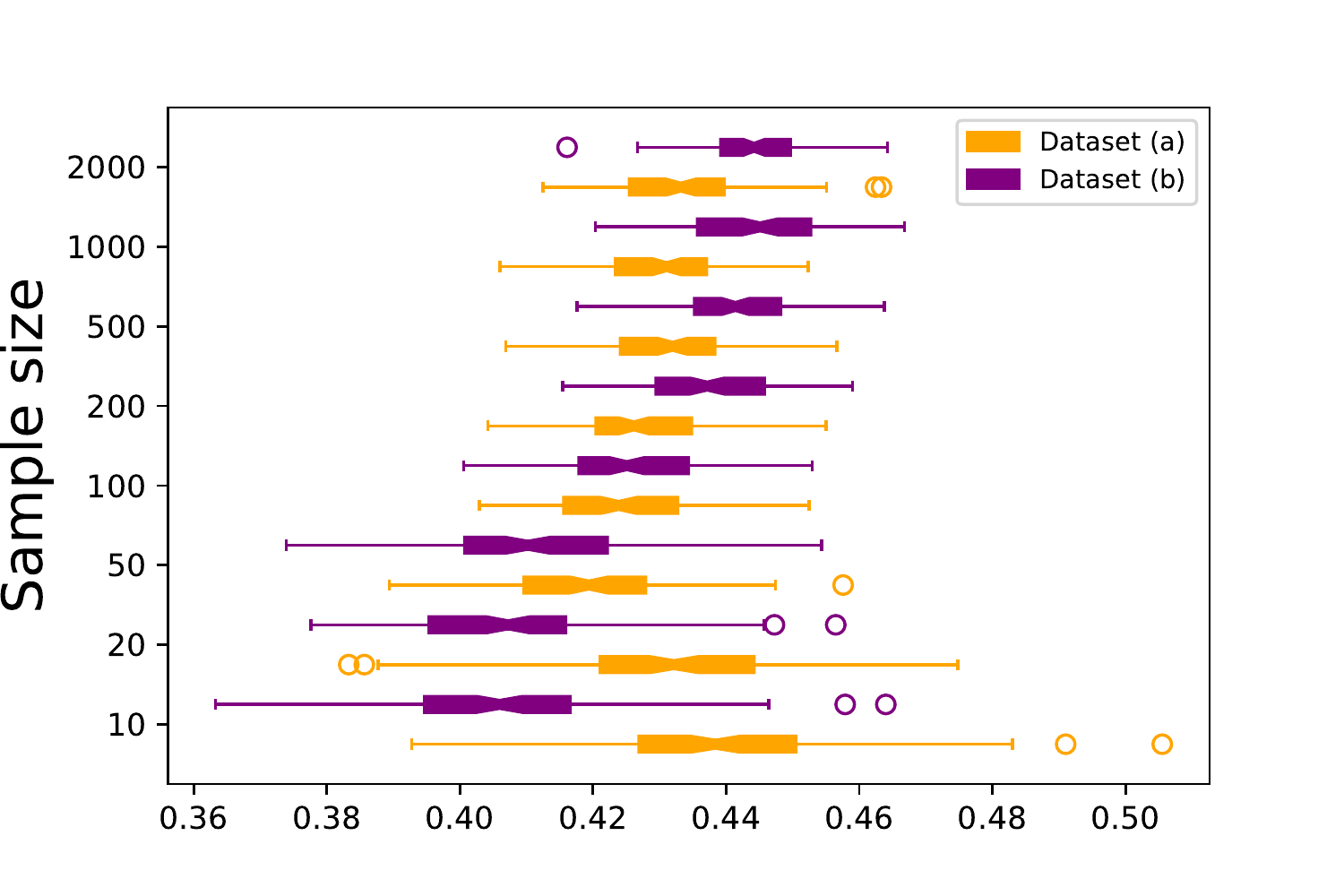} & \includegraphics[width=.4\textwidth, trim=0cm 0cm 1cm 1.2cm,clip=true]{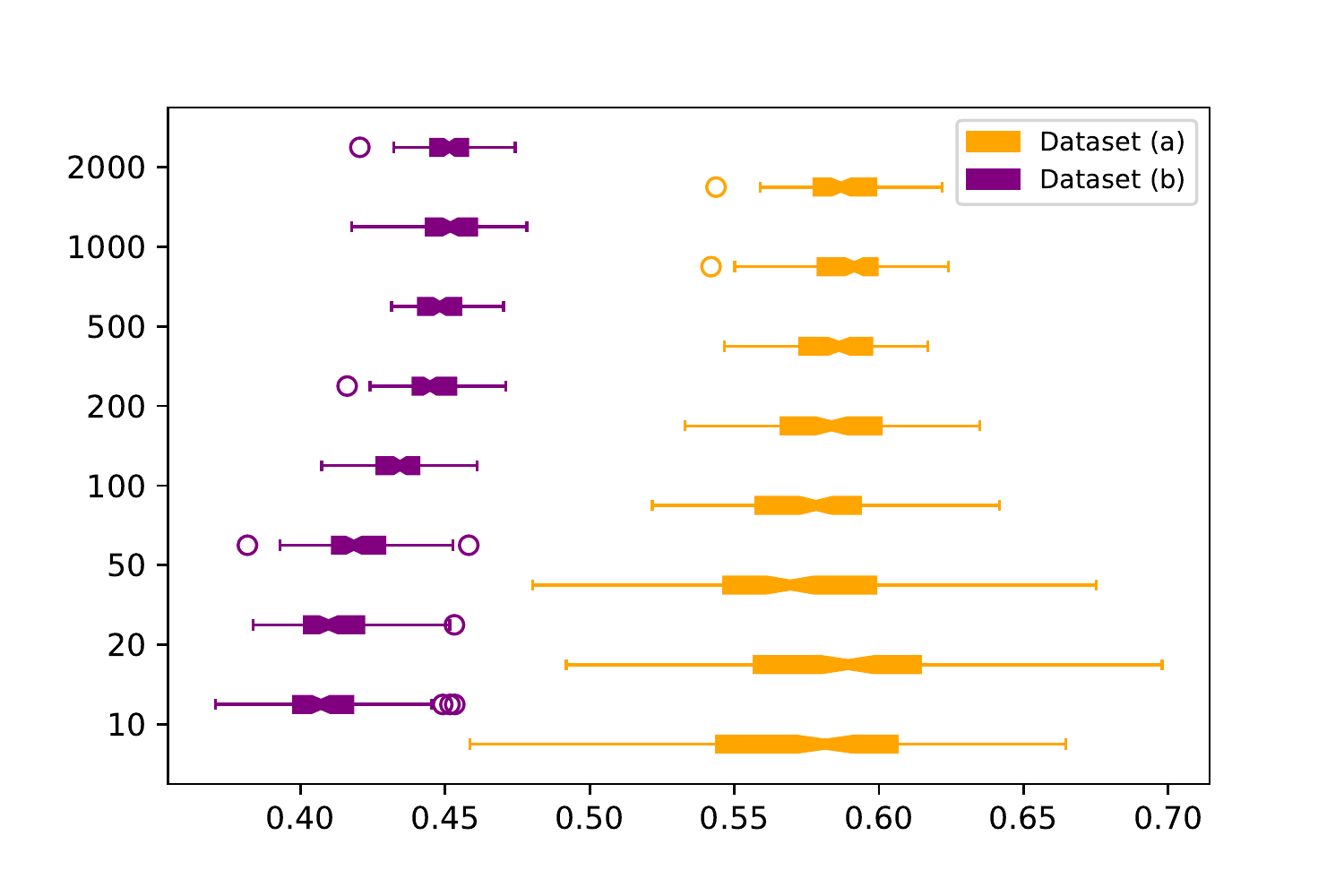} \\
\hspace*{0.5cm}$\mathbf{x}_2$ & \hspace*{0.5cm}$\mathbf{x}_3$ \\
  \includegraphics[width=.4\textwidth, trim=0cm 0cm 1cm 1.2cm,clip=true]{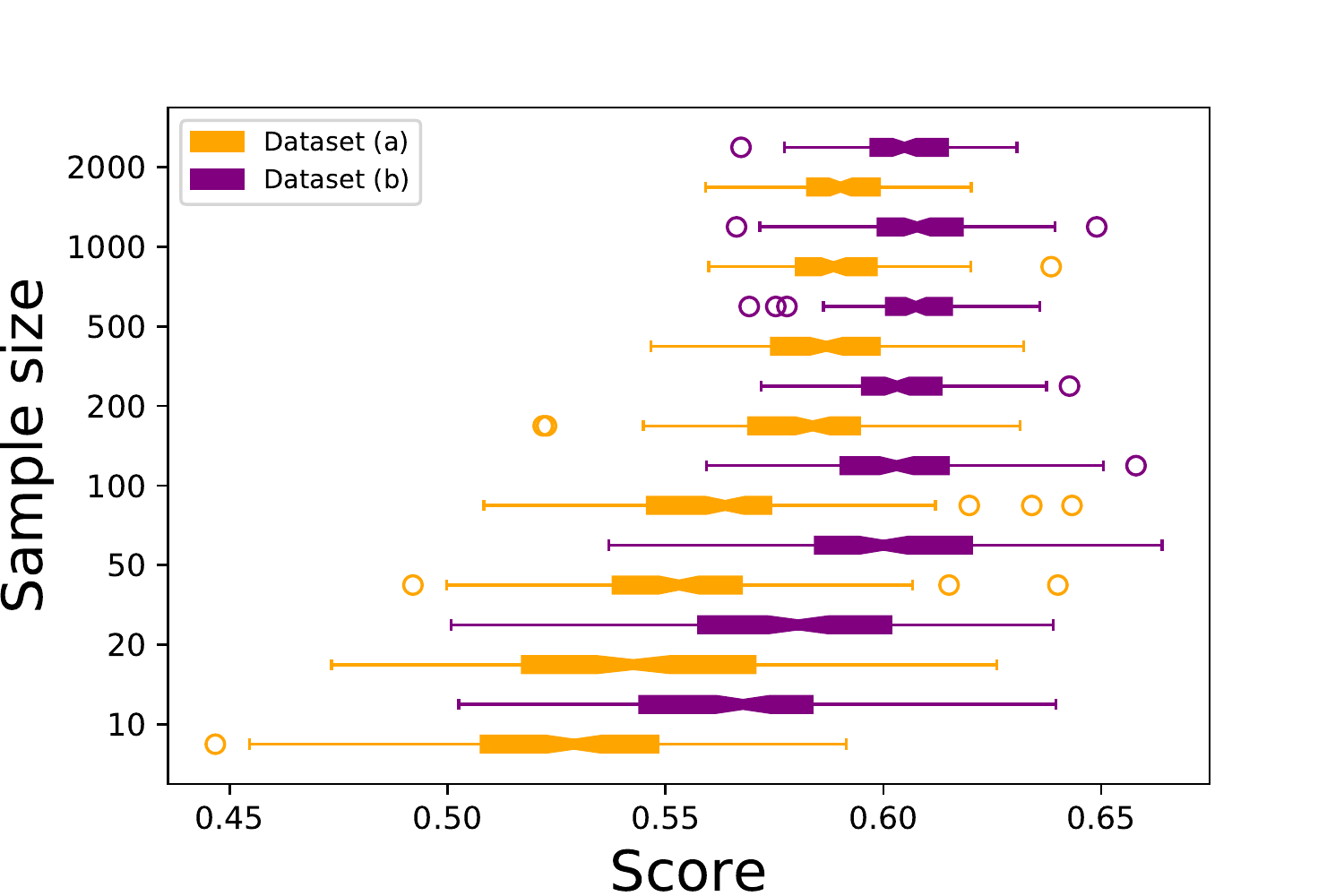}&\includegraphics[width=.4\textwidth, trim=0cm 0cm 1cm 1.2cm,clip=true]{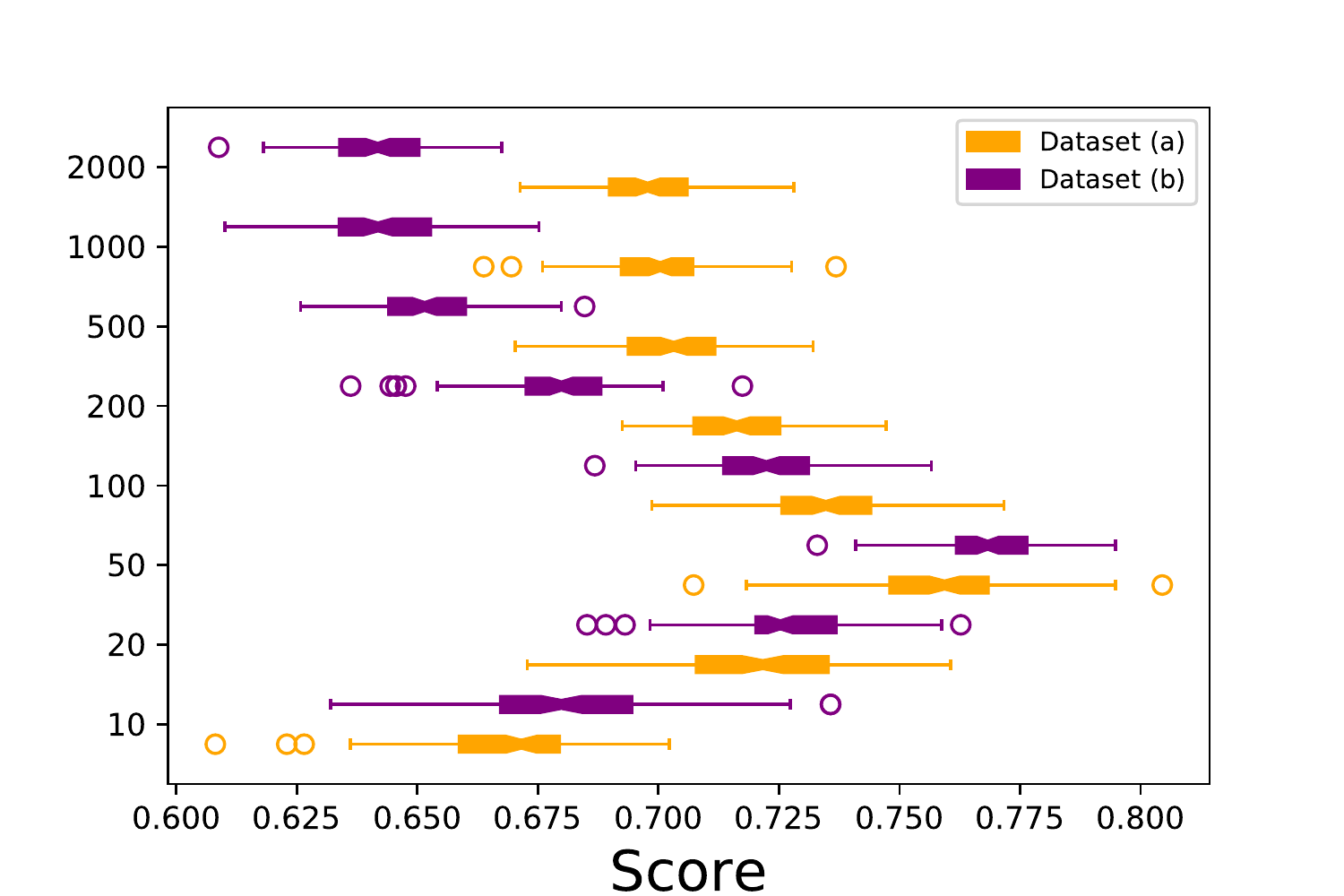}
\end{tabular}
\caption{Boxplot (over $100$ repetitions) of the FIF score for the observations $\mathbf{x}_0,\mathbf{x}_1,\mathbf{x}_2,\mathbf{x}_3$ for different sample sizes. The orange boxplots represent the dataset (a) while the purple boxplots represent the dataset (b).}
\label{fig:sizechange}
\end{center}
\end{figure}

Further, we provide an illustration of the empirical convergence of the score. All other parameters being fixed, we increase the number of observations $n$ when calculating the scores of the four selected observations; the empirical median and the boxplots of the scores computed over $100$ random draws of the dataset are shown in Fig.~\ref{fig:sizechange}.
First, one observes score convergence and variance decrease in $n$. Further, let us take a closer look at the score tendencies on the example of $\mathbf{x}_0$ and $\mathbf{x}_3$. The score of $\mathbf{x}_3$ first increases (for dataset (a)) and slightly decreases (for dataset (b)) with growing $n$ until $n$ reaches $\psi=64$, which happens because this abnormal observation is isolated quite fast (and thus has short path length) but the $c(\psi)$ in the denominator of the exponent of~\eqref{equ:cscore} increases in $\psi$. For $n>64$, the score of $\mathbf{x}_3$ decreases in $n$ since $h_i(\mathbf{x}_3)$ overestimates the real path length of $\mathbf{x}_3$ for subsamples in which it is absent; frequency of such subsamples grows in $n$ and equals, \textit{e.g.}, $0.872$ for $n=500$. On the other hand, this phenomenon allows to unmask grouped anomalies as mentioned in \citep{LiuTZ08}. The behavior is reciprocal for the typical observation $\mathbf{x}_0$. Its FIF anomaly score starts by decreasing in $n$ since $\mathbf{x}_0$ tends to belong to the deepest branches of the trees and is always selected while $\psi<n$. For larger $n$, the path length of $\mathbf{x}_0$ is underestimated for subsamples where it is absent when growing the tree, which explains slight increase in the score before it stabilizes.
\begin{figure}[!h]
\begin{center}
\begin{tabular}{cc}
\hspace*{0.5cm}$\mathbf{x}_0$ &\hspace*{0.5cm} $\mathbf{x}_1$ \\
\includegraphics[width=.4\textwidth, trim=0cm 0cm 1cm 1.2cm,clip=true]{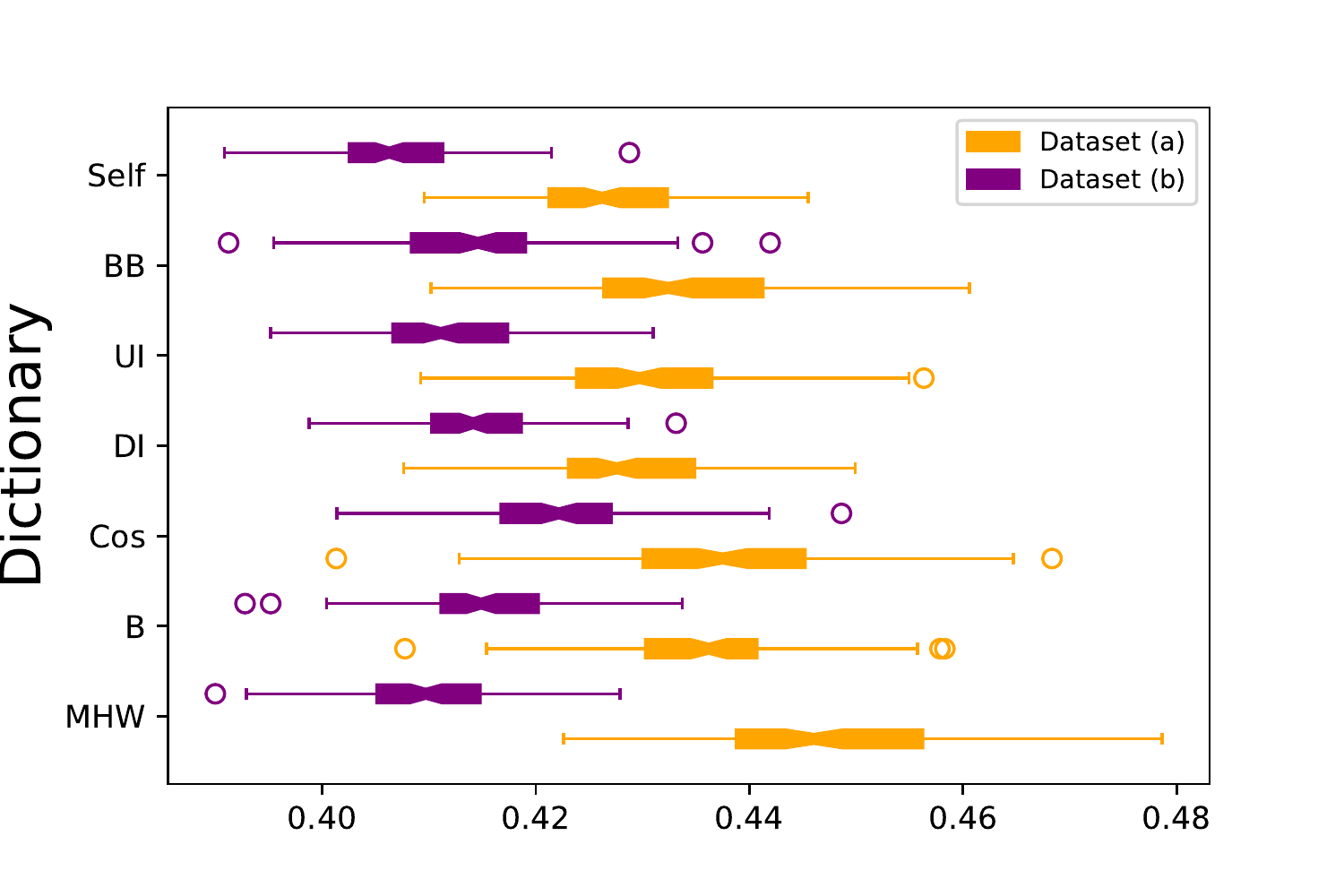} & \includegraphics[width=.4\textwidth, trim=0cm 0cm 1cm 1.2cm,clip=true]{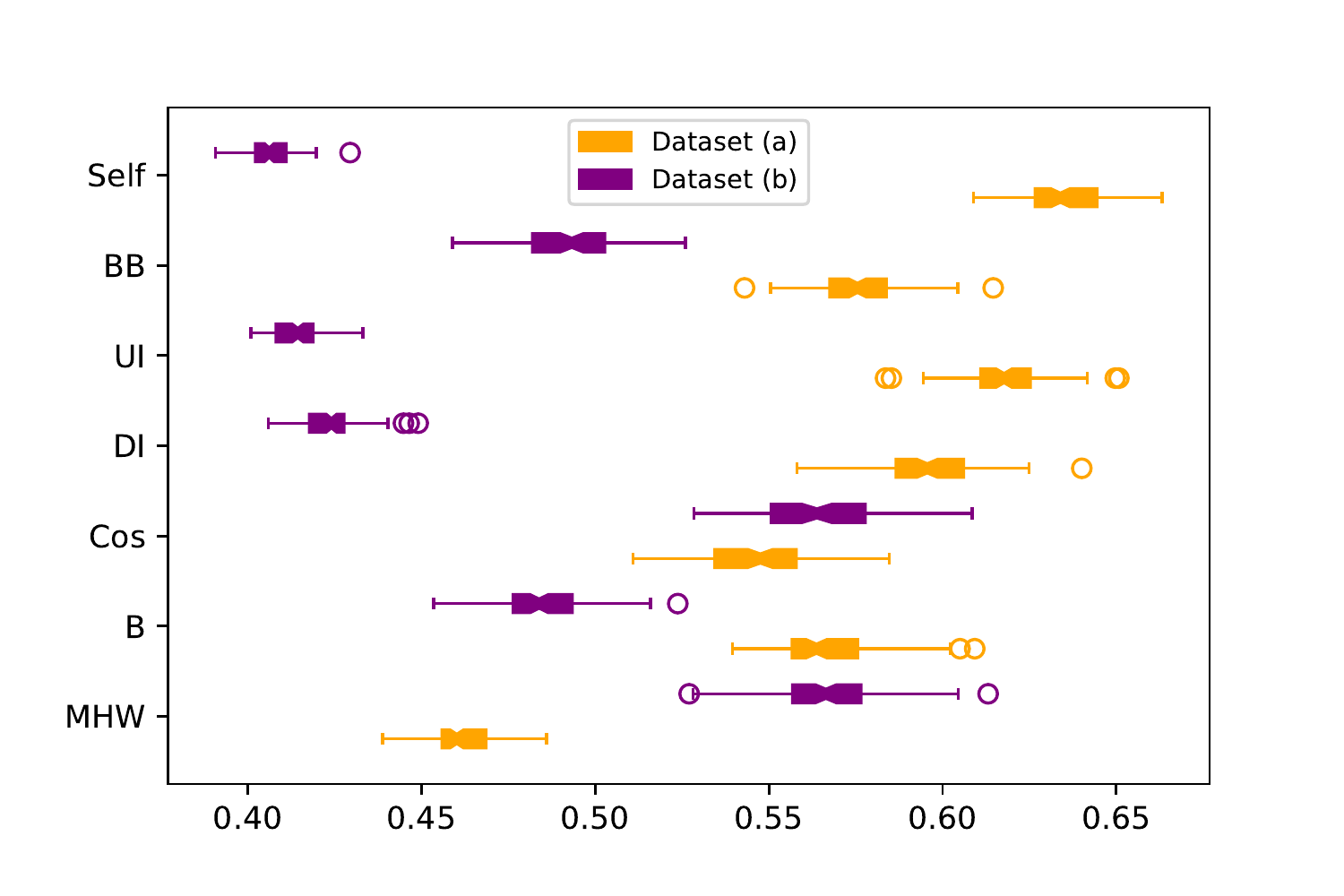} \\
\hspace*{0.5cm}$\mathbf{x}_2$ &\hspace*{0.5cm} $\mathbf{x}_3$ \\
\includegraphics[width=.4\textwidth, trim=0cm 0cm 1cm 1.2cm,clip=true]{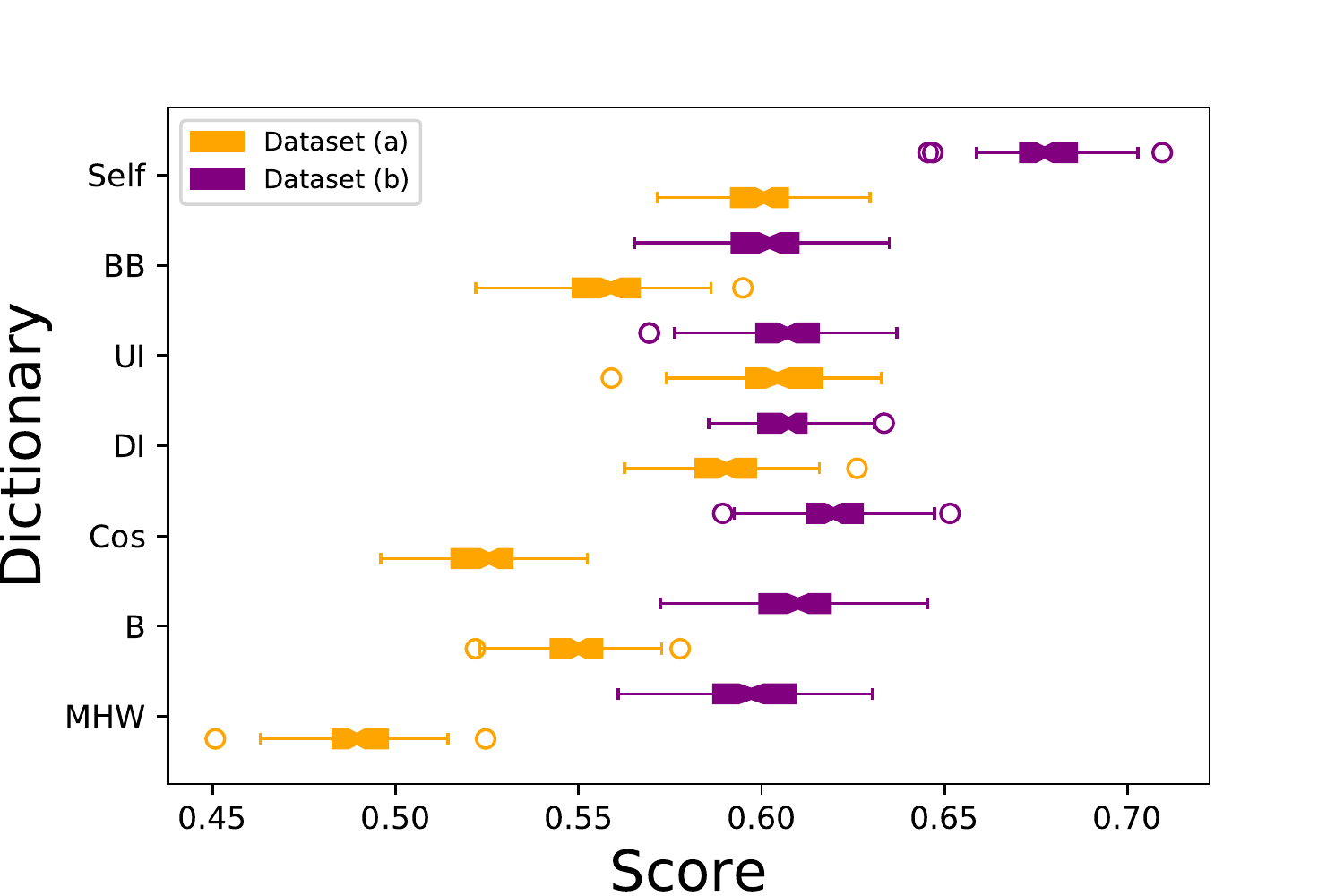} & \includegraphics[width=.4\textwidth, trim=0cm 0cm 1cm 1.2cm,clip=true]{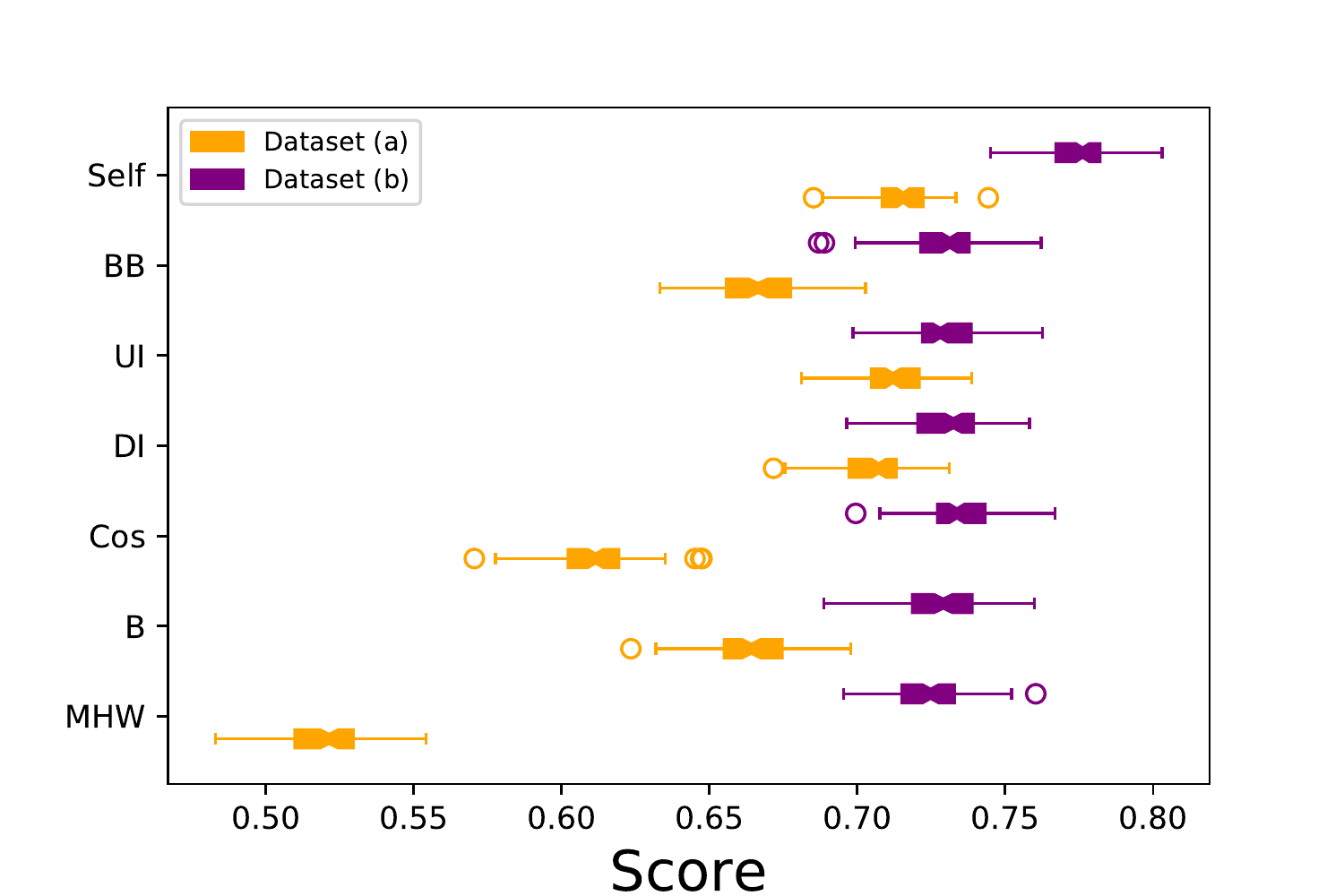}
\end{tabular}
\caption{Boxplot (over $100$ repetitions) of the FIF score for the observations $\mathbf{x}_0,\mathbf{x}_1,\mathbf{x}_2,\mathbf{x}_3$ for different dictionaries using the $L_{2}$ scalar product. The orange boxplots represent the dataset (a) while the purple boxplots represent the dataset (b).}
\label{fig:dictsdot}
\end{center}
\end{figure}
A second experiment illustrated in Fig.~\ref{fig:dictsdot} is conducted to measure the impact of various dictionaries shortly cited in Section~\ref{sec:fun} and more thoroughly described in Section~B of the Supplementary Materials; $L_2$ scalar product is used. 
One observes that the variance of the score seems to be mostly stable across dictionaries, for both datasets. Thus, random dictionaries like uniform indicator (UI) or Brownian motion (B) do not introduce additional variance into the FIF score. 
Since we know the expected ranking of the scores, we can observe that FIF relying on the Self, UI, and dyadic indicator (DI) dictionaries fail to make a strong difference between $\mathbf{x}_0$ and $\mathbf{x}_1$. Since $\mathbf{x}_1$ differs only slightly in the amplitude from the general pattern, these dictionaries seem insufficient to capture this fine dissimilarity: while Self and DI dictionaries simply do not contain enough elements, UI dictionary is to simple to capture this difference (it shares this feature with DI dictionary). For the scalar product $L_2$ on derivatives (see Fig.~18 in the Supplementary Materials), distinguishing anomalies for the Brownian motion becomes difficult since they differ mainly in location, while for a sine function the scores resemble those with the usual $L_2$ scalar product. Thus, even though---as seen in Section~\ref{subsec:sfif}---capturing different types of anomalies is one of the general strengths of the FIF algorithm, the dictionary may still have an impact on detection of functional anomalies in particular cases.

More experiments were run regarding the stability of the algorithm, but for sake of space, we describe them in Section~C of the Supplementary Materials.

\subsection{Real Data Benchmarking}\label{subsec:datasets}

To explore the performance of the proposed FIF algorithm, we conduct a comparative study using $13$ classification datasets from the UCR repository \citep{UCRArchive}. We consider the larger class as normal and some of others as anomalies (see Table~\ref{tab:datsets} for details). When classes are balanced, \textit{i.e} for 9 datasets out of 13, we keep only part of the anomaly class to reduce its size, always taking the same observations (at the beginning of the table) for a fair comparison. Since the datasets are already split into train/test sets, we use the train part (without labels) to build the FIF and compute the score on the test set. We assess the performance of the algorithm by measuring an Area Under the Receiver Operation Characteristic curve (AUC) on the test set. Both train and test sets are rarely used during learning in unsupervised setting since labels are unavailable when fitting the model. Thus, when fitting the models on unlabeled training data, good performances on the test set show a good generalization power.

\noindent{\bf Competitors}
FIF is considered with two finite size dictionaries \textit{dyadic indicator}, \textit{the self-data} and the infinite size dictionary \textit{cosines} (with $\alpha=1$ and $\alpha=0$); its parameters are set $N = 100$, $\psi = \text{min}(256,n)$ and the height limit to $=\lceil \log_2 (\psi) \rceil)$.
We contrast the FIF method with three most used multivariate anomaly detection techniques and two functional depths, with default settings. The multivariate methods---isolation forest (IF) \citep{LiuTZ08}, local outlier factor (LOF) \citep{Breunig}, and one-class support vector machine (OCSVM) \citep{SPSSW01}--- are employed after dimension reduction by Functional PCA keeping $20$ principal components with largest eigenvalues after a preliminary step of filtering using Haar basis. The depths are the random projection halfspace depth \citep{cuevas} and the functional Stahel-Donoho outlyingness \citep{rousseeuw}.

\noindent {\bf Analysis of the results}
Taking into account the complexity of the functional data, as expected there is no method performing generally best. Nevertheless, FIF performs well in most of the cases, giving best results for $10$ datasets and second best for $6$ datasets. It is worth to mention that the dictionary plays an important role in identifying anomalies, while FIF seems to be rather robust w.r.t. other parameters: The ``CinECGTorso'' dataset contains anomalies differing in location shift which are captured by the cosine dictionary. Dyadic indicator dictionary allows to detect local anomalies in ``TwoLeadECG'' and ``Yoga'' datasets. Self-data dictionary seems suited for Datasets ``SonyRobotAI2'' and ``StarlightCurves'' whose challenge is to cope with many different types of anomalies.

{\renewcommand{\arraystretch}{1} 
{\setlength{\tabcolsep}{0.1cm}
\begin{table}[!h]
\begin{center}
{\scriptsize
\begin{tabular}{|c||c|c|c|c|c|c|}
\hline
 & p & training : $n_a$/ $n$ & testing :  $n_a$/$n$&normal lab & anomaly lab \\
\hline
Chinatown & 24& 4 / 14 (29\%)  & 95 / 345 &  2& 1\\
\hline
Coffee & 286& 5 / 19 (26\%) & 6 / 19 & 1 & 0\\
\hline
ECGFiveDays & 136& 2 / 16 (12\%)& 53 / 481 & 1&  2  \\
\hline
ECG200 & 96&31 / 100 (31\%) & 36 / 100 &  1&  -1 \\
\hline
Handoutlines &2709 & 362 / 1000 (36 \%) & 133 / 370 &  1& 0\\
\hline
SonyRobotAI1& 70 & 6 / 20 (30 \%) & 343 / 601& 2 & 1\\
\hline
SonyRobotAI2 & 65 & 4 / 20 (20 \%) & 365 / 953 & 2 &  1\\
\hline
StarLightCurves & 1024 & 100 / 673 (15 \%) & 3482 / 8236 &  3 &  1 and 2\\
\hline
TwoLeadECG &82& 2 / 14 (14 \%) & 570 / 1139  &  1&  2\\
\hline
Yoga& 426 & 10 / 173 ( 06 \%) & 1393 / 3000 &  2 & 1\\
\hline
EOGHorizontal& 1250 & 10 / 40 (25 \%) & 30 / 61 &  5 &  6\\
\hline
CinECGTorso & 1639 & 4 / 16 (25 \%)  & 345 / 688&  3 &  4\\
\hline
ECG5000 & 140 & 31 / 323 (10 \%) & 283 / 2910 &  1 &  3,4 and 5 \\
\hline
\end{tabular}
}
\end{center}
\caption{Datasets considered in performance comparison: $n$ is the number of instances, $n_a$ is the number of anomalies. $p$ is the number of discretization points.}
\label{tab:datsets}
\end{table}

{\renewcommand{\arraystretch}{1} 
{\setlength{\tabcolsep}{0.1cm}
\begin{table}[!h]
\begin{center}
{\scriptsize
\begin{tabular}{|c|c|c|c|c|c|c|c|c|c|}
\hline
Methods : & DI$_{L_{2}}$   & Cos$_{Sob}$ & Cos$_{L_{2}}$ & Self$_{L_{2}}$ & IF & LOF&OCSVM &fHD$_{RP}$ & fSDO\\
\hline
Chinatown & \textit{\textbf{0.93}} &0.82 &0.74& 0.77 &0.69&0.68& 0.70&0.76&\textbf{0.98}\\
\hline
Coffee & 0.76& \textbf{0.87}& 0.73&\textit{\textbf{0.77}}&0.60&0.51&0.59&0.74&0.67\\
\hline
ECGFiveDays & 0.78 &0.75 &0.81& 0.56&0.81&\textit{\textbf{0.89}}&\textbf{0.90}&0.60&0.81\\
\hline
ECG200 & 0.86&\textbf{0.88}& \textbf{0.88} & 0.87 &0.80&0.80&0.79&0.85&0.86\\
\hline
Handoutlines &0.73&\textbf{0.76}&0.73&0.72&0.68&0.61&0.71&0.73&\textbf{0.76}\\
\hline
SonyRobotAI1 & \textit{\textbf{0.89}} &0.80&0.85&0.83 &0.79&0.69&0.74&0.83&\textbf{0.94}\\
\hline
SonyRobotAI2 & 0.77 &0.75&0.79&\textbf{0.92}&\textit{\textbf{0.86}}&0.78&0.80&\textit{\textbf{0.86}}&0.81\\
\hline
StarLightCurves &0.82 &0.81 &0.76 &\textbf{0.86} & 0.76 & 0.72&0.77&0.77&\textit{\textbf{0.85}} \\
\hline
TwoLeadECG & \textbf{0.71} & 0.61&0.61&0.56 &\textbf{0.71} &0.63&\textbf{0.71}&0.65&0.69\\
\hline
Yoga & \textbf{0.62}  &0.54 &0.60 &0.58&0.57&0.52&\textit{\textbf{0.59}}&0.55&0.55\\
\hline
EOGHorizontal & 0.72 & \textit{\textbf{0.76}} &\textbf{0.81}& 0.74  &0.70&0.69&0.74&0.73&0.75\\
\hline
CinECGTorso & 0.70   &\textbf{0.92}&\textit{\textbf{0.86}}& 0.43&0.51&0.46&0.41&0.64&0.80\\
\hline
ECG5000& 0.93 &\textbf{0.98}&\textbf{0.98}&0.95  &0.96&0.93&0.95&0.91&0.93\\
\hline
\end{tabular}
}
\end{center}
\caption{AUC of different anomaly detection methods calculated on the test set. Bold numbers correspond to the best result while italics to the second best.}
\label{tab:bench}
\end{table}

\section{Extensions of FIF}\label{sec:mulext}

\noindent{\bf Extension to multivariate functions} FIF can be easily extended to the multivariate functional data, \textit{i.e.} when the quantity of interest lies in $\mathbb{R}^d$ for each moment of time. For this, the coordinate-wise sum of the $d$ corresponding scalar products is used to project the data onto a chosen dictionary element: $\langle \mathbf{f},\mathbf{g} \rangle_{\mathcal{H}^{\otimes d}} := \sum_{i=1}^{d} \langle f^{(i)},g^{(i)} \rangle_{\mathcal{H}}$.
The dictionary is then defined in $(\mathcal{H}([0,1]))^{\otimes d}$, \textit{e.g.}, by componentwise application of one or several univariate dictionaries, see Section~\ref{sec:fun}. In the Supplementary Materials we give an illustration of multivariate functional anomaly detection on the MNIST~\citep{Lecun} dataset, each digit being seen as a 2D-curve.

\noindent{\bf Connection to data depth} Regarding FIF score as an anomaly ranking yields a connection to the notion of the \textit{statistical depth function} (see \citep{Mosler2013} for an overview), which has been successfully applied in outlier detection (see, \textit{e.g.}, \citep{rousseeuw}). Statistical data depth has been introduced as a measure of centrality (or depth) of an arbitrary observation $\mathbf{x}\in(\mathcal{H}([0,1]))^{\otimes d}$ with respect to the data at hand $\mathcal{S}$. A data depth measure based on FIF score can be defined for (multivariate) functional data as: $D_{FIF}(\mathbf{x};\mathcal{S}) = 1 - s_n(\mathbf{x};\mathcal{S})$.
\noindent
Data depth proves to be a useful tool for a low-dimensional data representation called {\it depth-based map}. Using this property, \cite{LI} and \cite{mozha} define a $DD$-plot classifier which consists in applying a multivariate classifier to the depth-based map.  Low-dimensional representation is of particular interest for functional data and a $DD$-plot classifier can be defined using the FIF-based data depth. Let $\mathcal{S}^{trn} = \mathcal{S}^{1} \cup ... \cup \mathcal{S}^{q}$ be a training set for supervised classification containing $q$ classes, each subset  $\mathcal{S}^{j}$ standing for class $j$. The depth map is defined as follows:
\[
	\mathbf{x} \mapsto \phi(\mathbf{x}) = \bigl(D_{FIF}(\mathbf{x};\mathcal{S}^{1}),...,D_{FIF}(\mathbf{x};\mathcal{S}^{q})\bigr)\in[0,1]^q\,.
\]

As an illustration, we apply the depth map to $3$ digits ($1$, $5$ and $7$, $100$ observations per digit for training and $100$ testing) of the MNIST dataset after their transformation to two-variate functions using \texttt{skimage} python library (see Figure ~\ref{fig:mnist2} ). One observes appealing geometrical interpretation (observe, \textit{e.g.}, the location of the abnormally distant---from their corresponding classes---observations) and a clear separation of the classes. To illustrate separability, we apply linear multiclass (one-against-all) SVM in the depth space, which delivers the accuracy of $99\%$ on the test data.

\begin{figure}[!h]
\begin{center}
\includegraphics[width=.45\textwidth, trim=2cm 1cm 2cm 3cm,clip=true]{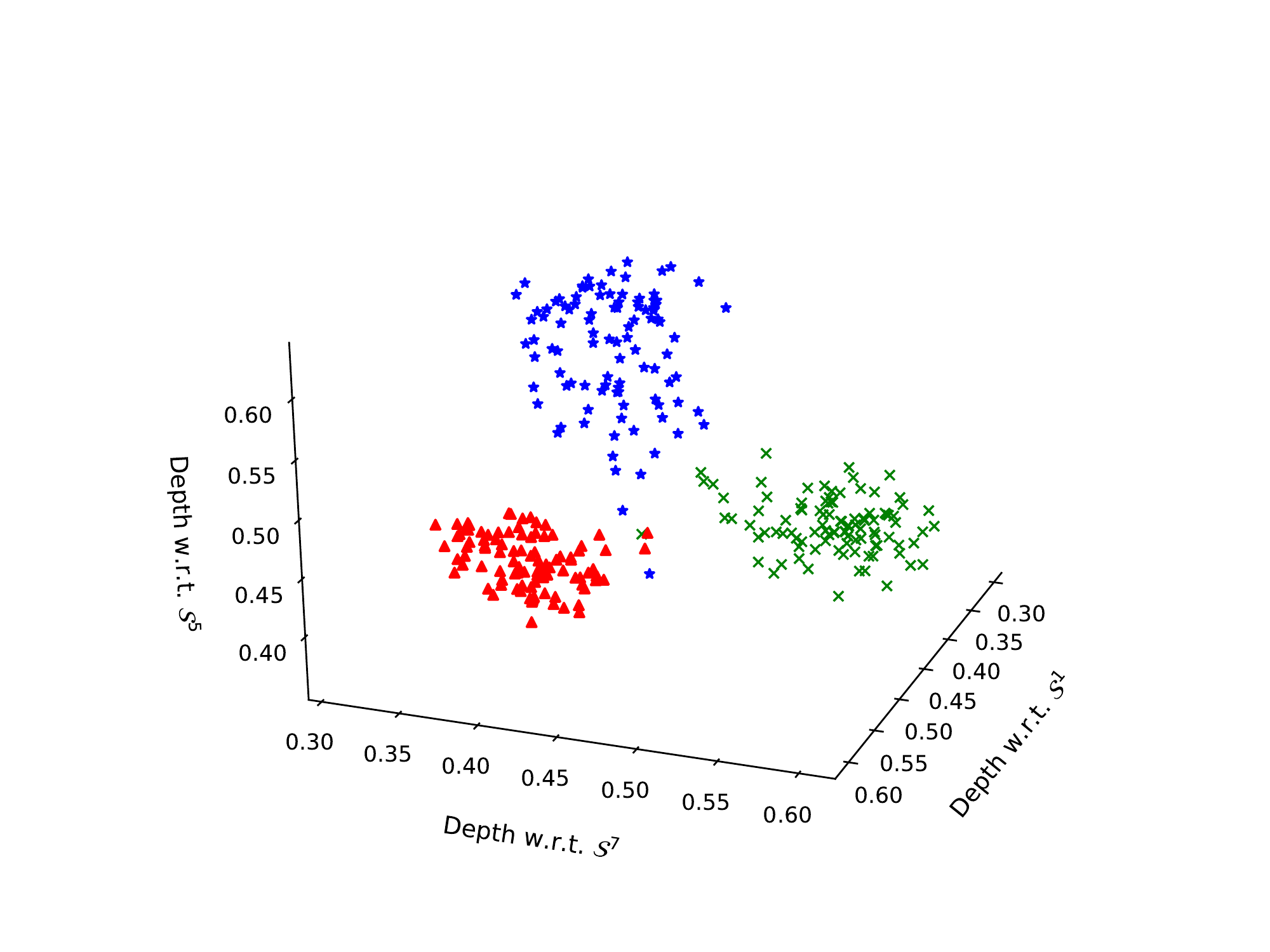}
\end{center}
\caption{Depth space embedding of the three digits ($1$, $5$ and $7$) of the MNIST dataset.}
\label{fig:mnist2}
\end{figure}

\section{Conclusion}\label{sec:conc}

The Functional Isolation Forest algorithm has been proposed, which is an extension of Isolation Forest to functional data. The combined choice of the dictionary itself, the probability distribution used to pick a {\it Split variable} and the scalar product used for the projection enables FIF to exhibit a great flexibility  in detecting anomalies for a variety of tasks. FIF is extendable to multivariate functional data. When transformed in a data depth definition, FIF can be used for supervised classification via a low-dimensional representation---the depth space. The open-source implementation of the method, along with all reproducing scripts, can be accessed at https://github.com/Gstaerman/FIF.

\bibliography{FiF-ACML_Arxiv}

\clearpage

\setcounter{section}{0}
\renewcommand{\thesection}{\Alph{section}}

\section*{\LARGE Supplementary material}

\section{Illustrative figures}

An example of a functional dataset---the ``Chinatown'' dataset form the UCR \cite{UCRArchive} repository---and evaluation points is plotted in Figure~\ref{fig:china}. Figure~\ref{fig:treestr} depicts a sample isolation tree with a selected terminal node.

\begin{figure}[!h]
\begin{center}
\includegraphics[trim=4cm 2cm 4cm 2cm,clip=true,scale=0.2]{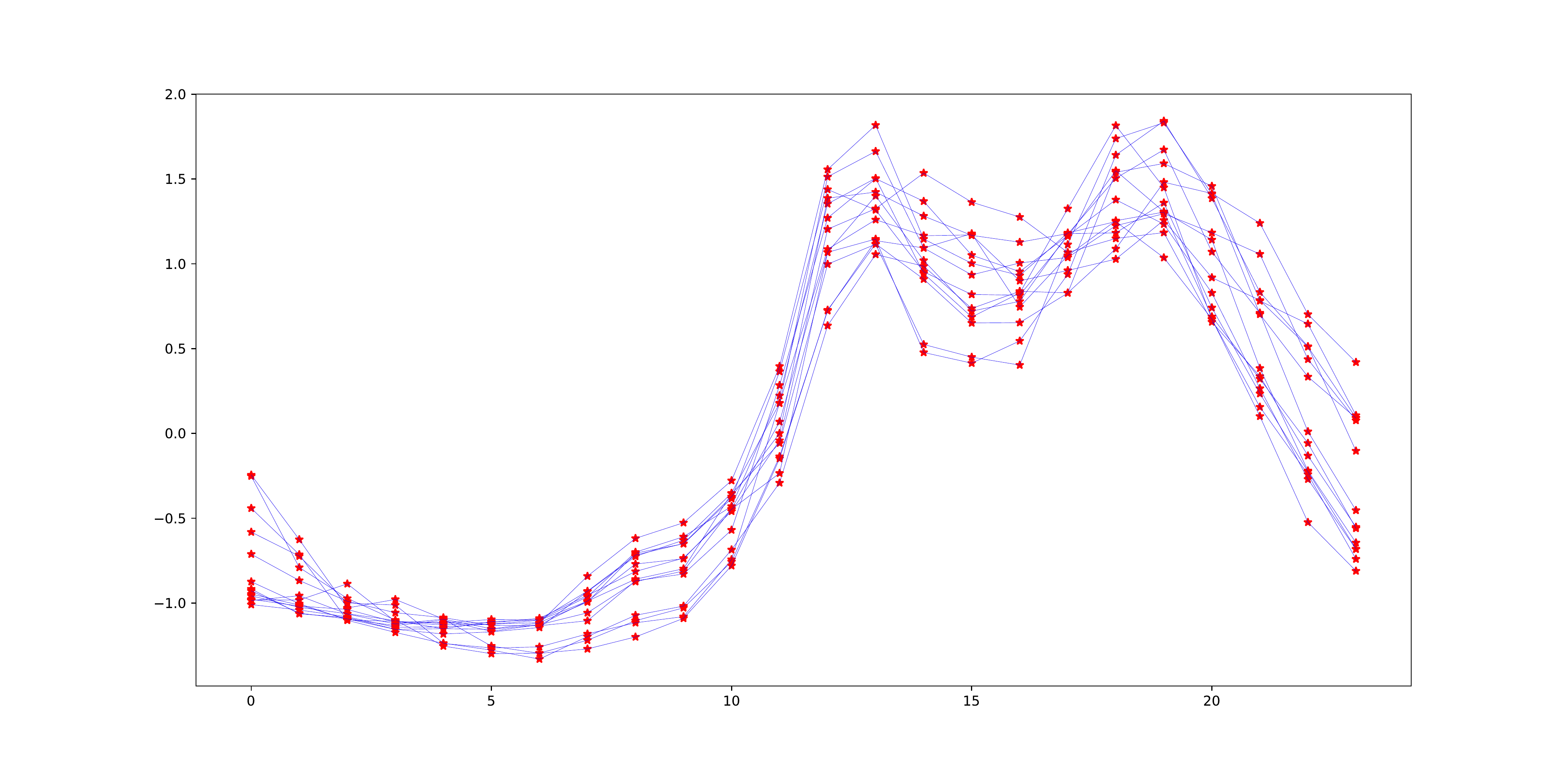}
\end{center}
\caption{Example of a functional dataset. Red stars indicate the observed values (at the same equispaced time points), while the blue curves are obtained by linearly interpolating them.}
\label{fig:china}
\end{figure}

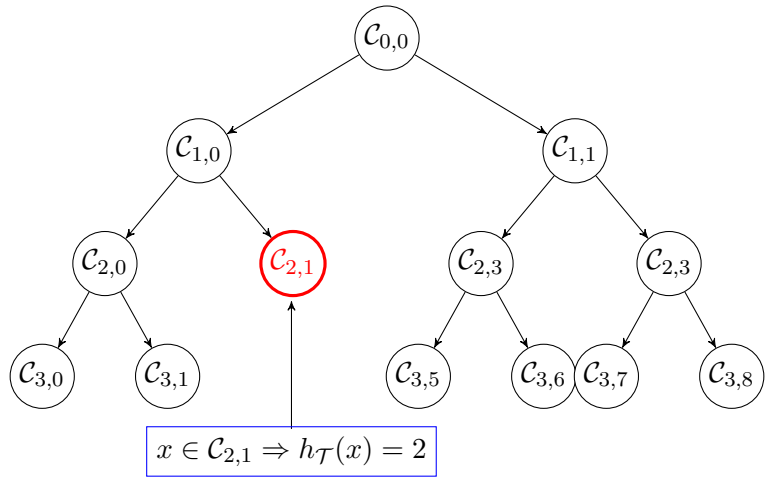
\begin{figure}[!h]
\begin{center}
\begin{tikzpicture}[->,>=stealth',level/.style={sibling distance = 5cm/#1,
  level distance = 1.5cm}] 
  \draw(-1.27,-5.2) -- (-1.27,-3.5);
  \node[draw = blue] at (-1.27,-5.5)  {$x\in\mathcal{C}_{2,1} \Rightarrow h_{\mathcal{T}}(x)=2$};

\node [arn_n] {$\mathcal{C}_{0,0}$}
    child{ node [arn_n] {$\mathcal{C}_{1,0}$} 
            child{ node [arn_n] {$\mathcal{C}_{2,0}$} 
            	child{ node [arn_n] {$\mathcal{C}_{3,0}$} 
                         } 
							child{ node [arn_n] {$\mathcal{C}_{3,1}$}}
            }
            child{ node [arn_r] {$\mathcal{C}_{2,1}$}
							child[missing]{}
							child[missing]{}
            }                            
    }
    child{ node [arn_n] {$\mathcal{C}_{1,1}$}
            child{ node [arn_n] {$\mathcal{C}_{2,3}$} 
							child{ node [arn_n] {$\mathcal{C}_{3,5}$}}
							child{ node [arn_n] {$\mathcal{C}_{3,6}$}}
            }
            child{ node [arn_n] {$\mathcal{C}_{2,3}$}
							child{ node [arn_n] {$\mathcal{C}_{3,7}$}}
							child{ node [arn_n] {$\mathcal{C}_{3,8}$}}
            }
		}
; 
\end{tikzpicture}
\end{center}
\caption{An example of a functional isolation tree structure denoted by $\mathcal{T}$. Here, $\mathcal{C}_{2,1}$ is a cell associated with a terminal node.}
\label{fig:treestr}
\end{figure}

\section{Presentation of used dictionaries}

In this part, we define properly every dictionaries used in the paper.

\textit{Self-data dictionary} (Self) consisting of the training dataset itself.

\vspace*{0.4cm}

\textit{Brownian motion dictionary} (B) is a combination of the space of continuous function $\mathcal{D}=\mathcal{C}([0,1])$ and the Wiener measure $\mathcal{W}$ on $\mathcal{D}$.

\vspace*{0.4cm}

\textit{Brownian bridge dictionary} (BB) is a combination of the space of continuous function $\mathcal{D}=\mathcal{C}([0,1])$ and the Brownian bridge measure $\mathcal{G}$ on $\mathcal{D}$.

\vspace*{0.4cm}

\textit{Cosine dictionary} (Cos) consisting of curves with the following forms:

$$ \mathbf{x}_{a, \omega}(t) = a \cos(2\pi \omega t)$$

with $a \in [0,1]$ and $\omega \in [0,10]$.

\vspace*{0.4cm}

\textit{Mexican hat wavelet dictionary} (MHW) consists of the negative second derivatives of the normal density, shifted and scaled in a appropriate fashion:
\[
	\mathbf{x}_{\theta, \sigma}(t) = \frac{2}{\sqrt{3\sigma}\pi^{1/4}} \left(1-\left(\frac{t-\theta}{\sigma} \right)^{2} \right) \exp \left( \frac{-(t-\theta)^{2}}{2\sigma^{2}}\right)\,
\]
with $\theta\in [-0.8,0.8]$ and $\sigma \in ([0.04,0.2])$.

\vspace*{0.4cm}

\textit{Dyadic indicator dictionary} (DI) consisting of a set of indicator functions on the elements of binary partitioning, for a given $J$ (chosen according to the granularity to be captured or from the discretisation considerations) having as elements $\{ \left(\mathbf{x}_{k,j}\right)_{0 \leq k < 2^{j} } \}_{1 \leq j \leq J}$:
\[
	\mathbf{x}_{k,j}(t) = \mathbbm{1}\Bigl(t\in \Bigl[\frac{k}{2^{j}},\frac{k+1}{2^{j}}\Bigr]\Bigr)\,.
\]

\vspace*{0.4cm}

\textit{Uniform indicator dictionary} (UI) consists of indicator function on $[a,b]$ where $a$ and $b$ are choosen uniformly on $[0,1]$ such that $a<b$.

\vspace*{0.4cm}

\textit{Dyadic indicator derivative} (DId) consisting of a set of indicator functions on the elements of binary partitioning, for a given $J$ (chosen according to the granularity to be captured or from the discretisation considerations) having as elements $\{ \left(\mathbf{x}_{k,j}\right)_{0 \leq k < 2^{j} } \}_{1 \leq j \leq J}$:
\[
	\mathbf{x}_{k,j}(t) = t\mathbbm{1}\Bigl(t\in \Bigl[\frac{k}{2^{j}},\frac{k+1}{2^{j}}\Bigr]\Bigr)\,.
\]

\vspace*{0.4cm}

\textit{Uniform indicator derivative} (UId) consists of functions $t\mapsto t$ on $[a,b]$ where $a$ and $b$ are choosen uniformly on $[0,1]$ such that $a<b$.

\section{Further discussion on the choice of dictionary}
To illustrate the dicussion on dictionaries, especially the incorporation of stochastic elements and external informations, we bring an example of the use of the \textit{Brownian motion dictionary}. Let $\mathcal{W}$ be the Wiener measure defined on $\mathcal{C}([0,1])$ the space of continuous function on [0,1] and $\mathcal{H}$ be the $L_{2}$ space. We define by \textit{Brownian motion dictionary} (B) the \textit{Split variables} space induced by $\boldsymbol{\nu} = \mathcal{W}$ and $\dict=\mathcal{C}([0,1])$.
Although seeming universal, this dictionary explores almost the entire argument space equivalently, and in practice can be unable to detect \textit{isolated anomalies}. On Fig.~\ref{fig:isolated} we plot the following synthetic dataset:

\begin{itemize}
\item 30 curves defined by $\mathbf{x}(t) = 30 (1-t)^{q}t^{q}$ on $t\in [0,0.2]$ and $\mathbf{x}(t)= 30 (0.8)^{q}0.2^{q}+ \mathcal{N}(0,0.3^{2})$ on $t\in [0.2,0.7]$ with $q$ equispaced in $[0.5,0.55]$.
\item 1 \textit{abnormal} curve with the same shape but that is shifted at the beginning and whose continuation is deep in the $30$ preceding curves.
\end{itemize}

\begin{figure}
\begin{center}
\includegraphics[height=.2\textheight, trim=.5cm .6cm 1cm 1cm,clip=true]{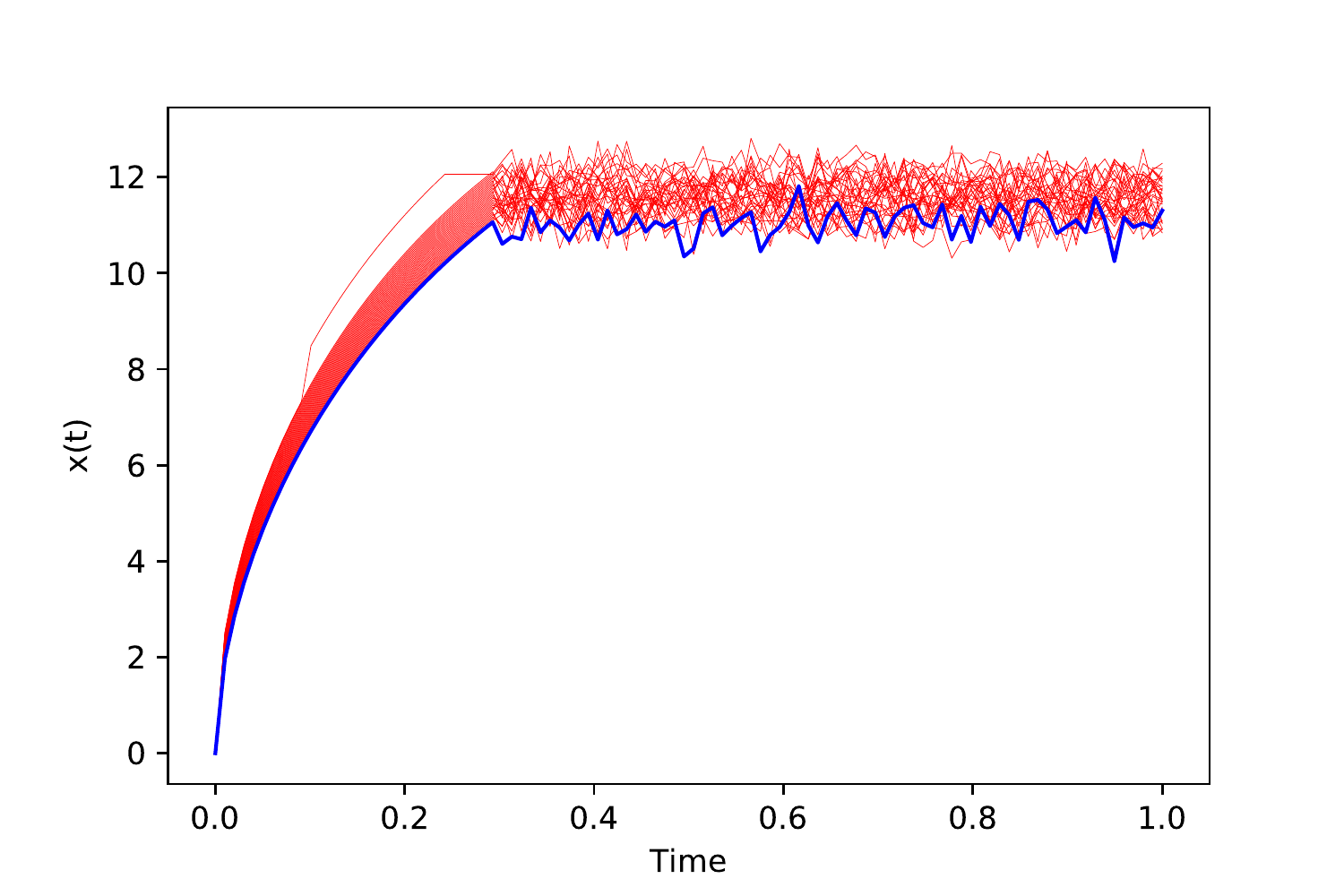}\,\includegraphics[height=.2\textheight, trim=1cm .6cm 1cm 1cm,clip=true]{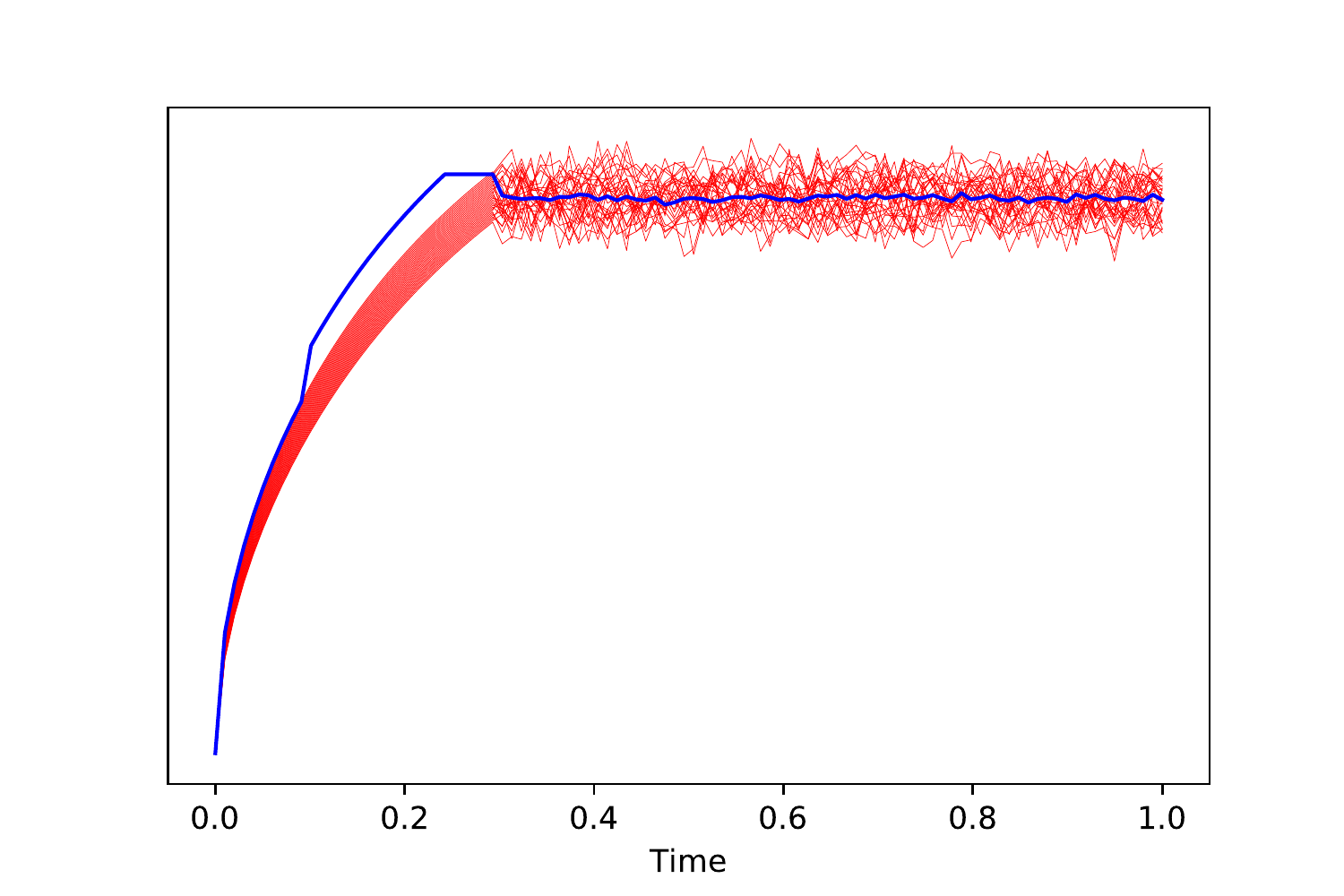}
\end{center}
\caption{Synthetic data containing an isolated anomaly and the observation having the highest anomaly score, with dictionary being pure Brownian motion (left) and Brownian motion mixed with an indicator function in the area of interest in proportions $4$ to $1$ (right).}
\label{fig:isolated}
\end{figure}

One can see that the anomaly is not detected, that indicated as anomaly curve (the one with highest anomaly score) is on the fringe of the dataset though. Illustrative incorporation of the prior knowledge, in its simplified version, can consist, \textit{e.g.}, in adding to the measure $\mathcal{W}$, a Dirac of the indicator function on the interval of interest $\tilde{\mathbf{x}}(t)=\mathbbm{1}(t\le0.25)$ with weights: $\widetilde{\mathcal{W}}:=0.2\mathcal{W}+0.8\delta_{\tilde{\mathbf{x}}}$  ; this assigns the highest anomaly score to the desired observation. In the sequel, $\boldsymbol{\nu}$ follows a uniform distribution if is not explicitly mentioned.

\vspace*{0.8cm}

When having not enough prior knowledge, \textit{e.g.} just knowing to stick to local features of functional data but not the precise interval, one would like to use a dictionary exploring different localities. To illustrate possible advantage of this approach, we use \textit{Mexican hat wavelet} and  \textit{Dyadic indicator} dictionaries. Regard the ``Chinatown'' dataset \cite{UCRArchive}, which represents pedestrian count in Chinatown-Swanston St North for $12$ months during year $2017$. 
With $14$ functions (working days) representing normal observations and taking $4$ functions (weekends) as anomalies (Figure~\ref{fig:Chinatown}). One observes that while the Mexican hat wavelet dictionary correctly detects part of the anomalies, due to its smooth nature it is distracted by two normal curves with high deviation on the second half of the domain. Having straight fronts and begin non-zero only in a small part of the domain, the dyadic indicator dictionary detects all four abnormal observations. Nevertheless, it is not adapted to scalar product that involves derivative. To adapt the dyadic indicator dictionary and the uniform indicator dictionary for the scalar product involving derivatives, we define their \textit{slope} versions being $\mathbf{x}_{k,j}(t) = t\mathbbm{1}\Bigl(t\in \Bigl[\frac{k}{2^{j}},\frac{k+1}{2^{j}}\Bigr]\Bigr)$ and $\mathbf{x}(t) = t\mathbbm{1}(t\in[t_1,t_2])$, respectively, with the notation defined above (\textit{i.e.} their derivatives become indicator functions). Clearly, this list can be extended with further task-specific dictionaries.

\begin{figure}[h]
\begin{center}
\includegraphics[height=.1375\textheight, trim=0cm 0cm 1cm 1.15cm,clip=true]{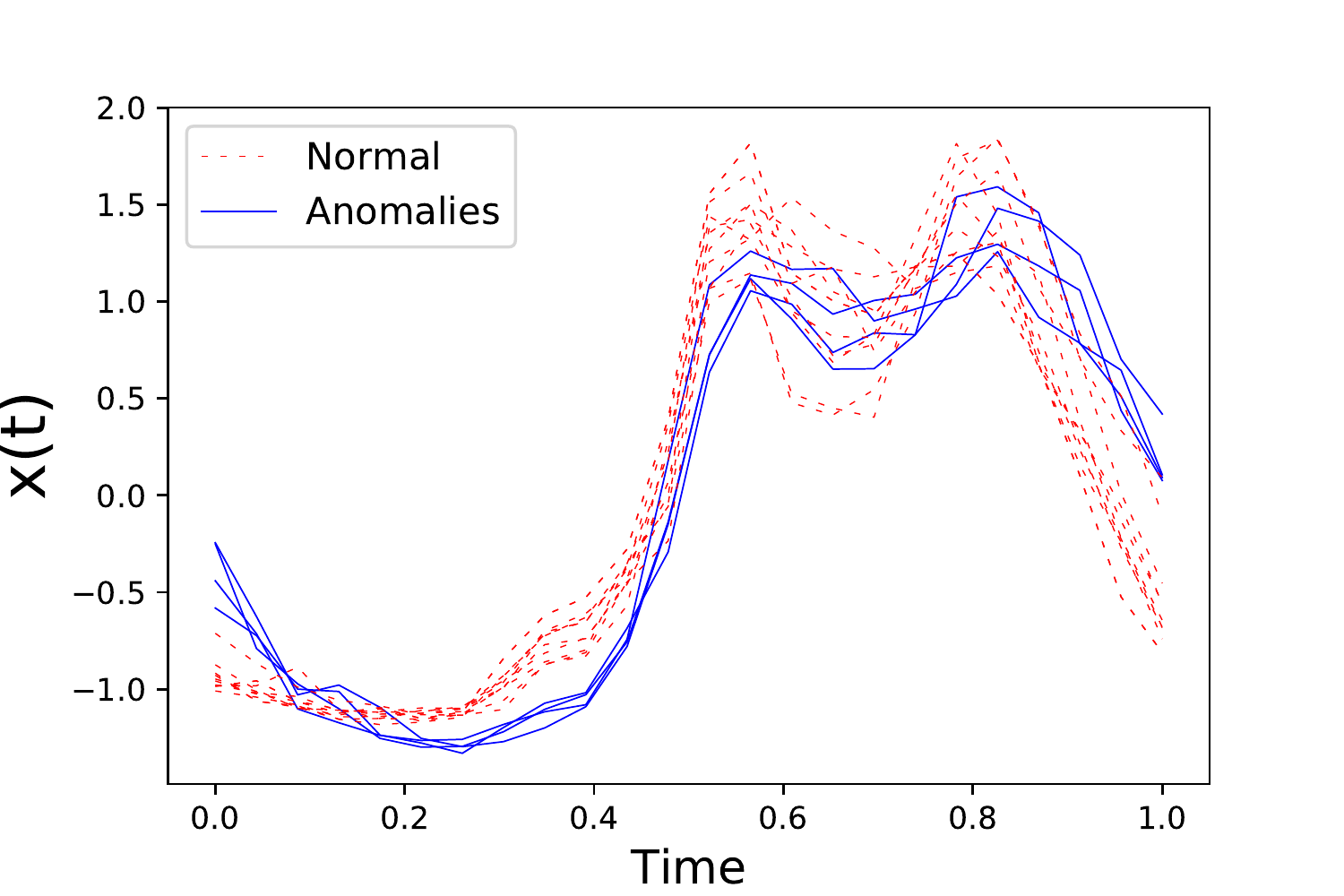}\,\includegraphics[height=.1375\textheight, trim=1cm 0cm 1cm 1.15cm,clip=true]{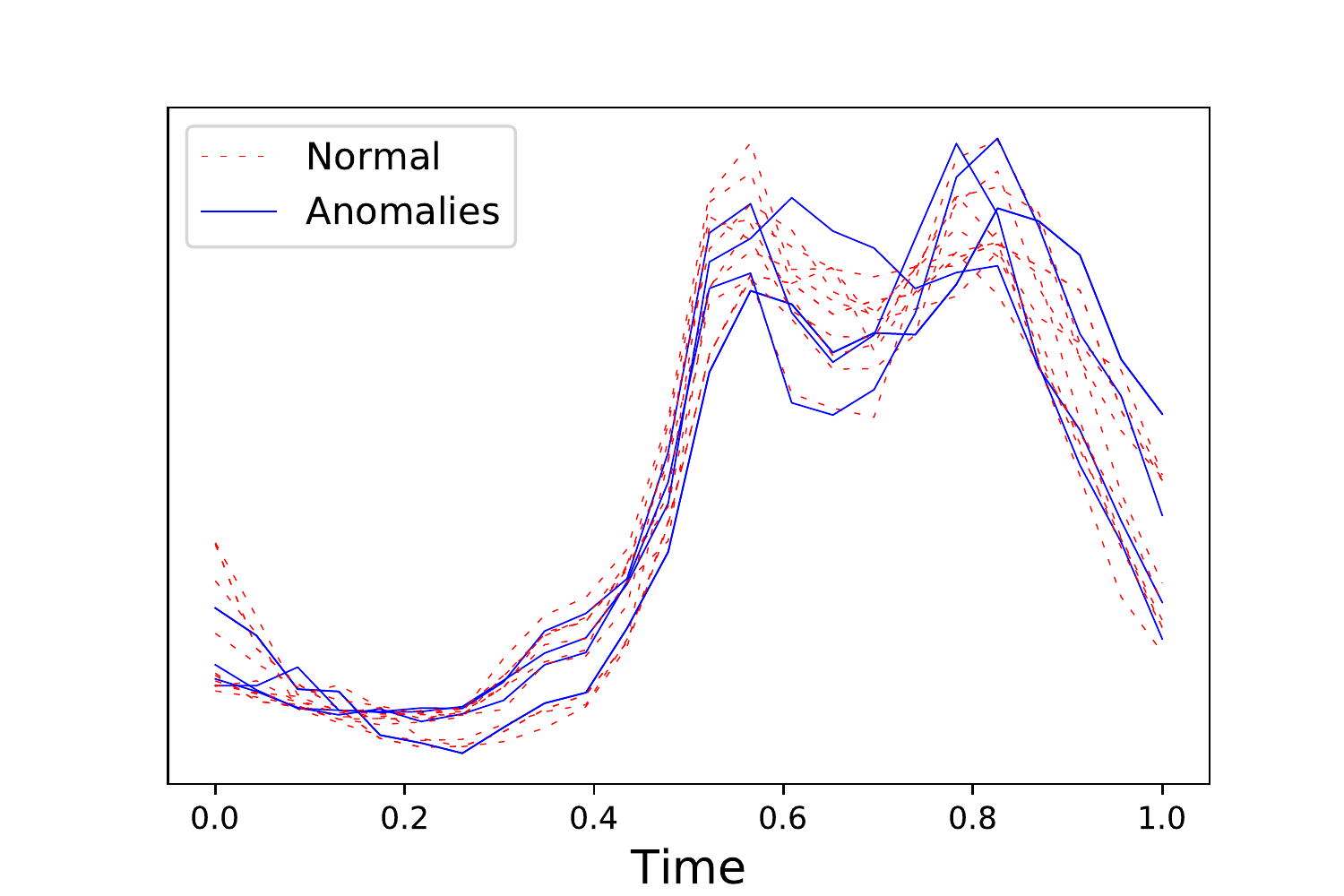}\,\includegraphics[height=.1375\textheight, trim=1cm 0cm 1cm 1.15cm,clip=true]{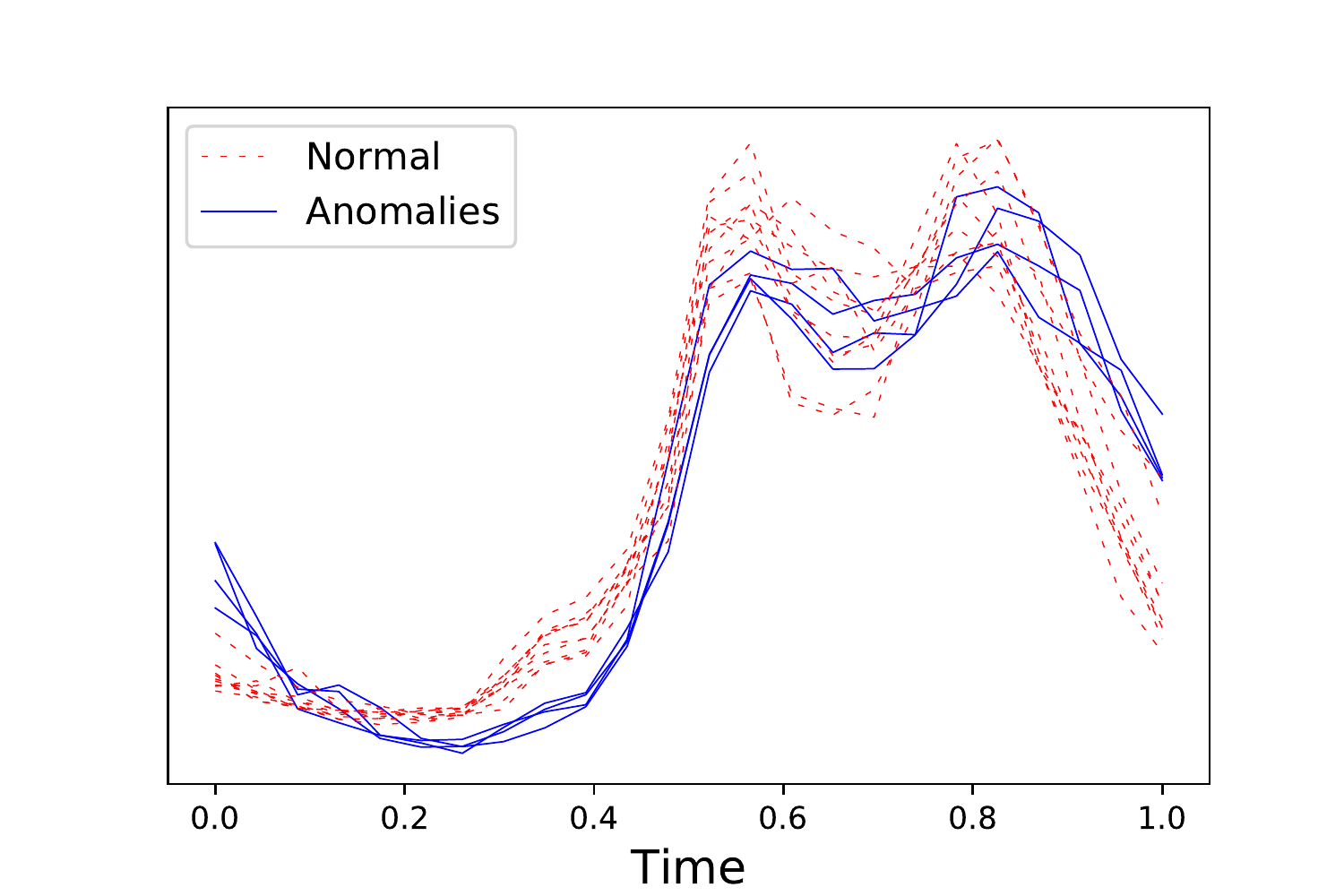}
\end{center}
\caption{ The ``Chinatown'' dataset, with normal observations in red and anomalies in blue: The data and the true anomalies (left), anomalies detected using the Mexican hat wavelet dictionary (middle), and anomalies detected using the dyadic indicator dictionary (right).}\label{fig:Chinatown}
\end{figure}

Before, we were considering dictionaries that are independent of data. Nevertheless one can use observations or their certain transform as a dictionary itself: projections on both normal and abnormal observations shall differ for normal ones and for anomalies; this suggests the \textit{self-data dictionary} (Self). This can be extended to the \textit{local self-data dictionary} which consists of the product of the self-data dictionary with the uniform indicator dictionary. As an example, we apply this to the ``ECG5000'' dataset plotted in Figure~\ref{fig:selfdict}, where, different to the cosine dicitonary, it allows to detect all abnormal observations.

\begin{figure}[h]
\begin{center}
\includegraphics[height=.1375\textheight, trim=0cm 0cm 1cm 1.15cm,clip=true]{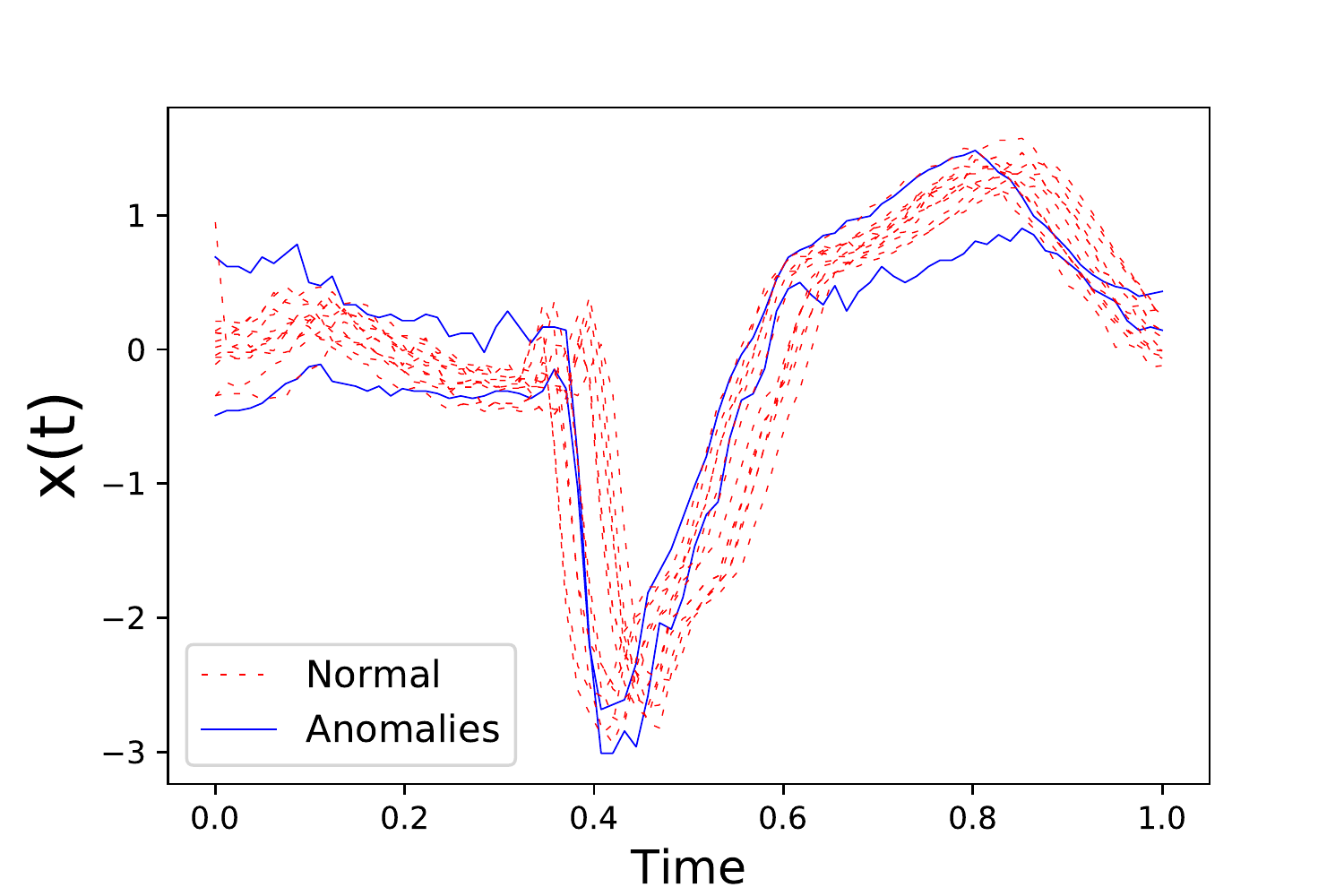}\,\includegraphics[height=.1375\textheight, trim=1cm 0cm 1cm 1.15cm,clip=true]{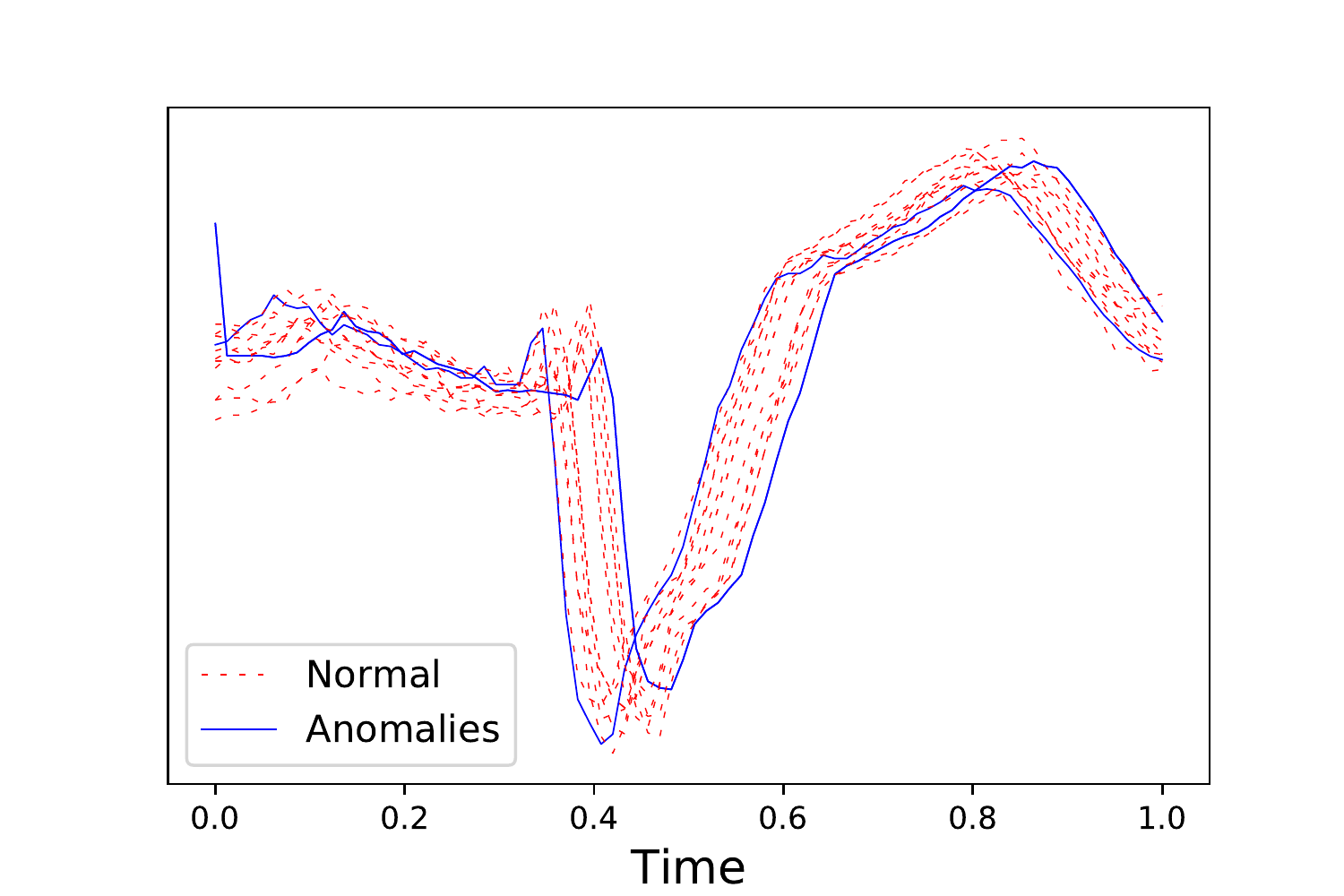}\,\includegraphics[height=.1375\textheight, trim=1cm 0cm 1cm 1.15cm,clip=true]{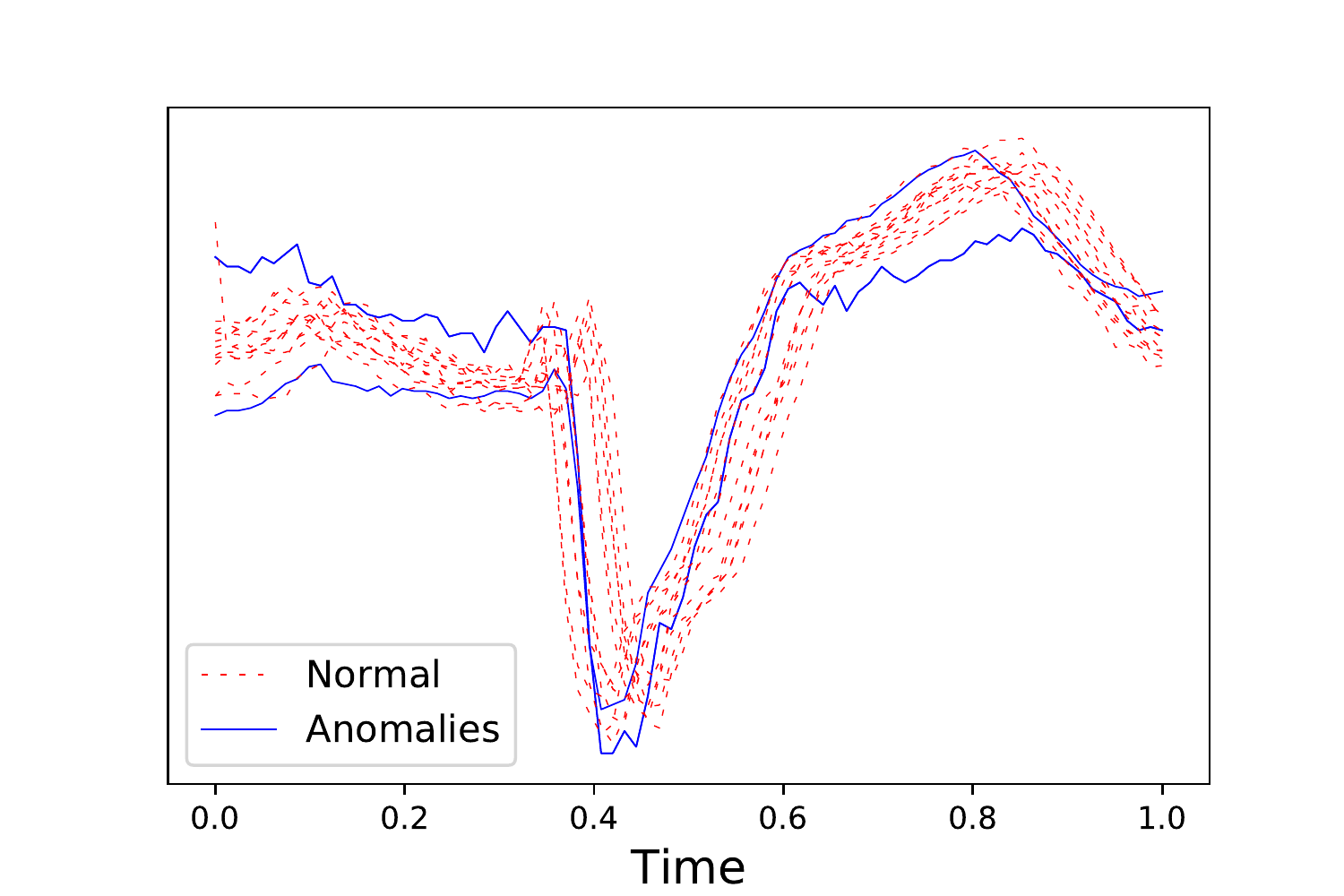}
\end{center}
\caption{The ``ECG5000'' dataset, with normal observations in red and anomalies in blue: The data and the true anomalies (left), anomalies detected using the cosine dictionary (middle), and anomalies detected using the self data dictionary (right).}
\label{fig:selfdict}
\end{figure}

To conclude this, we provide a last example (see Fig.~\ref{fig:cmpalpha}) where we highlight the impact of the scalar product choice. To illustrate the score change caused by different values of $\alpha$, we calculate the FIF anomaly scores with $\alpha=1$ and $\alpha=0$ for a sample consisting of $100$ curves as follows (inspired by \cite{cuevas}, see Fig.~\ref{fig:cmpalpha}):
\begin{itemize}
\item 90 curves defined by $\mathbf{x}(t) = 30 (1-t)^{q}t^{q}$ with $q$ equispaced in $[1,1.4]$,
\item 10 \textit{abnormal} curves defined by $\mathbf{x}(t) = 30 (1-t)^{1.2}t^{1.2}$ noised by $\varepsilon \sim \mathcal{N}(0,0.3^{2})$ on the interval $[0.2,0.8]$.
\end{itemize}
One can see that even though the $10$ noisy curves are abnormal for the majority of the data, they are considered as normal ones when only location is taken into account. On the other hand, they are easily distinguished with the high anomaly score when derivatives are examined.
\begin{figure}[t]
\begin{center}
\includegraphics[width=.45\textwidth, trim=0.8cm .6cm 1cm 1cm,clip=true]{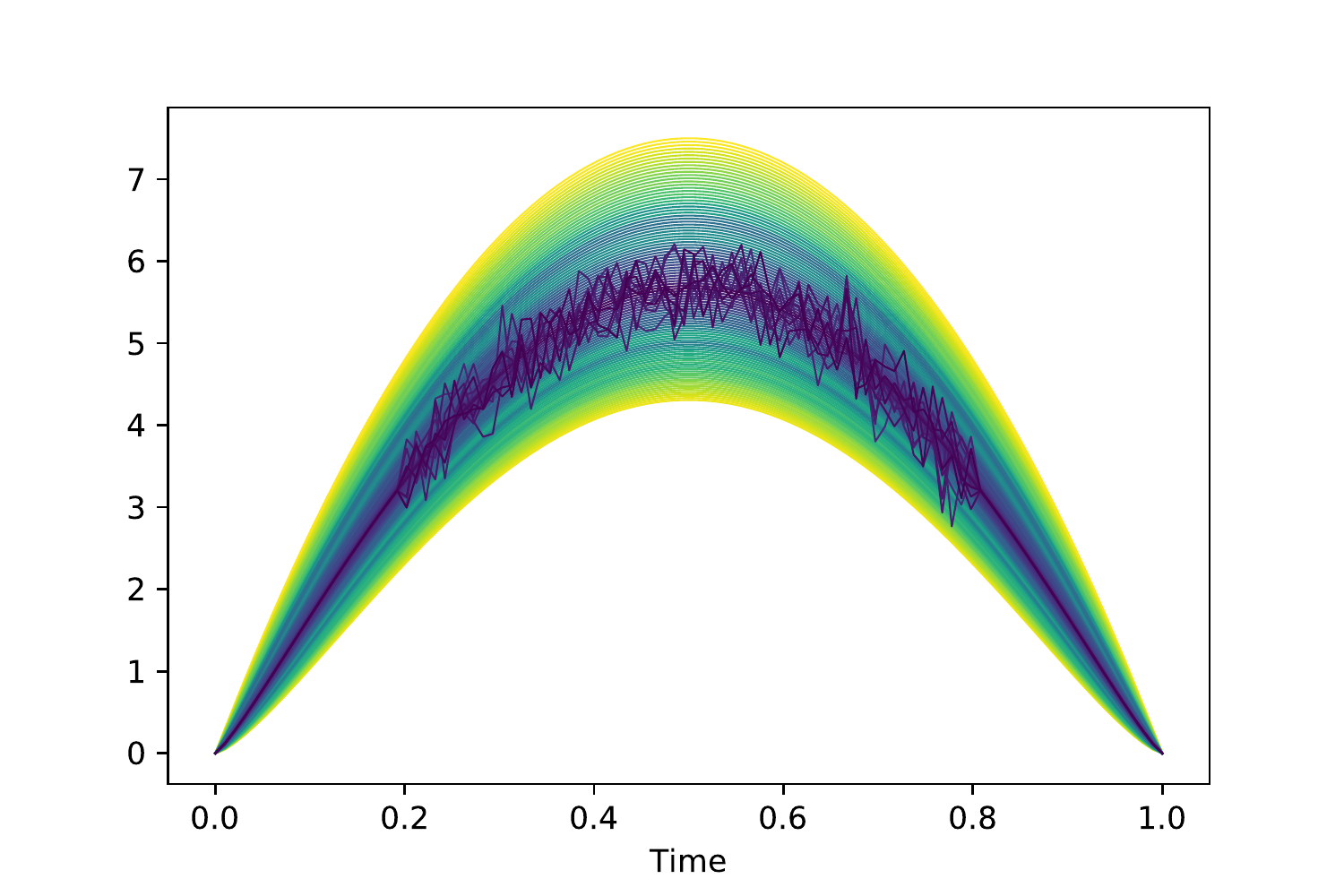}\,\includegraphics[width=.45\textwidth, trim=0.8cm .6cm 1cm 1cm,clip=true]{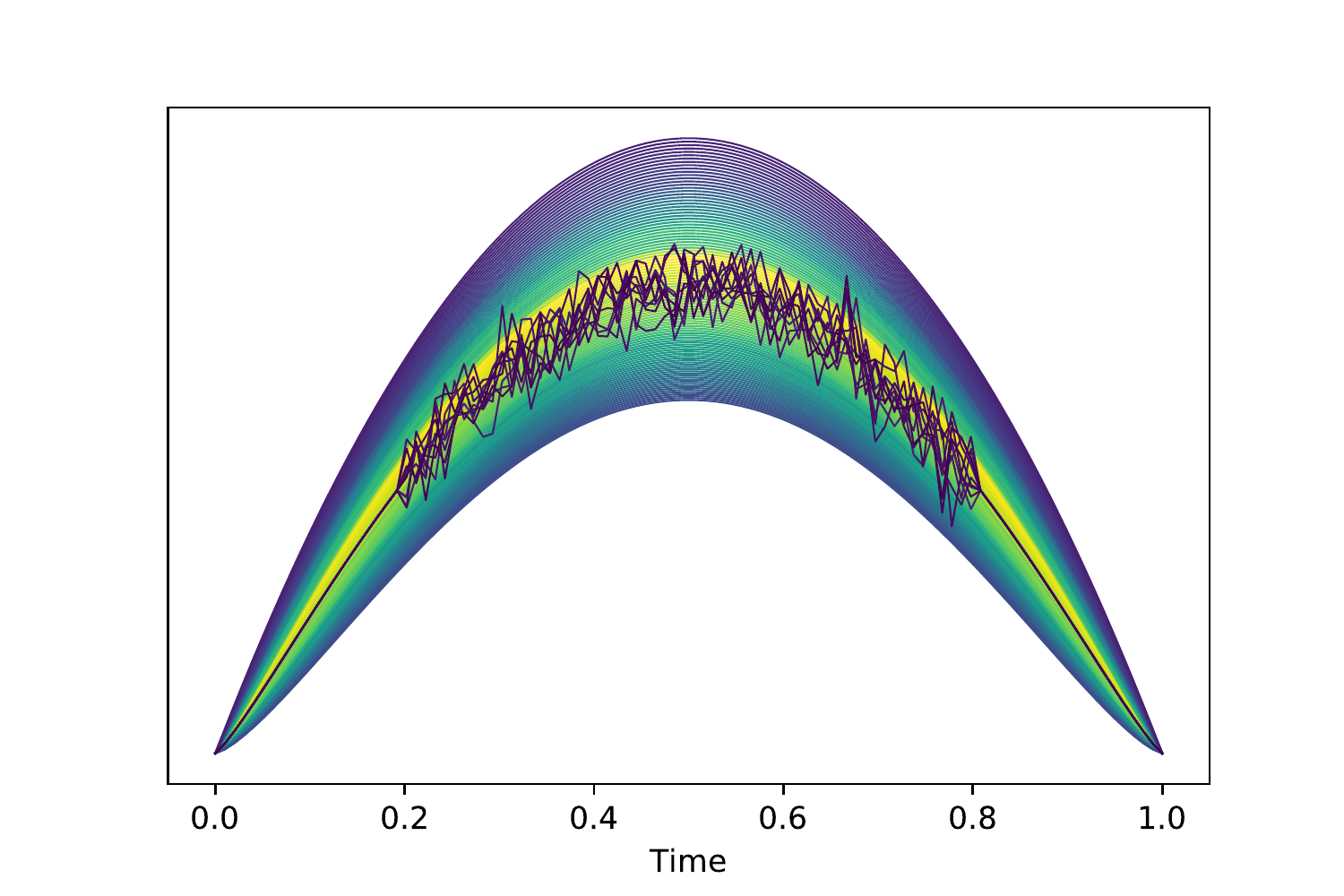}
\end{center}
\caption{FIF anomaly scores for a sample of $100$ curves with $\alpha=1$ (left) and $\alpha=0$ (right). Anomaly score increases from magenta to yellow in the left plot and decreases in the right plot.}\label{fig:cmpalpha}
\end{figure}

\subsection{Direction importance of finite size Dictionaries}\label{subsec:dirimportance}

Although feature importance have been tackled in supervised random trees (see \textit{e.g.} \cite{Br01}, \cite{Geurts06}), this has not been adressed in the \textit{Isolation Forest} literature (see  \cite{LiuTZ08}, \cite{LiuTZ12} and \cite{Hariri}). As a very randomized procedure, there is no incrementally way to define feature importance from the supervised setting. Nevertheless, it is a matter of interest in many anomaly detection applications to get interpretability of models, especially when dealing with functional data where many information are contained in curves. Thus, it is rewarding to get an \textit{a posteriori} sparse representation of the dictionary $\mathcal{D}$ which corresponds to the discriminating directions that have great importance in the construction of the model. Furthermore, it could bring some information on the distribution of normal data by studying the dispersion of the projection coefficients on a direction $d$ (\textit{e.g} multi-modality). To extend this notion to the \textit{Functional Isolation Forest} algorithm , we propose two ways to evaluate the importance of the elements of $\mathcal{D}$ to discriminate anomaly curves. The general idea is to give importance to elements of $\mathcal{D}$ which allows to discriminate between the sample. The naive idea is to add "+1" to the elements of $\mathcal{D}$ where an instance of the node sample is isolated (except for the cells with only two instances) such that good directions are those with a high score (after the forest construction). A clever one, more adaptive, would be to get weighted gain since curves isolated at nodes closer to the root should be more rewarding. To do this, we choose to give a reward depending on the size of the sample node where a curve is isolated. Precisely, the given reward is equal to the size of the node sample divided by the (sub)-sample used to build the tree. An example of the latter is given in Figure ~\ref{fig:feature_importance}. The experiment is conducted on the real-world  CinECGTorso dataset (more details in Section 4.2). We use FIF with the \textit{Dyadic indicator dictionary} and the $L_2$ scalar product. As we can see, the two most important elements of the dictionary are indicator functions which localize the peak around $t=0.4$ where anomalies are really different from the normal ones. These leads to some interpretability of a "black box" procedure.

\begin{figure}[!h]
\begin{center}
\begin{tabular}{ccc}

(a)&(b)&(c)\\
\includegraphics[height=.1375\textheight, trim=0cm 0cm 1cm 1.15cm,clip=true]{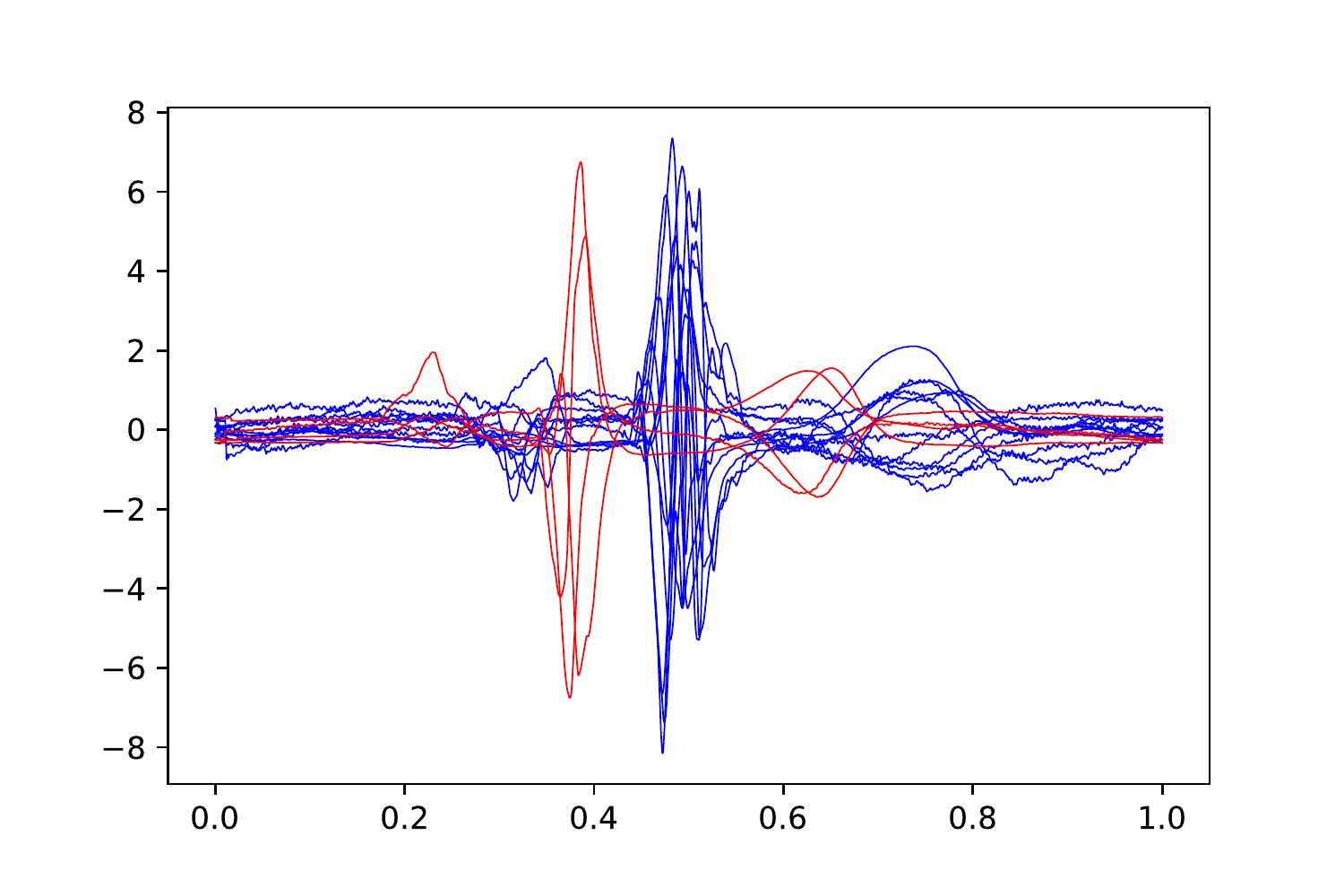}\,&\includegraphics[height=.1375\textheight, trim=1cm 0cm 1cm 1.15cm,clip=true]{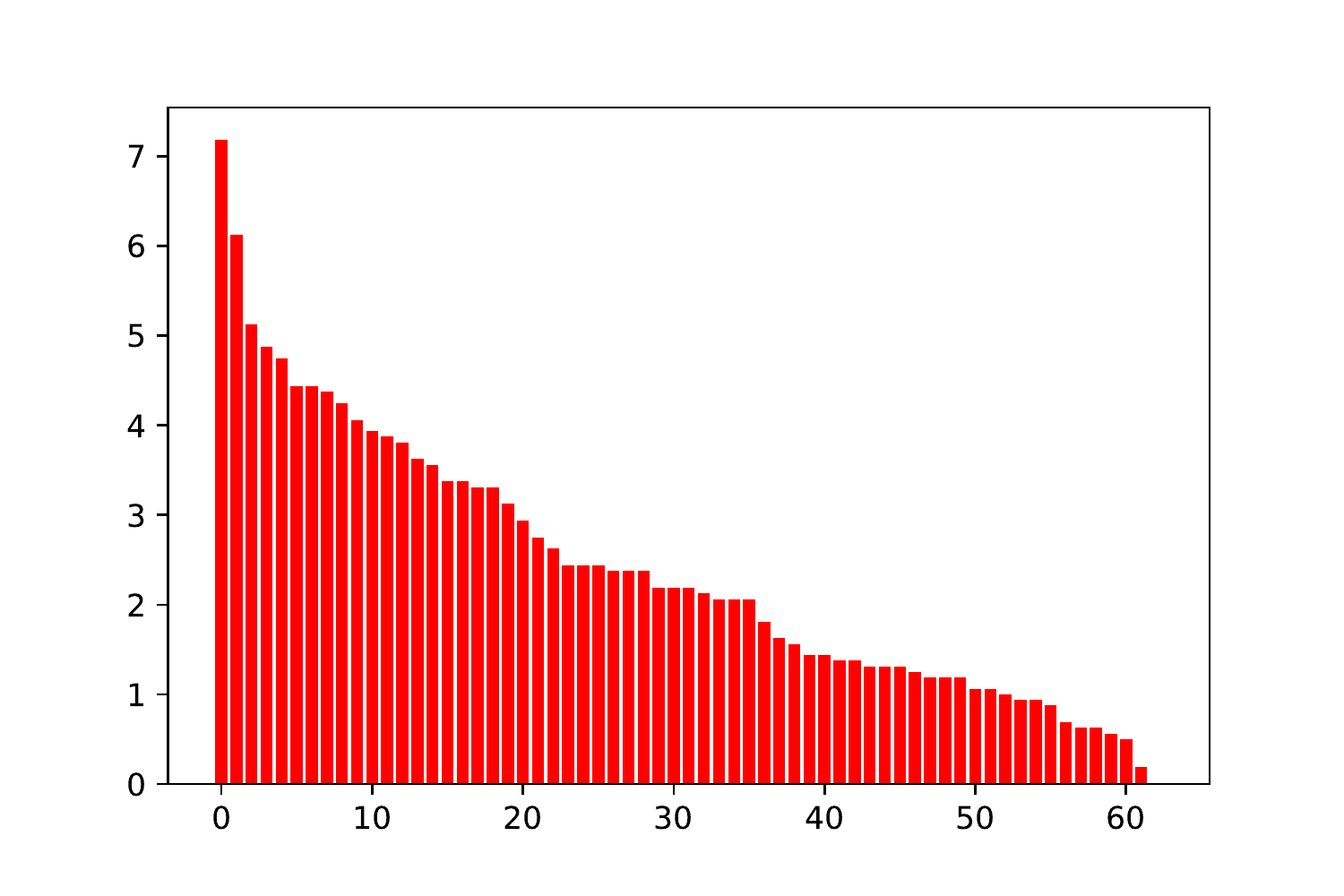}\,&\includegraphics[height=.1375\textheight, trim=1cm 0cm 1cm 1.15cm,clip=true]{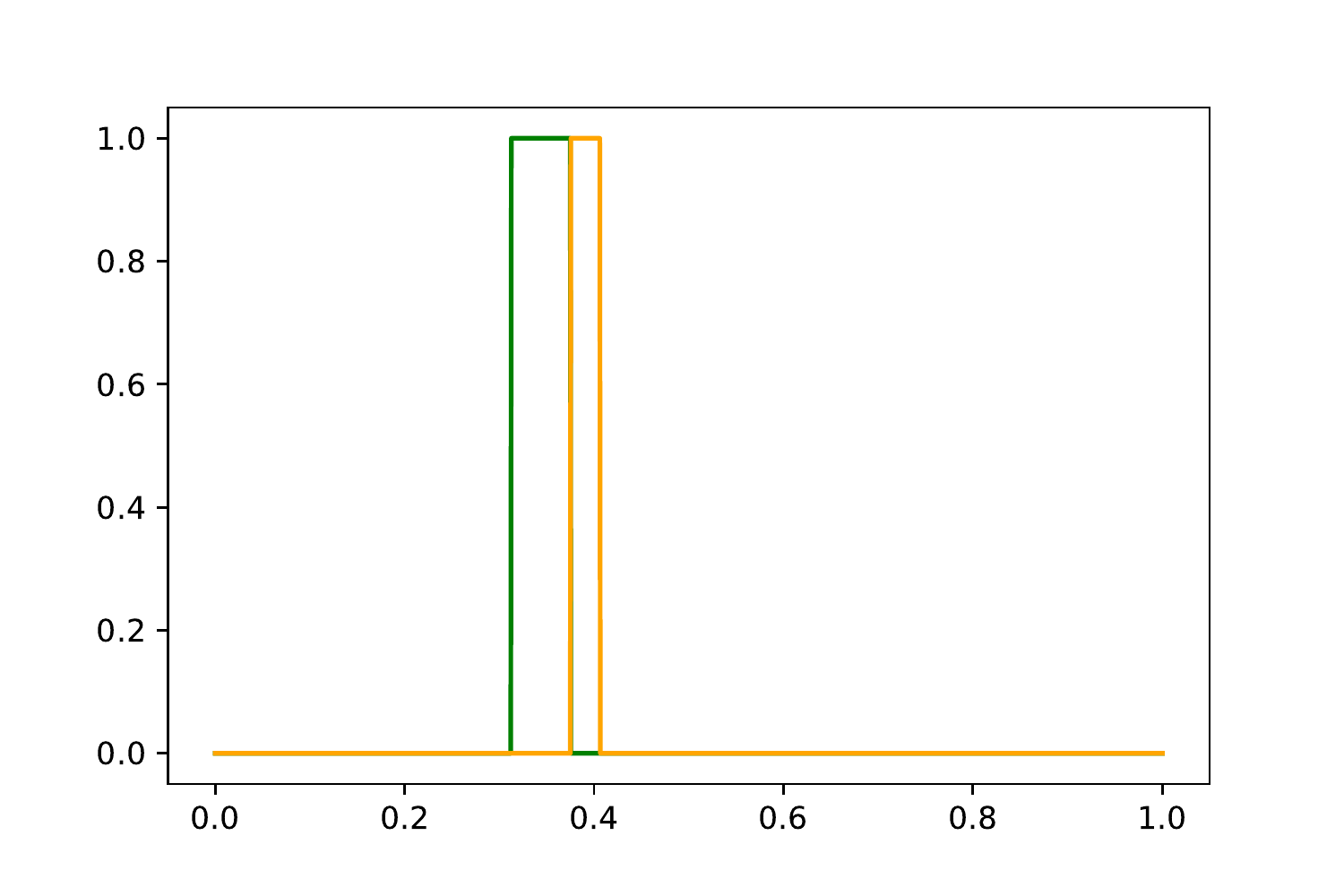}
\end{tabular}
\end{center}
\caption{CinECGTorso training dataset (a), red curves correspond to anomalies while blue curves to normal data. The direction importances given by the "adaptative" way are represented by (b) and the two most important functions (from the dyadic dictionary) used by FIF to build the model are plotted in (c). }\label{fig:feature_importance}
\end{figure}

\section{Study of the parameters of FIF}

In this section, we present results of a simulation study of the variance of the FIF algorithm. The experiments were conducted on the datasets (a) and (b) from Section~4 (see also Figure~3 of Section~4), for each of the four specified observations $\mathbf{x}_0$, $\mathbf{x}_1$, $\mathbf{x}_2$, $\mathbf{x}_3$ using the following settings (except varying parameter):
\begin{itemize}
\item[] \textit{Dictionary}: Gaussian wavelets (negative second derivative of the standard Gaussian density) with random variance selected in an uniform way in $[0.2,1]$ and a translation parameter selected randomly in $[-4,4]$. We fixed the size of the dictionary to 1000.
\item[]  \textit{Scalar product}: $L^{2}$ dot product.
\item[] \textit{Size of the dataset}: $n=500$.
\item[] \textit{Subsampling size}: $\psi =64$.
\item[] \textit{The number of trees}: $N = 100$.
\item[] \textit{The height limit}: fixed to $l=\lceil\log_2(\psi)\rceil$.
\end{itemize}
The figures below indicate boxplots of the FIF anomaly score, over $100$ runs. Empirical study of the FIF anomaly score and its variance when increasing the number of F-{\it i}trees is depicted in Figure~\ref{fig:treechange}.

\begin{figure}[h]
\begin{center}
\begin{tabular}{cc}
\hspace*{0.5cm}$\mathbf{x}_0$ &\hspace*{0.5cm} $\mathbf{x}_1$ \\
\includegraphics[width=.4\textwidth, trim=0cm 0cm 1cm 1.2cm,clip=true]{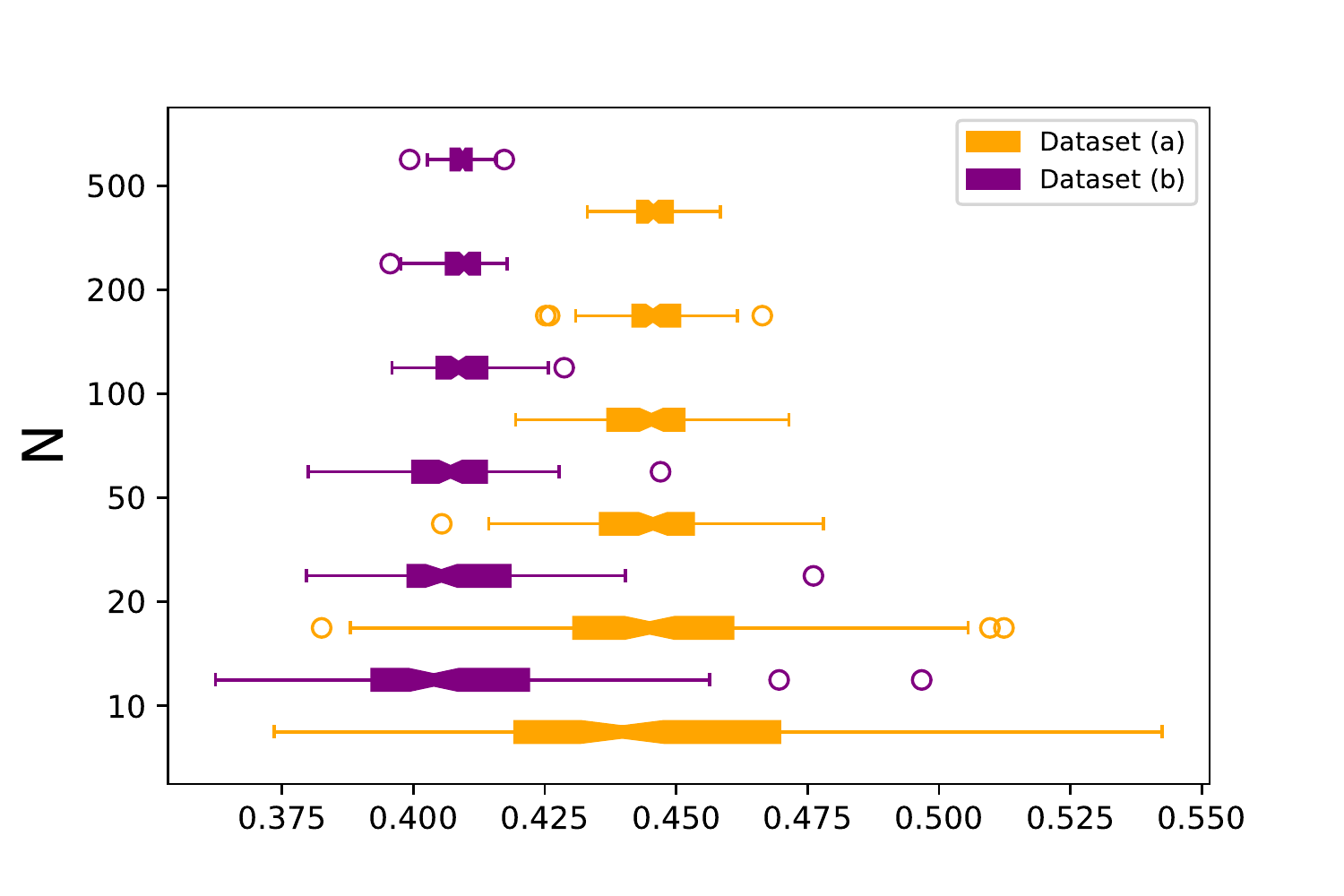} & \includegraphics[width=.4\textwidth, trim=0cm 0cm 1cm 1.2cm,clip=true]{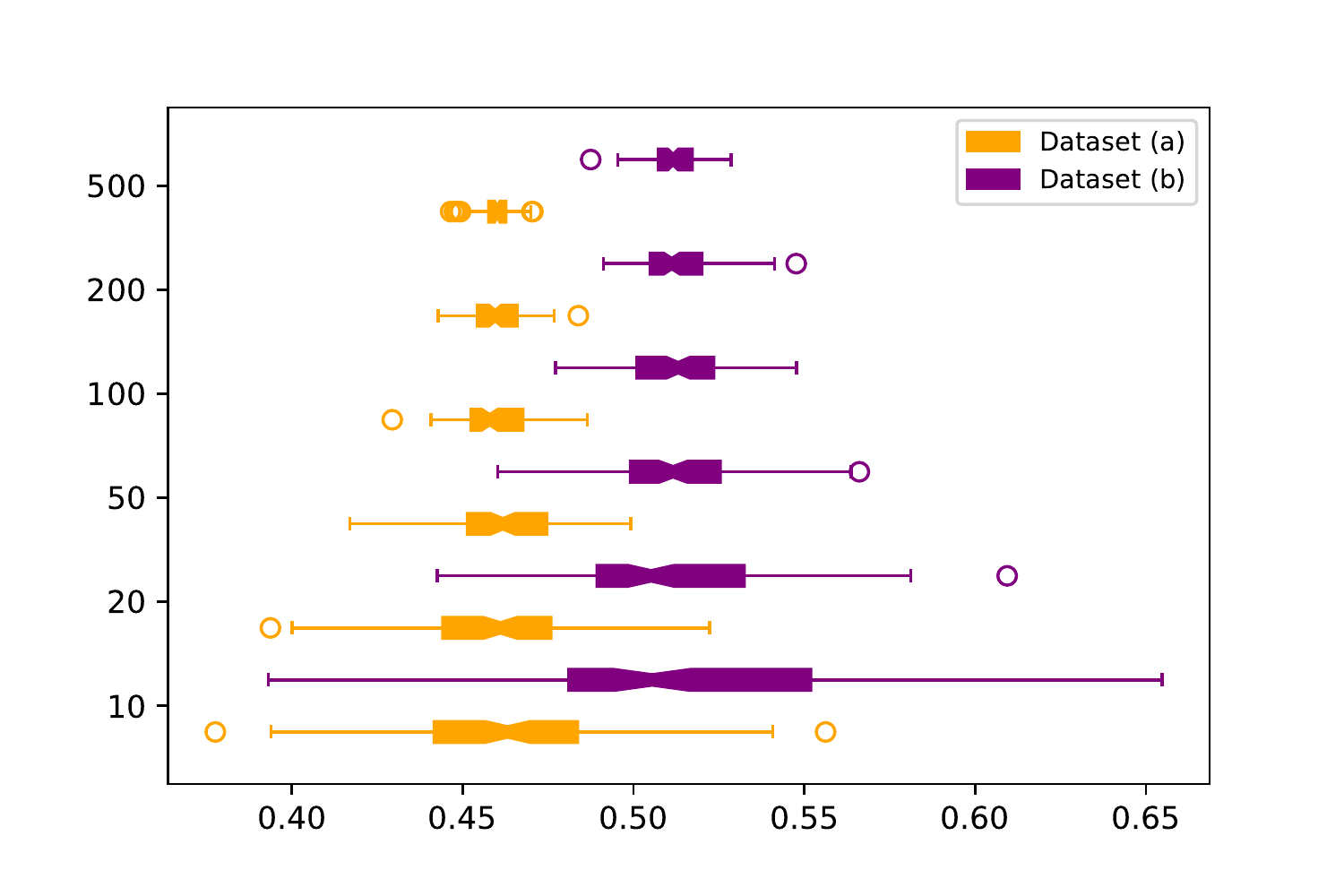} \\
\hspace*{0.5cm}$\mathbf{x}_2$ &\hspace*{0.5cm} $\mathbf{x}_3$ \\
  \includegraphics[width=.4\textwidth, trim=0cm 0cm 1cm 1.2cm,clip=true]{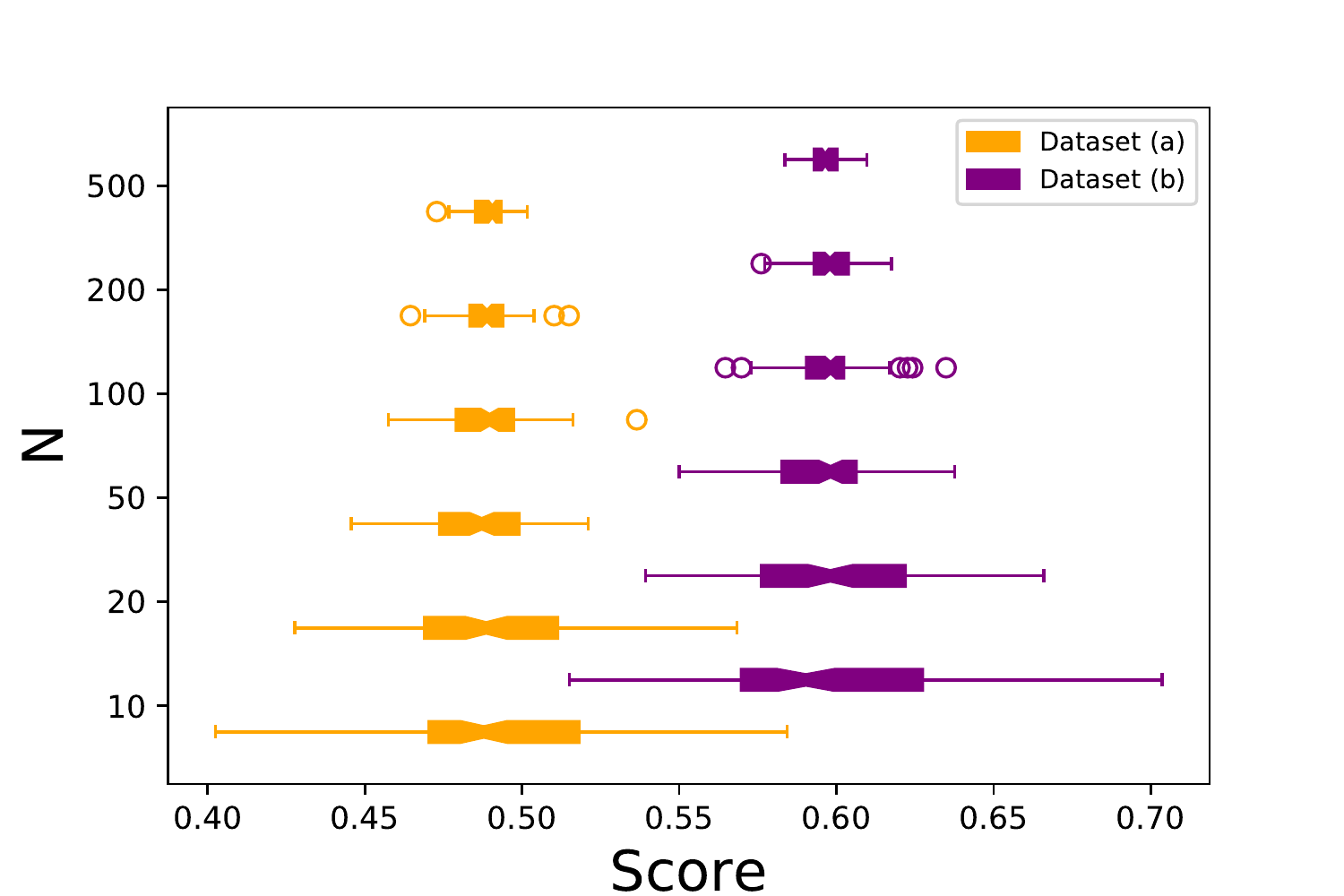}&\includegraphics[width=.4\textwidth, trim=0cm 0cm 1cm 1.2cm,clip=true]{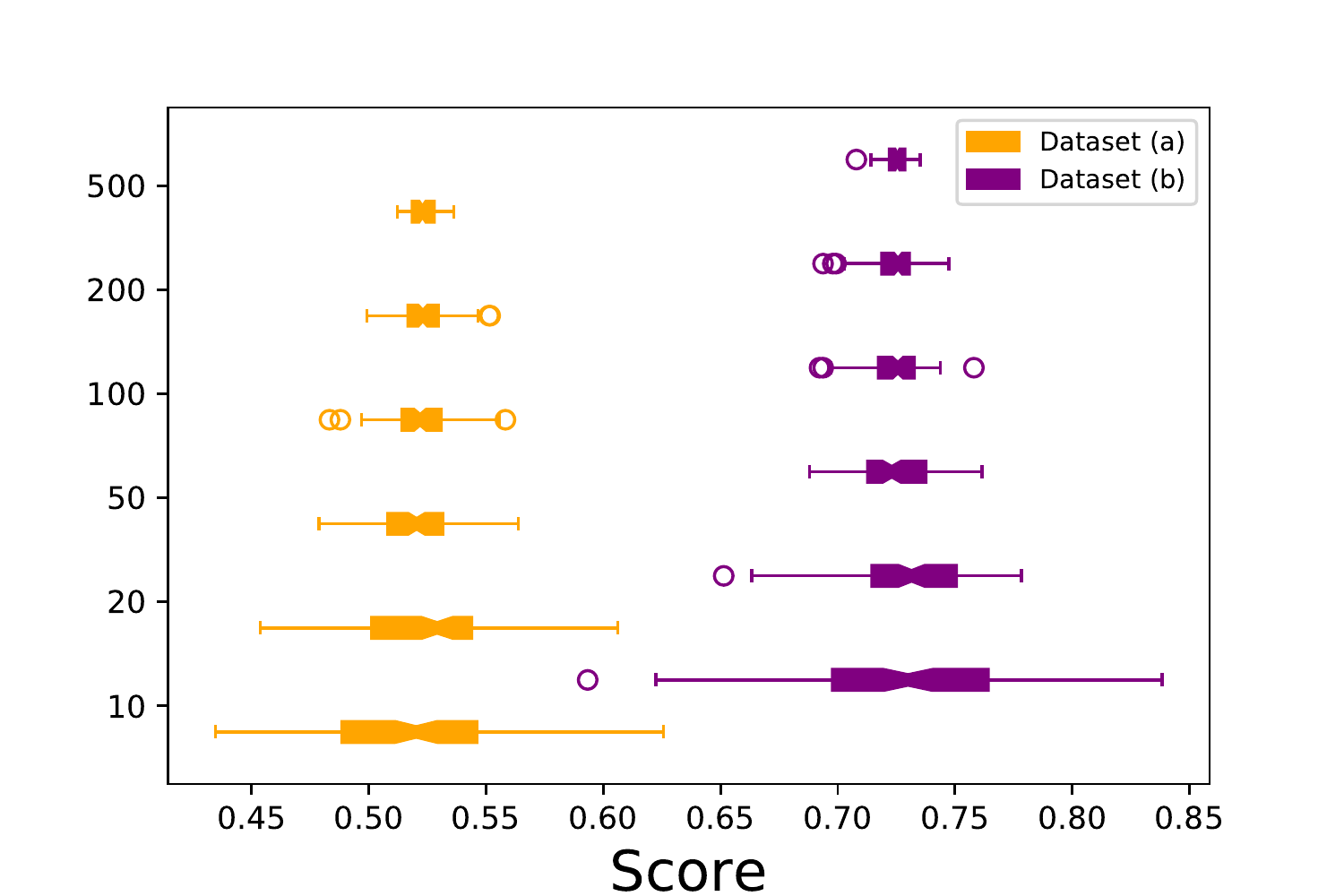}
\end{tabular}
\caption{Boxplot (over $100$ repetitions) of the FIF score for the observations $\mathbf{x}_0,\mathbf{x}_1,\mathbf{x}_2,\mathbf{x}_3$ for different sample sizes. The orange boxplots represent the dataset (a) while the purple boxplots represent the dataset (b).}
\label{fig:treechange}
\end{center}
\end{figure}

\clearpage

Empirical study of the FIF anomaly score and its variance when increasing the subsample size is depicted in Figure~\ref{fig:psichange}.

\begin{figure}[h]
\begin{center}
\begin{tabular}{cc}
\hspace*{0.5cm}$\mathbf{x}_0$ & \hspace*{0.5cm}$\mathbf{x}_1$ \\
\includegraphics[width=.4\textwidth, trim=0cm 0cm 1cm 1.2cm,clip=true]{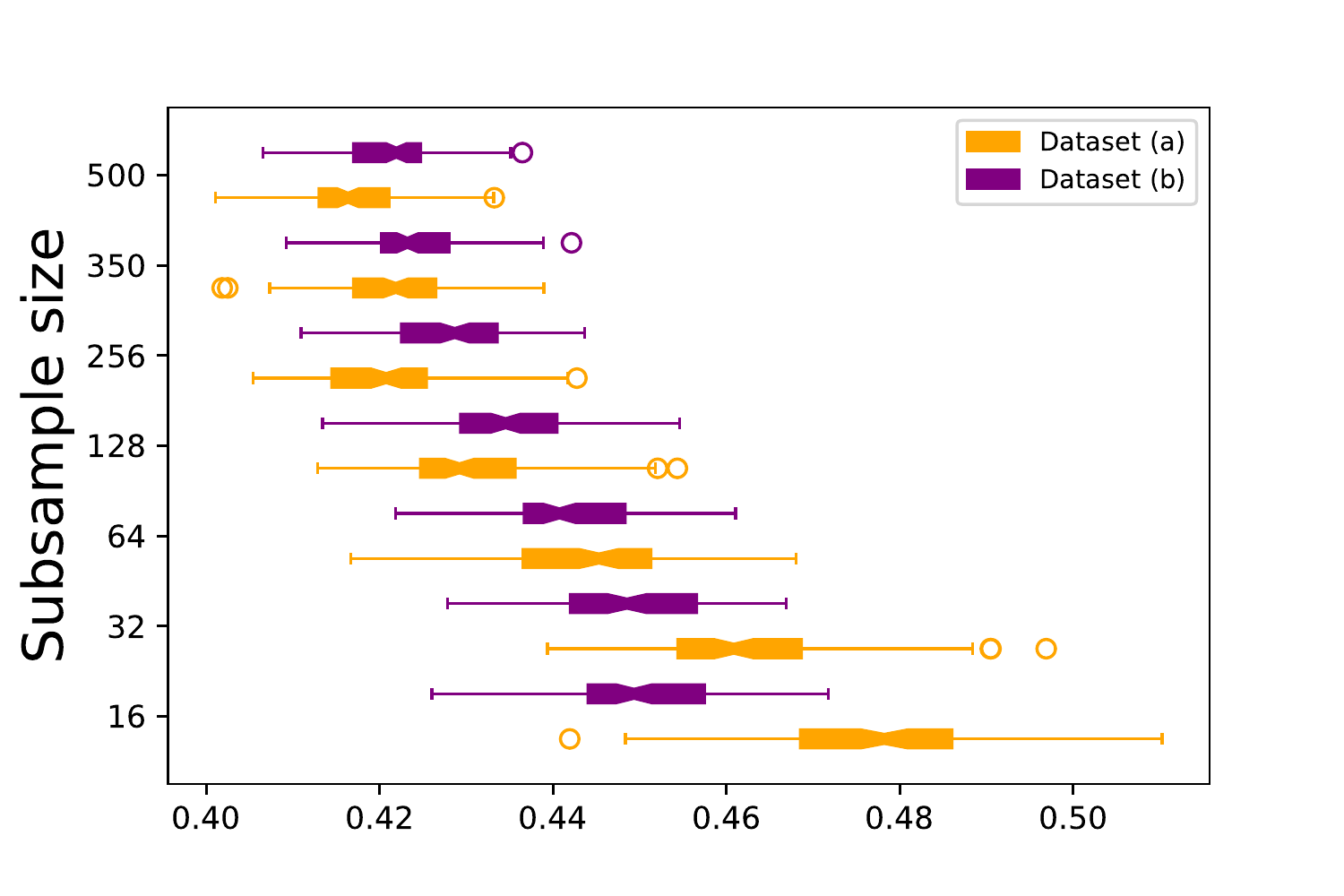} & \includegraphics[width=.4\textwidth, trim=0cm 0cm 1cm 1.2cm,clip=true]{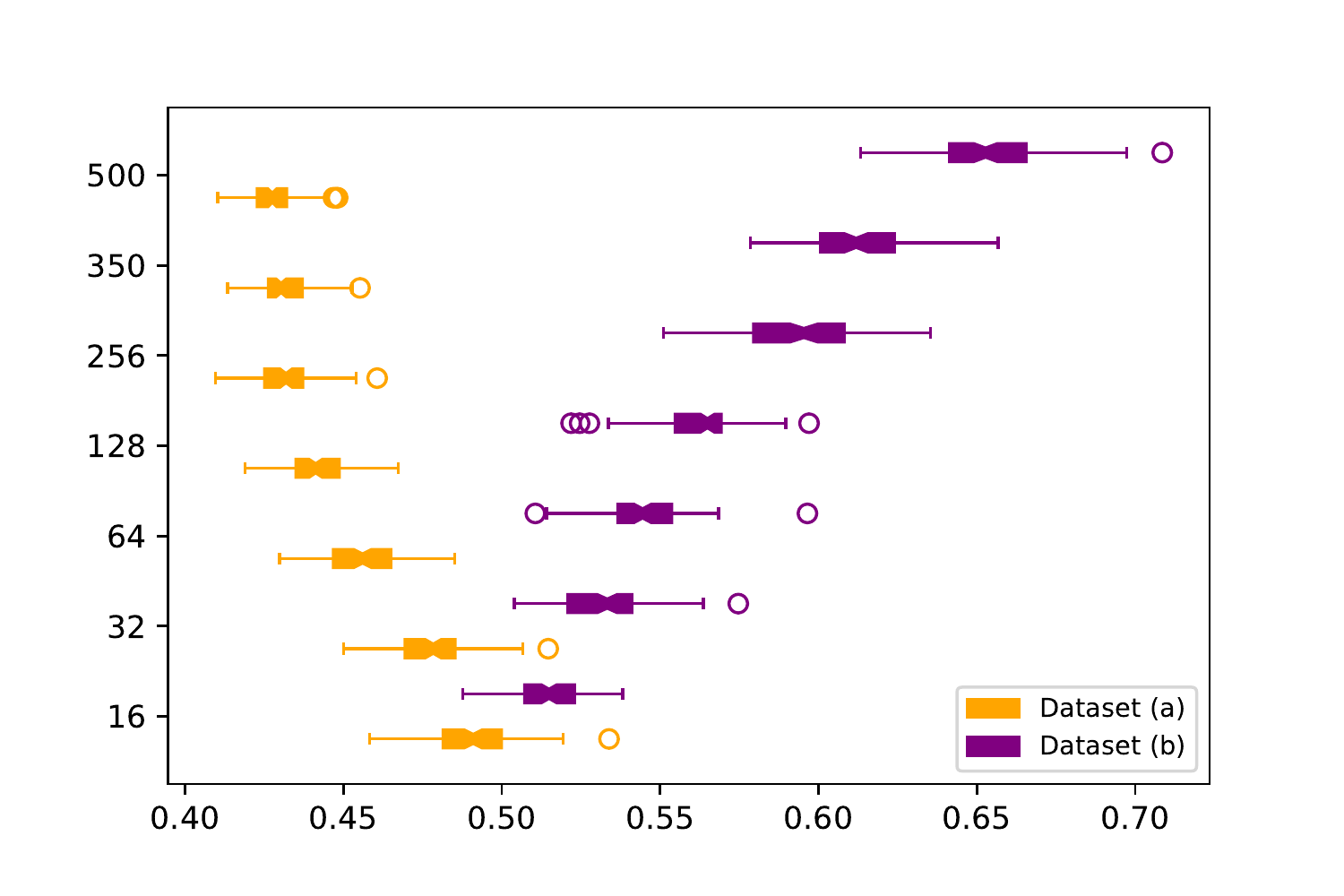} \\
\hspace*{0.5cm}$\mathbf{x}_2$ & \hspace*{0.5cm}$\mathbf{x}_3$ \\
  \includegraphics[width=.4\textwidth, trim=0cm 0cm 1cm 1.2cm,clip=true]{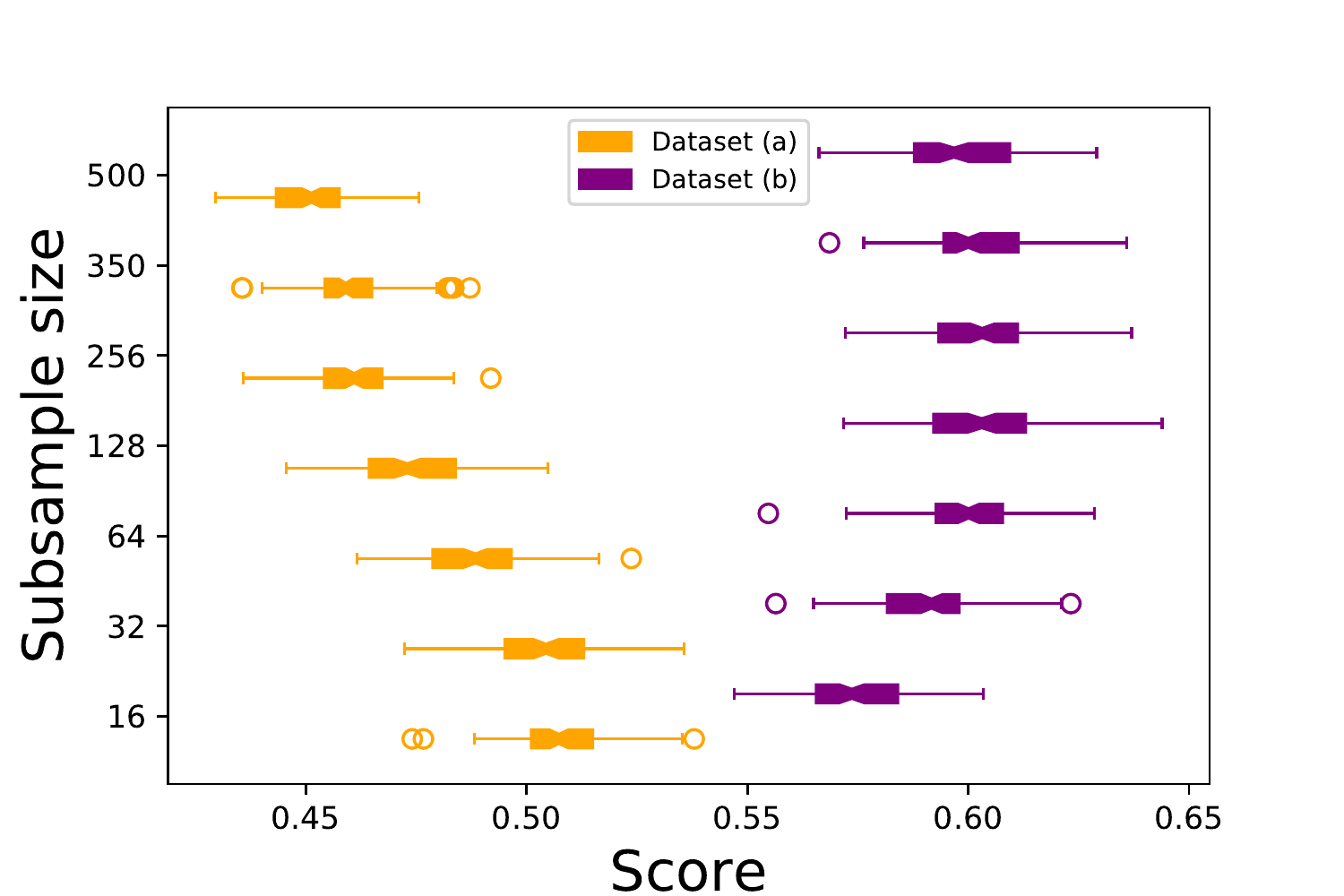}&\includegraphics[width=.4\textwidth, trim=0cm 0cm 1cm 1.2cm,clip=true]{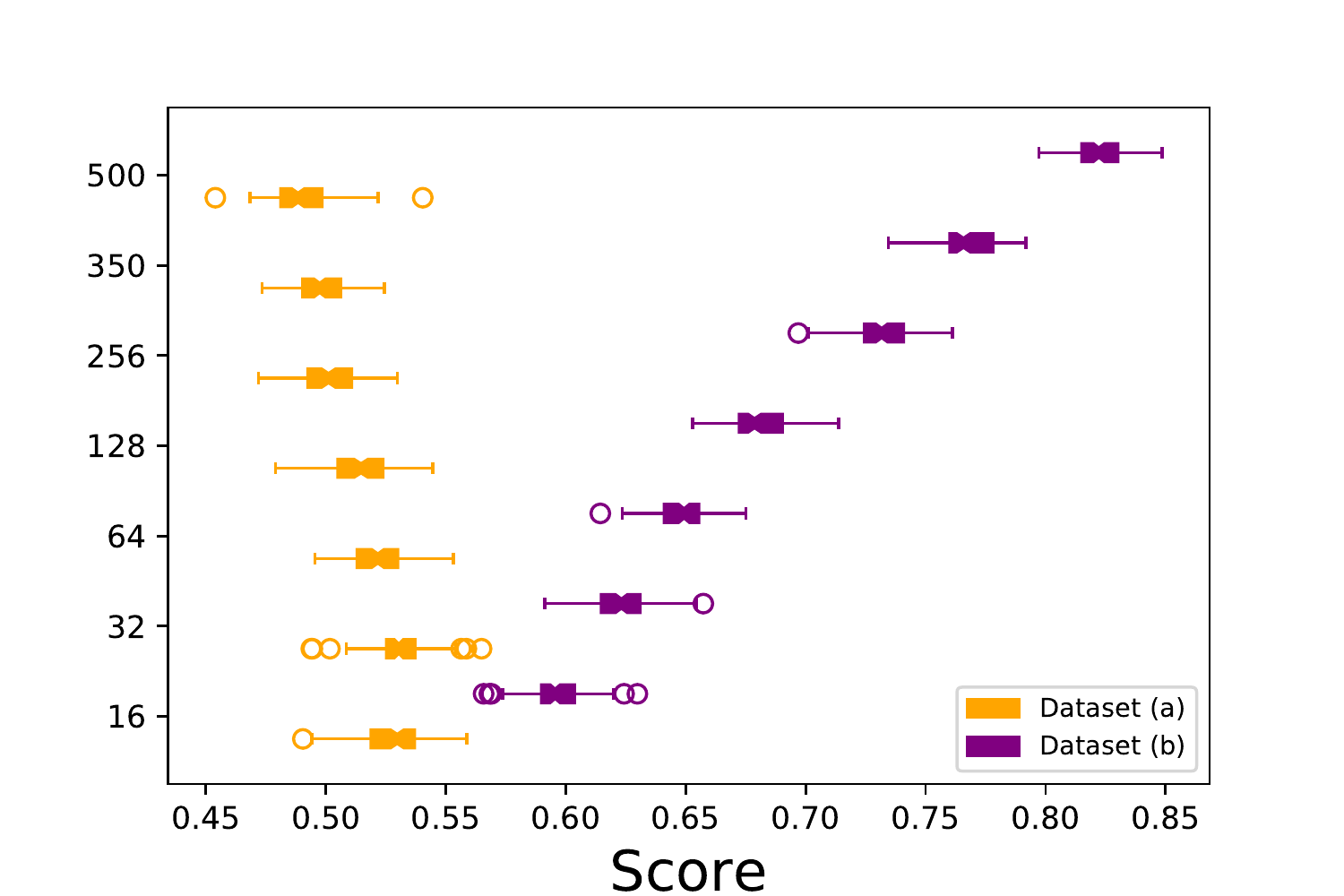}
\end{tabular}
\caption{Boxplot (over $100$ repetitions) of the FIF score for the observations $\mathbf{x}_0,\mathbf{x}_1,\mathbf{x}_2,\mathbf{x}_3$ for different sample sizes. The orange boxplots represent the dataset (a) while the purple boxplots represent the dataset (b).}
\label{fig:psichange}
\end{center}
\end{figure}

\clearpage

Empirical study of the FIF anomaly score and its variance when increasing the height limit of the F-{\it i}tree is depicted in Figure~\ref{fig:hlchange}.

\begin{figure}[h]
\begin{center}
\begin{tabular}{cc}
\hspace*{0.5cm}$\mathbf{x}_0$ &\hspace*{0.5cm} $\mathbf{x}_1$ \\
\includegraphics[width=.4\textwidth, trim=0cm 0cm 1cm 1.2cm,clip=true]{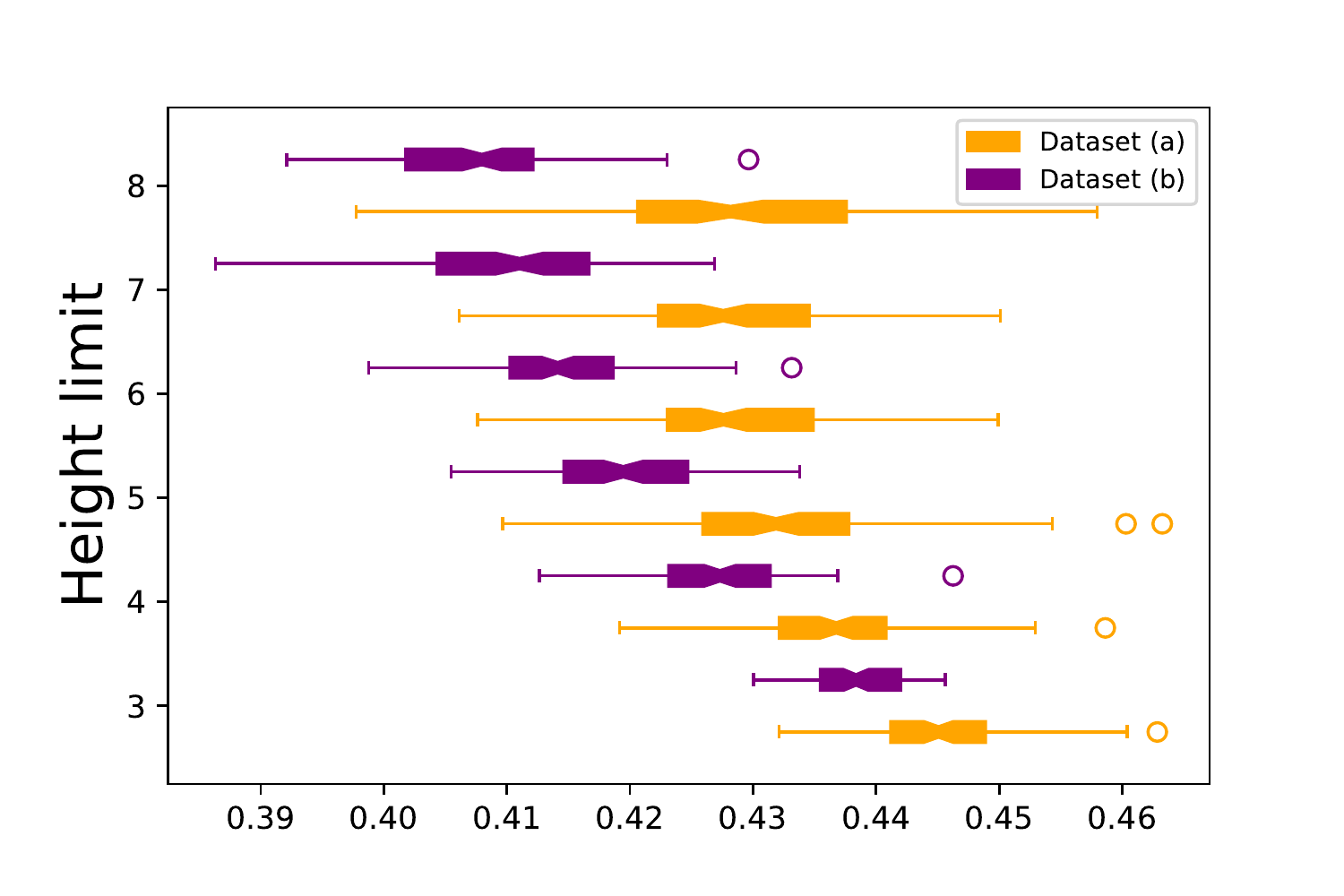} & \includegraphics[width=.4\textwidth, trim=0cm 0cm 1cm 1.2cm,clip=true]{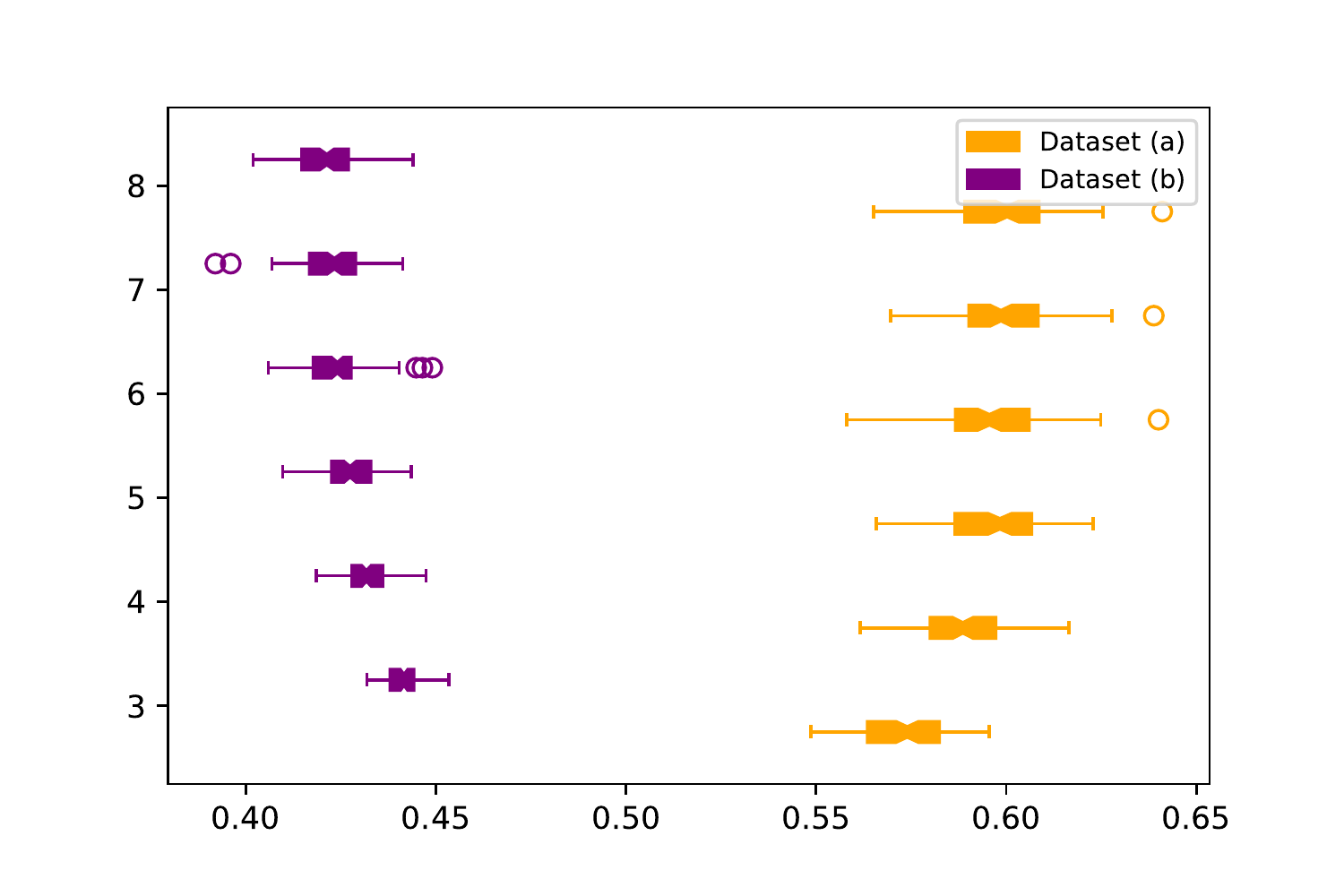} \\
\hspace*{0.5cm}$\mathbf{x}_2$ &\hspace*{0.5cm} $\mathbf{x}_3$ \\
  \includegraphics[width=.4\textwidth, trim=0cm 0cm 1cm 1.2cm,clip=true]{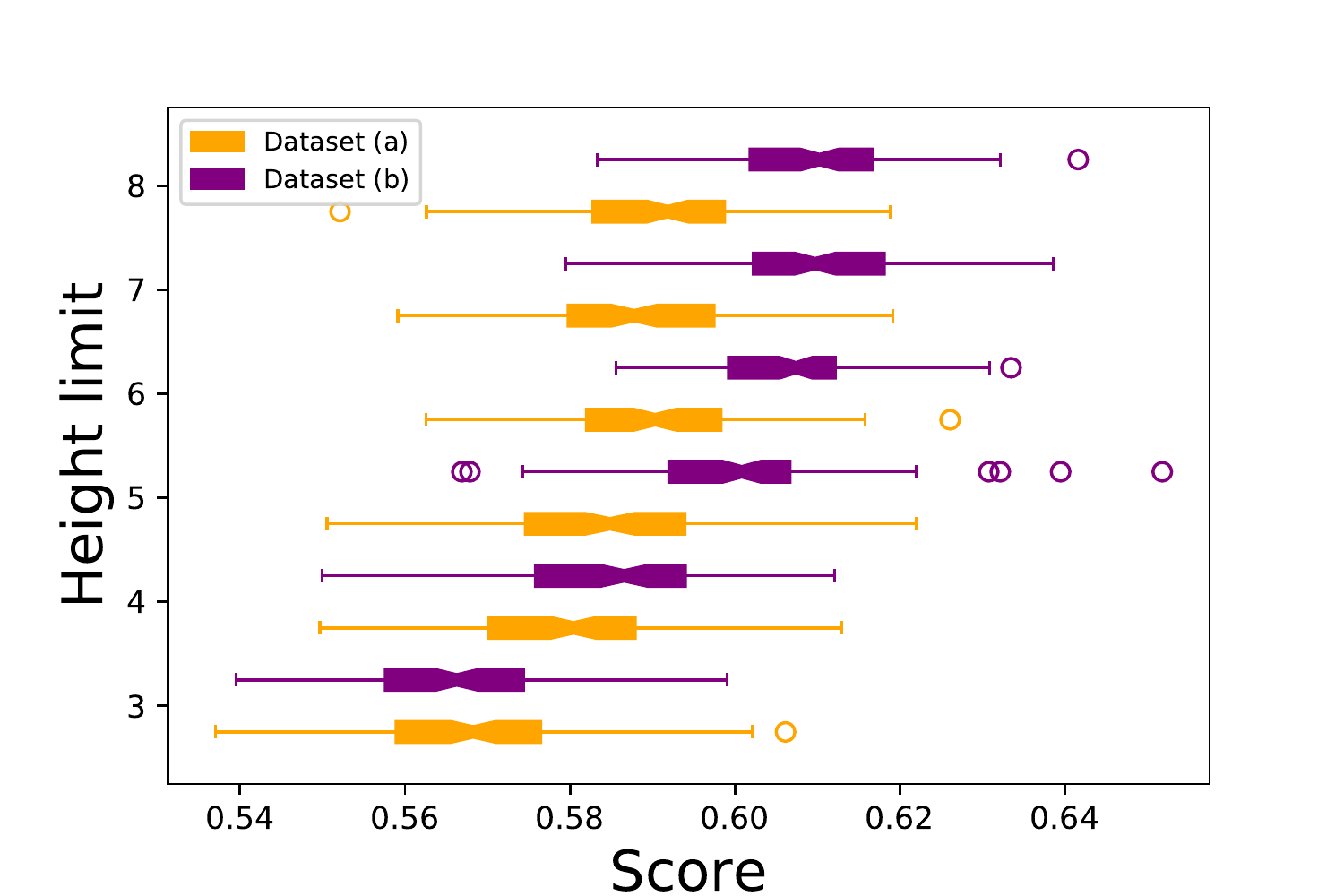}&\includegraphics[width=.4\textwidth, trim=0cm 0cm 1cm 1.2cm,clip=true]{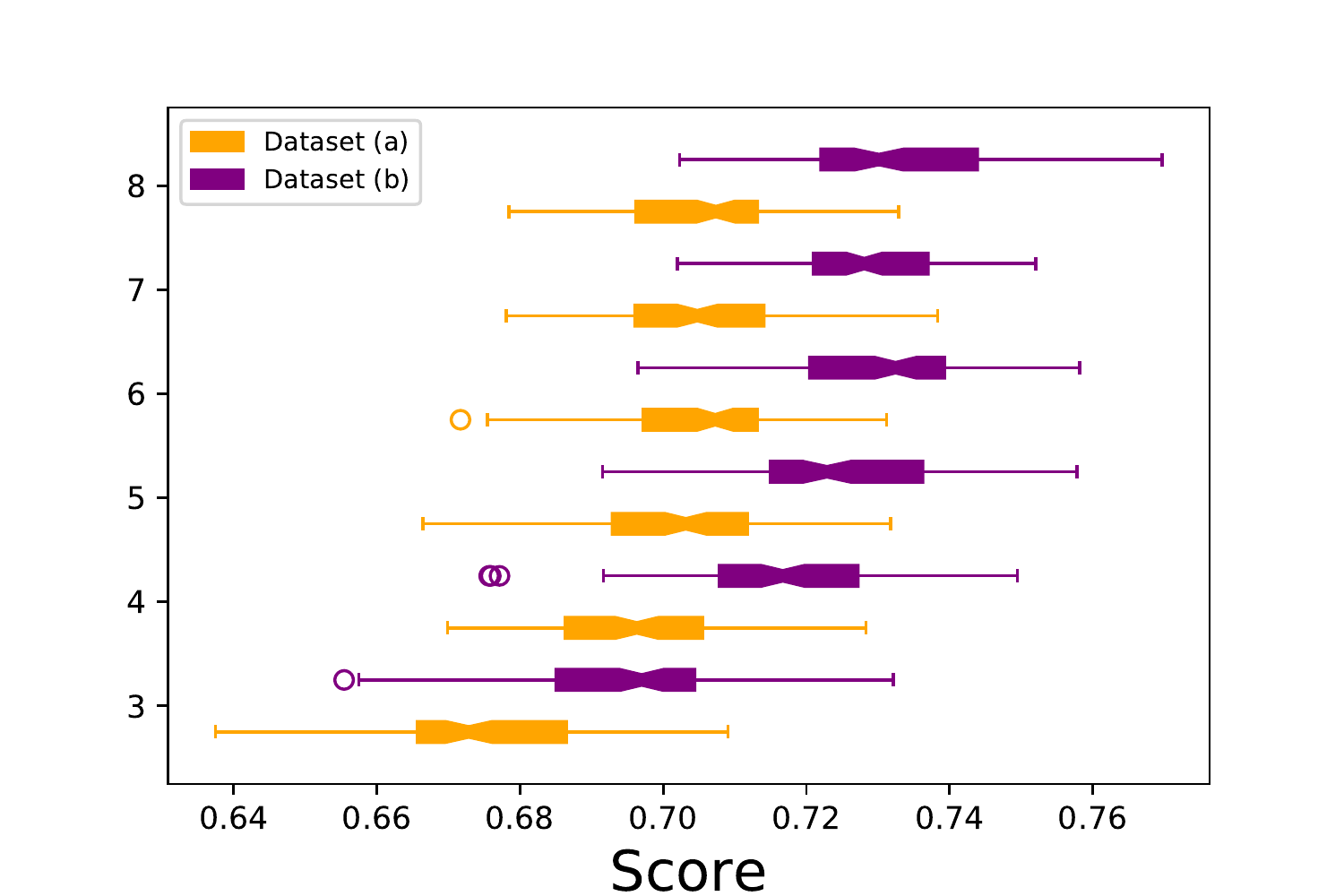}
\end{tabular}
\caption{Boxplot (over $100$ repetitions) of the FIF score for the observations $\mathbf{x}_0,\mathbf{x}_1,\mathbf{x}_2,\mathbf{x}_3$ for different sample sizes. The orange boxplots represent the dataset (a) while the purple boxplots represent the dataset (b).}
\label{fig:hlchange}
\end{center}
\end{figure}

\clearpage

Taking  finite size versions of the infinite \textit{gaussian wavelets} dictionary, an empirical study of the FIF anomaly score and its variance when increasing the size of the dictionary is depicted in Figure~\ref{fig:dsizechange}.

\begin{figure}[h]
\begin{center}
\begin{tabular}{cc}
\hspace*{0.5cm}$\mathbf{x}_0$ & \hspace*{0.5cm}$\mathbf{x}_1$ \\
\includegraphics[width=.4\textwidth, trim=0cm 0cm 1cm 1.2cm,clip=true]{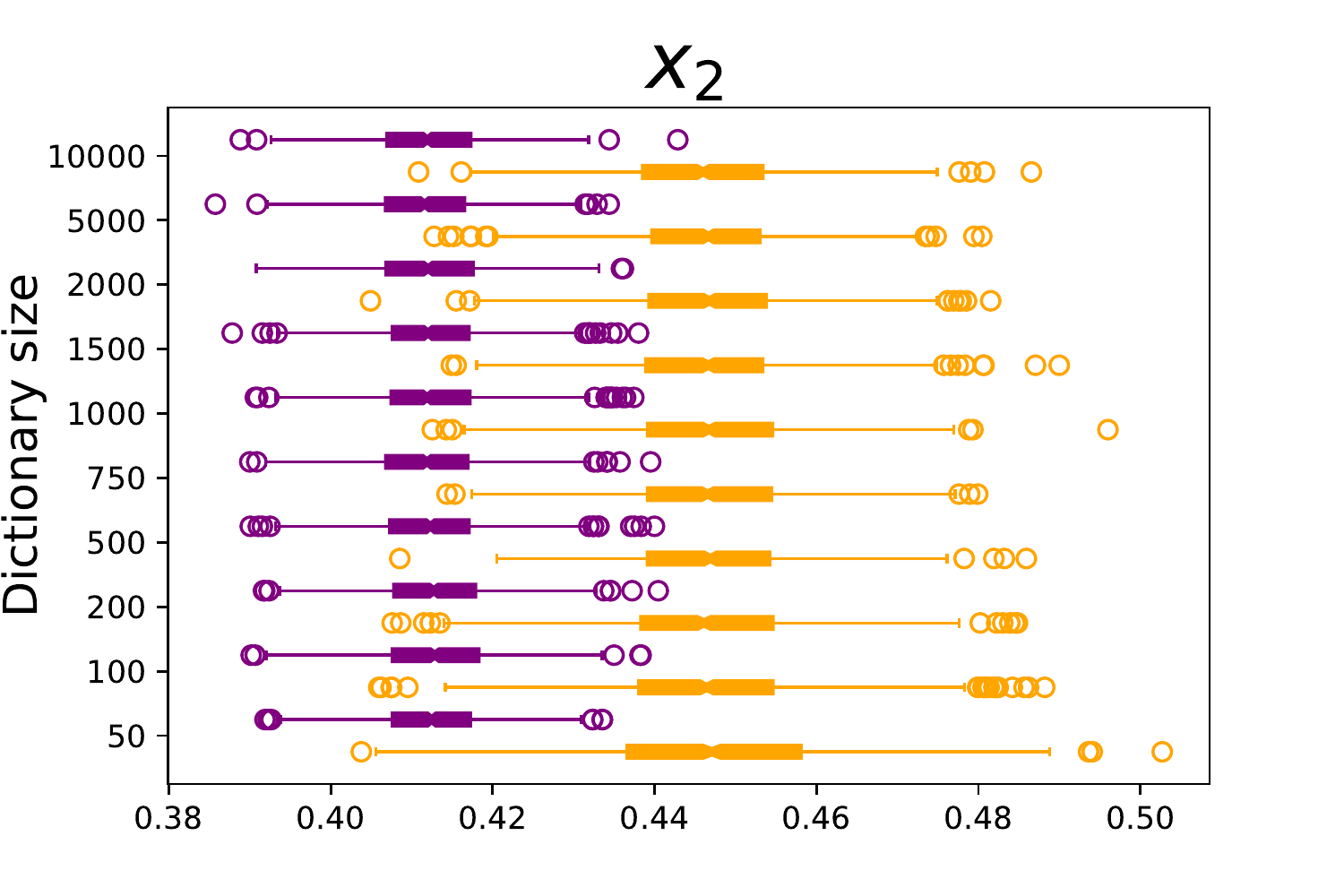} & \includegraphics[width=.4\textwidth, trim=0cm 0cm 1cm 1.2cm,clip=true]{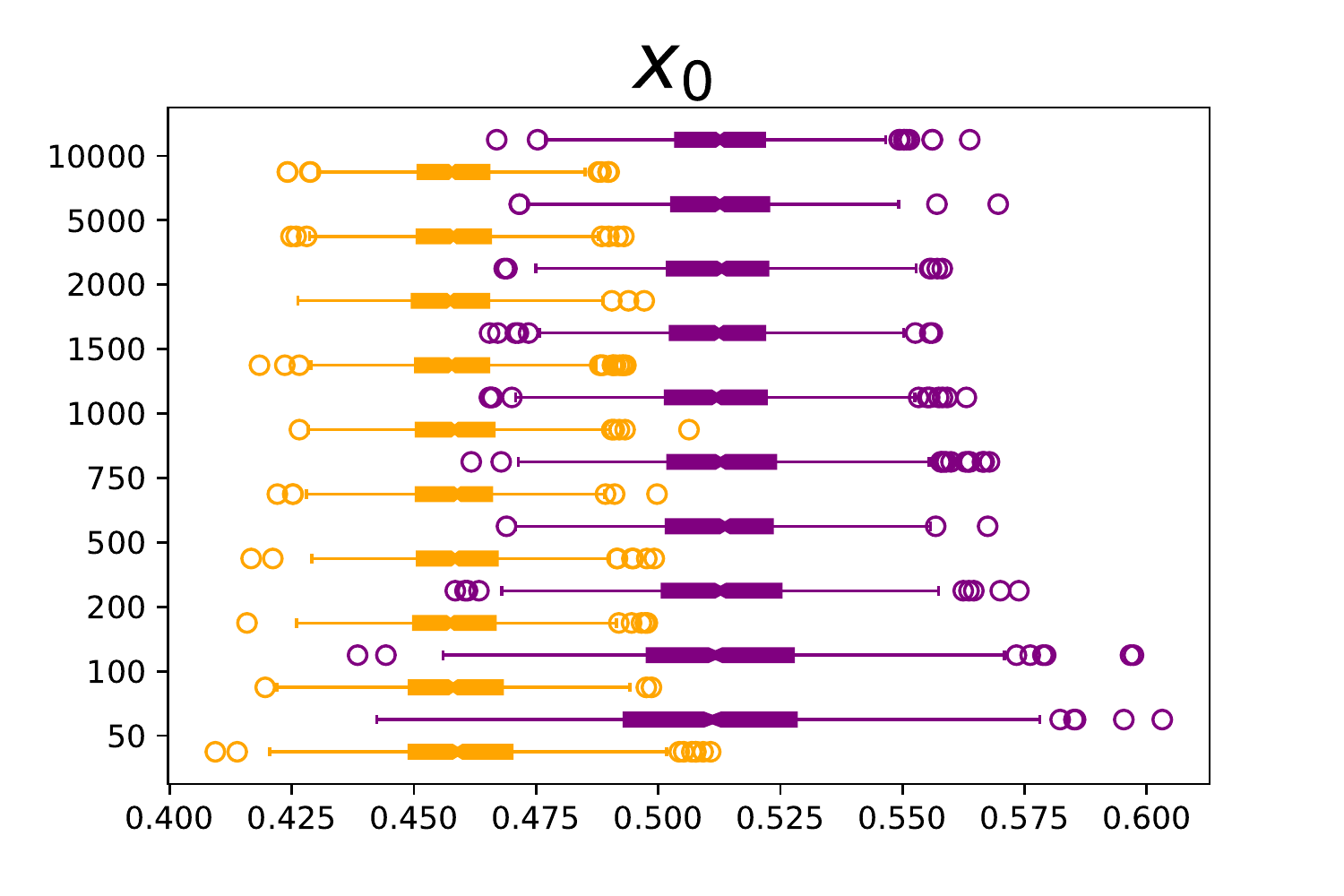} \\
\hspace*{0.5cm}$\mathbf{x}_2$ & \hspace*{0.5cm}$\mathbf{x}_3$ \\
  \includegraphics[width=.4\textwidth, trim=0cm 0cm 1cm 1.2cm,clip=true]{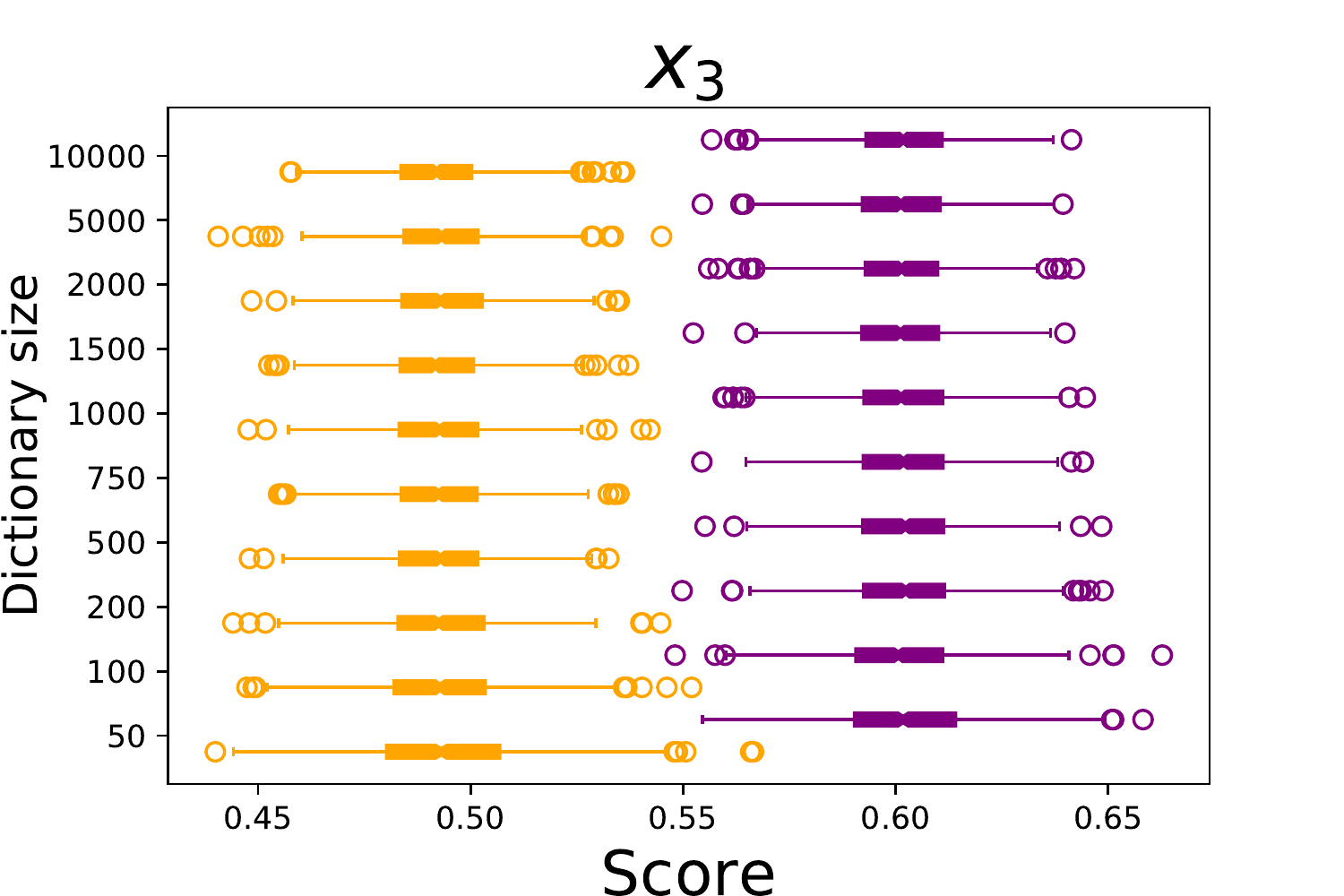}&\includegraphics[width=.4\textwidth, trim=0cm 0cm 1cm 1.2cm,clip=true]{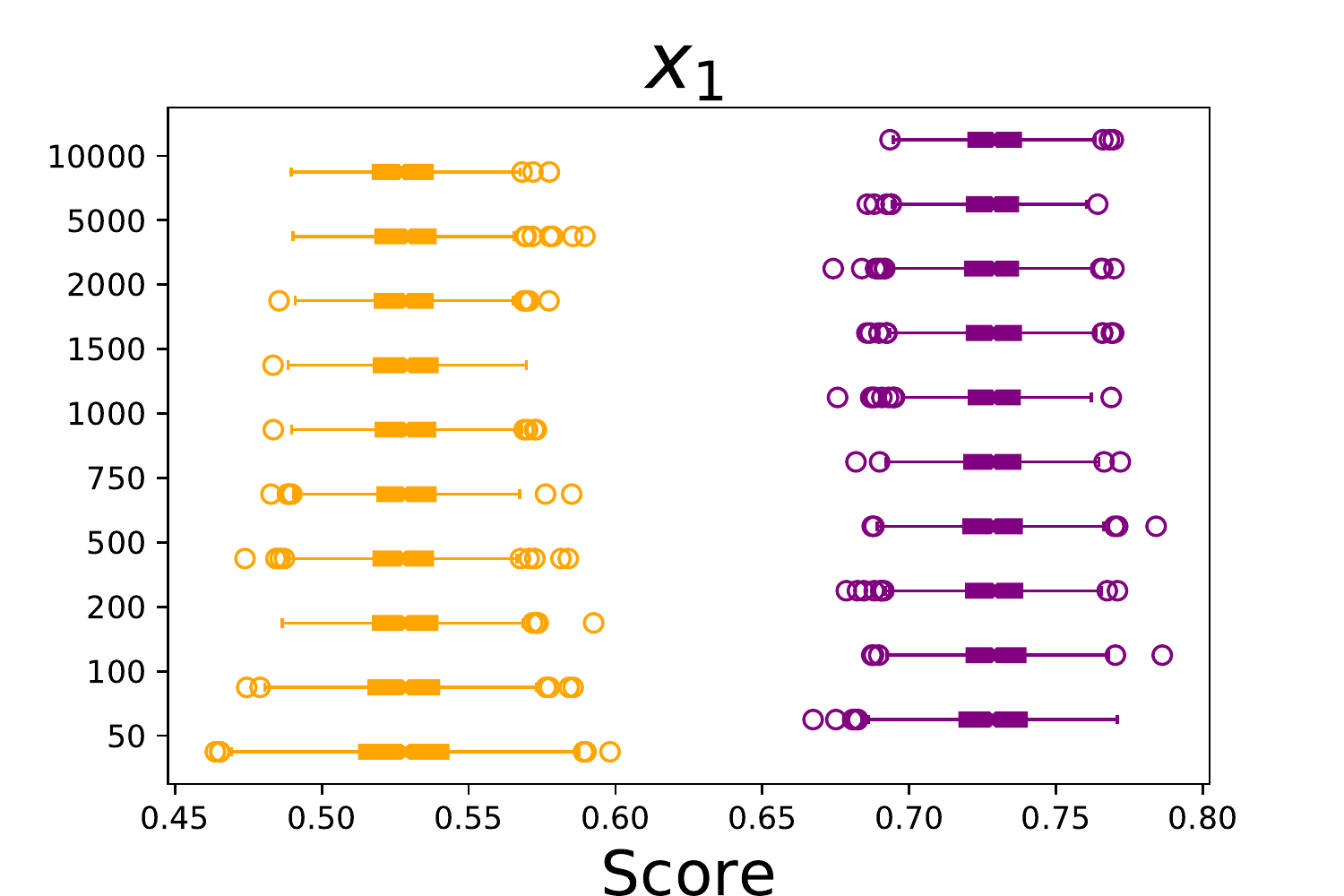}
\end{tabular}
\caption{Boxplot (over $100$ repetitions) of the FIF score for the observations $\mathbf{x}_0,\mathbf{x}_1,\mathbf{x}_2,\mathbf{x}_3$ for different sample sizes. The orange boxplots represent the dataset (a) while the purple boxplots represent the dataset (b).}
\label{fig:dsizechange}
\end{center}
\end{figure}

\clearpage

Empirical study of the FIF anomaly score for a variety of dictionaries with the $L_{2}$ scalar product of the derivatives is depicted in Figure~\ref{fig:dictsdotder}.

\begin{figure}[h]
\begin{center}
\begin{tabular}{cc}
\hspace*{0.5cm}$\mathbf{x}_0$ &\hspace*{0.5cm} $\mathbf{x}_1$ \\
\includegraphics[width=.4\textwidth, trim=0cm 0cm 1cm 1.2cm,clip=true]{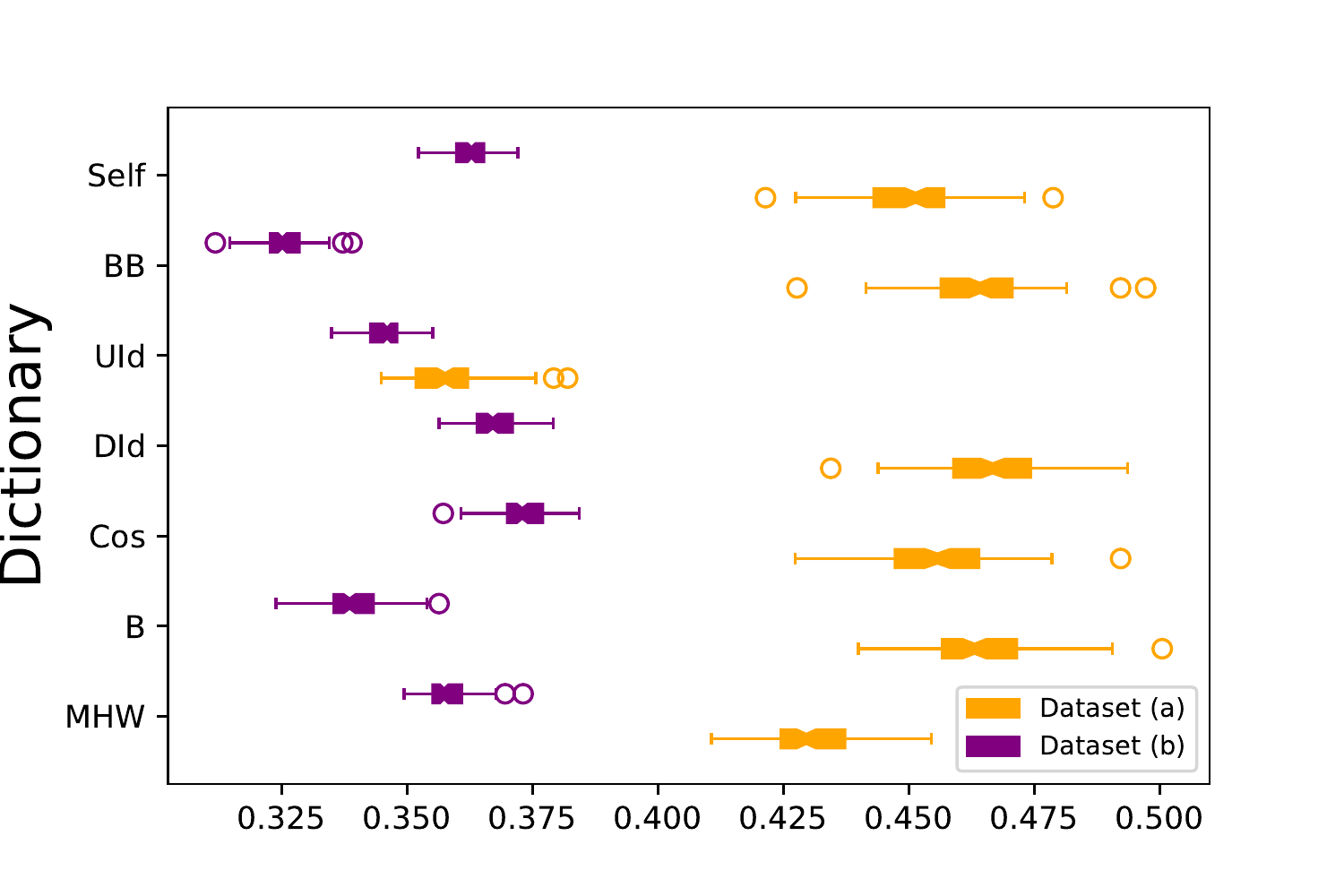} & \includegraphics[width=.4\textwidth, trim=0cm 0cm 1cm 1.2cm,clip=true]{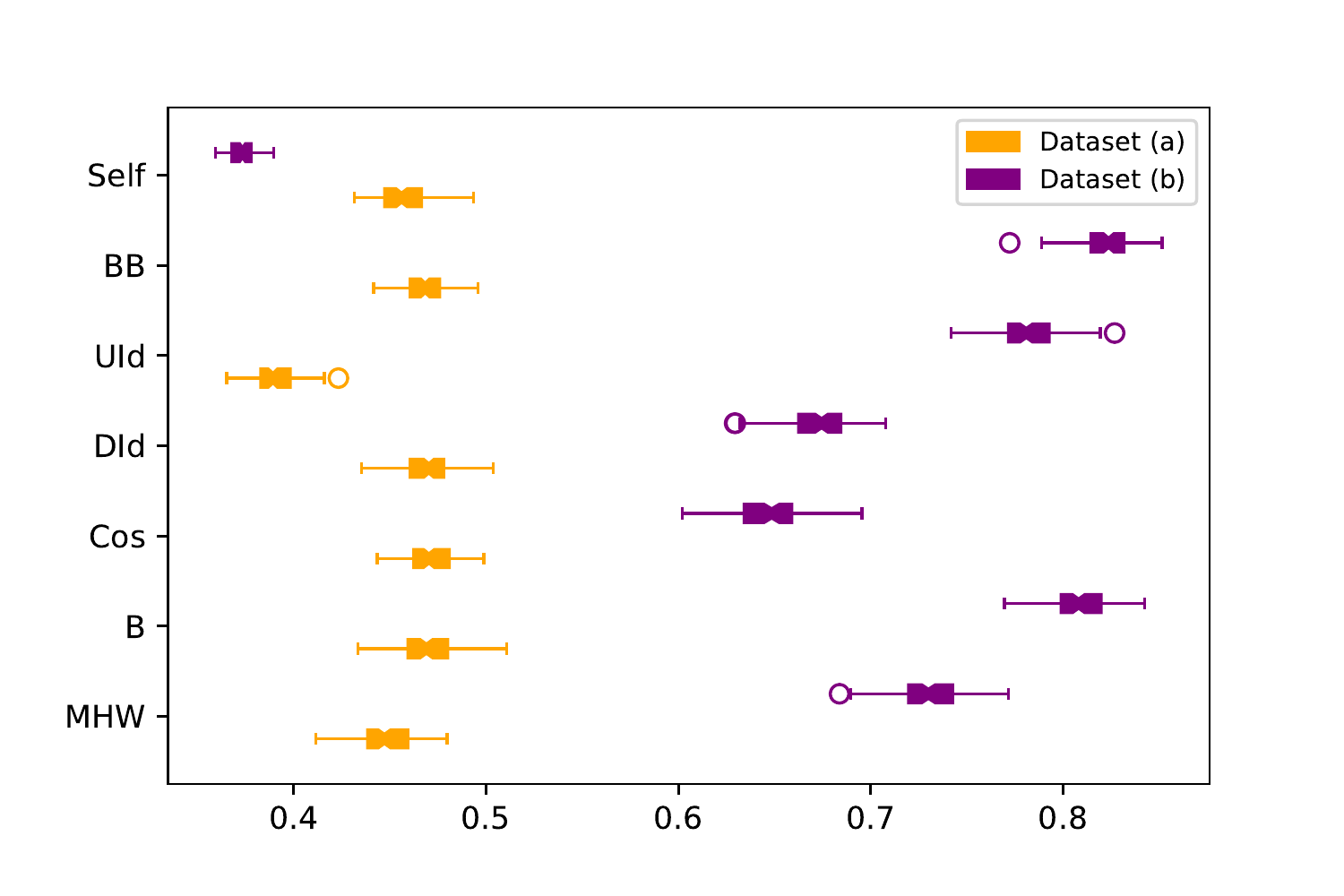} \\
\hspace*{0.5cm}$\mathbf{x}_2$ &\hspace*{0.5cm} $\mathbf{x}_3$ \\
\includegraphics[width=.4\textwidth, trim=0cm 0cm 1cm 1.2cm,clip=true]{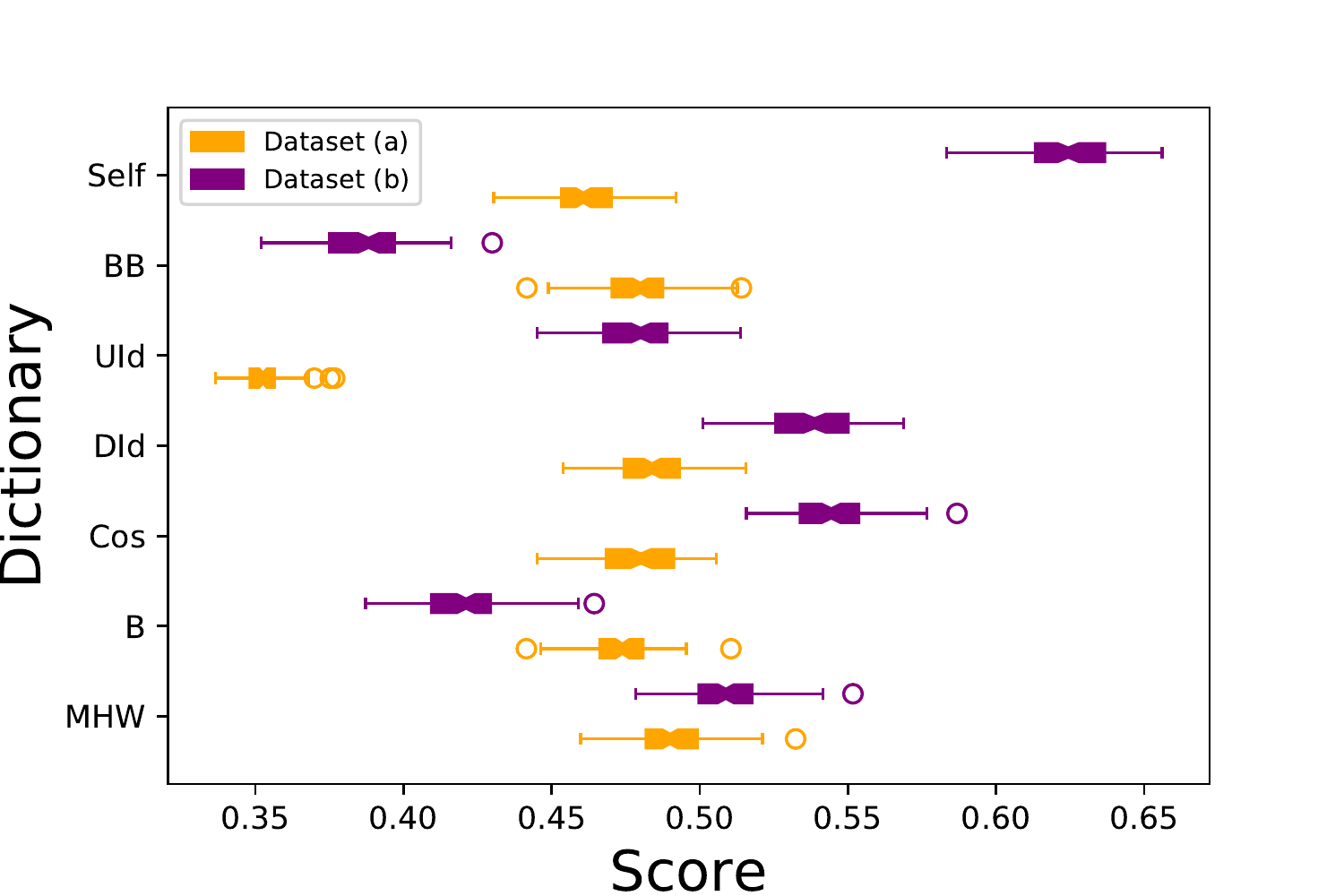} & \includegraphics[width=.4\textwidth, trim=0cm 0cm 1cm 1.2cm,clip=true]{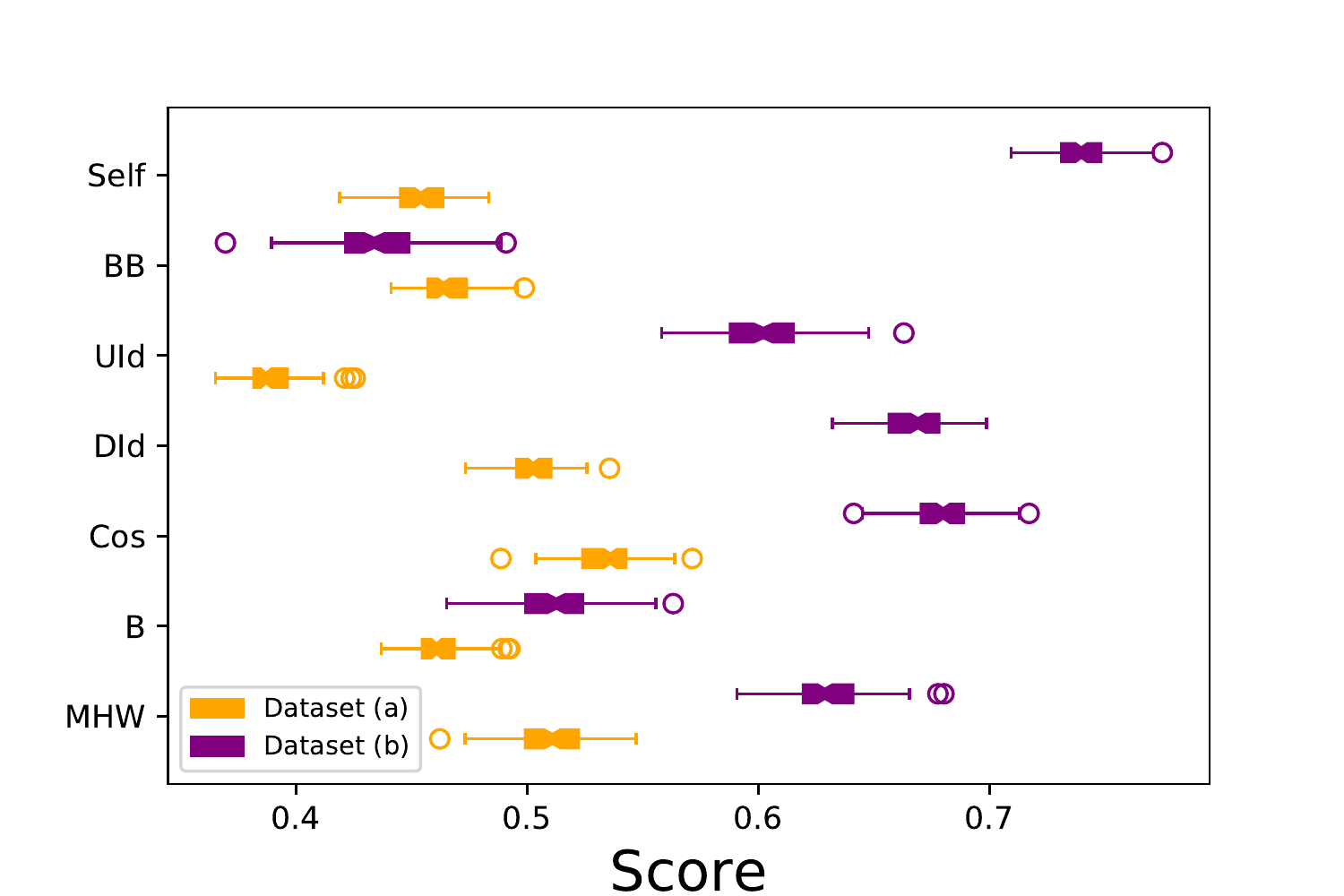}
\end{tabular}
\caption{Boxplot (over $100$ repetitions) of the FIF score for the observations $\mathbf{x}_0,\mathbf{x}_1,\mathbf{x}_2,\mathbf{x}_3$ for different dictionaries using the $L_{2}$ scalar product of the derivatives. The orange boxplots represent the dataset (a) while the purple boxplots represent the dataset (b).}
\label{fig:dictsdotder}
\end{center}
\end{figure}

\noindent{\bf Analysis of the results of Section C} In a first experiment, we show the boxplots of the score estimated by FIF when increasing the number of F-{\it i}trees and observe that, as expected, the variance diminishes when $N$ grows (see Figure ~\ref{fig:treechange}). We also see in  Figure ~\ref{fig:psichange} that with an increasing subsample size $\psi$ the FIF anomaly score increases for anomalies since these are more often present in the subsample and thus isolated faster (with shorter path length) when calculating the score than when they were absent in the subsample; this effect is reciprocal for normal observations. A similar behavior is observed with increasing height limit $l$ in Figure ~\ref{fig:hlchange}. The variance of the score tends to slightly increase with $\psi$ and $l$ because of more observations/branching possibilities. If the dictionary is sufficiently rich, its size does not influence the FIF anomaly score and its variance stabilizes relatively fast while growing the size of the dictionary (see Figure ~\ref{fig:dsizechange}) which encourages the use of massive (and infinite size) dictionaries.

\section{Complementary results on the performance comparison}

\subsection{benchmark datasets}

Here, we plot the thirteen benchmark train datasets used in the experiment. Anomalies are represented by blue color while normal data are drawn in red.

\vspace*{0.3cm}

\begin{tabular}{cccc}
{\tiny Chinatown}&{\tiny Coffee}&{\tiny ECGFiveDays}&{\tiny ECG200}\\
\includegraphics[width=.2\textwidth, trim=0cm 0cm 1cm 1.2cm,clip=true]{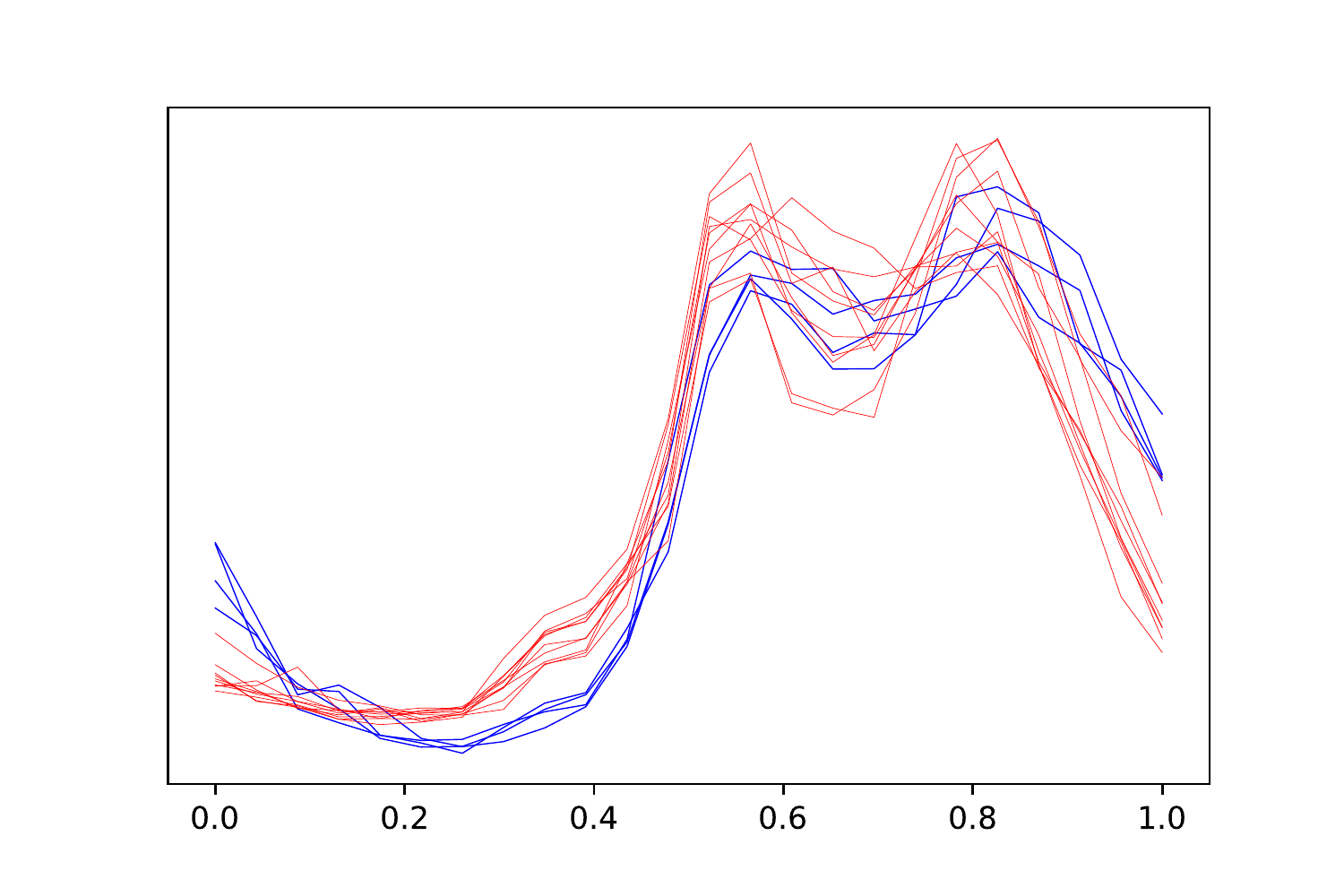}&\includegraphics[width=.2\textwidth, trim=0cm 0cm 1cm 1.2cm,clip=true]{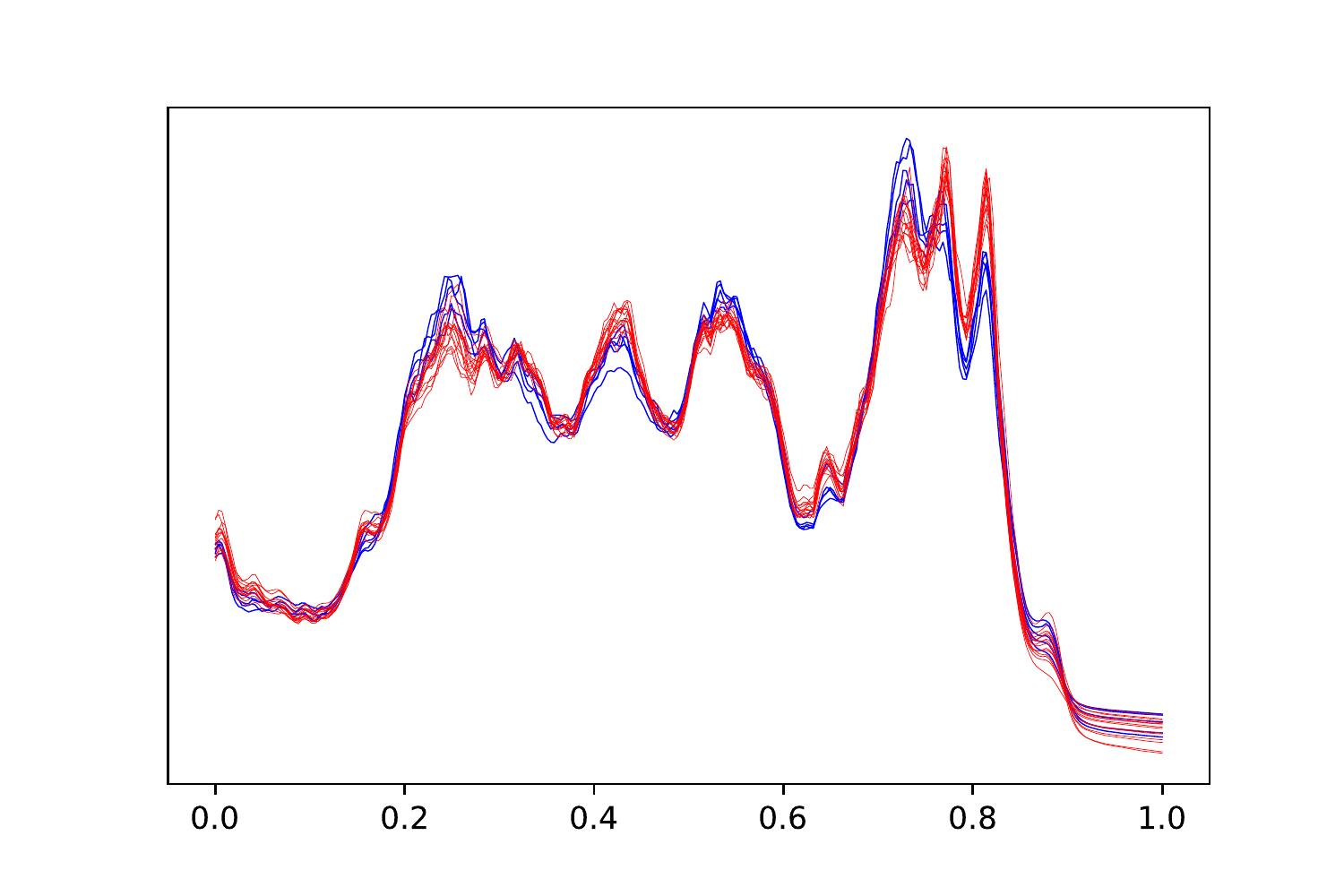}&\includegraphics[width=.2\textwidth, trim=0cm 0cm 1cm 1.2cm,clip=true]{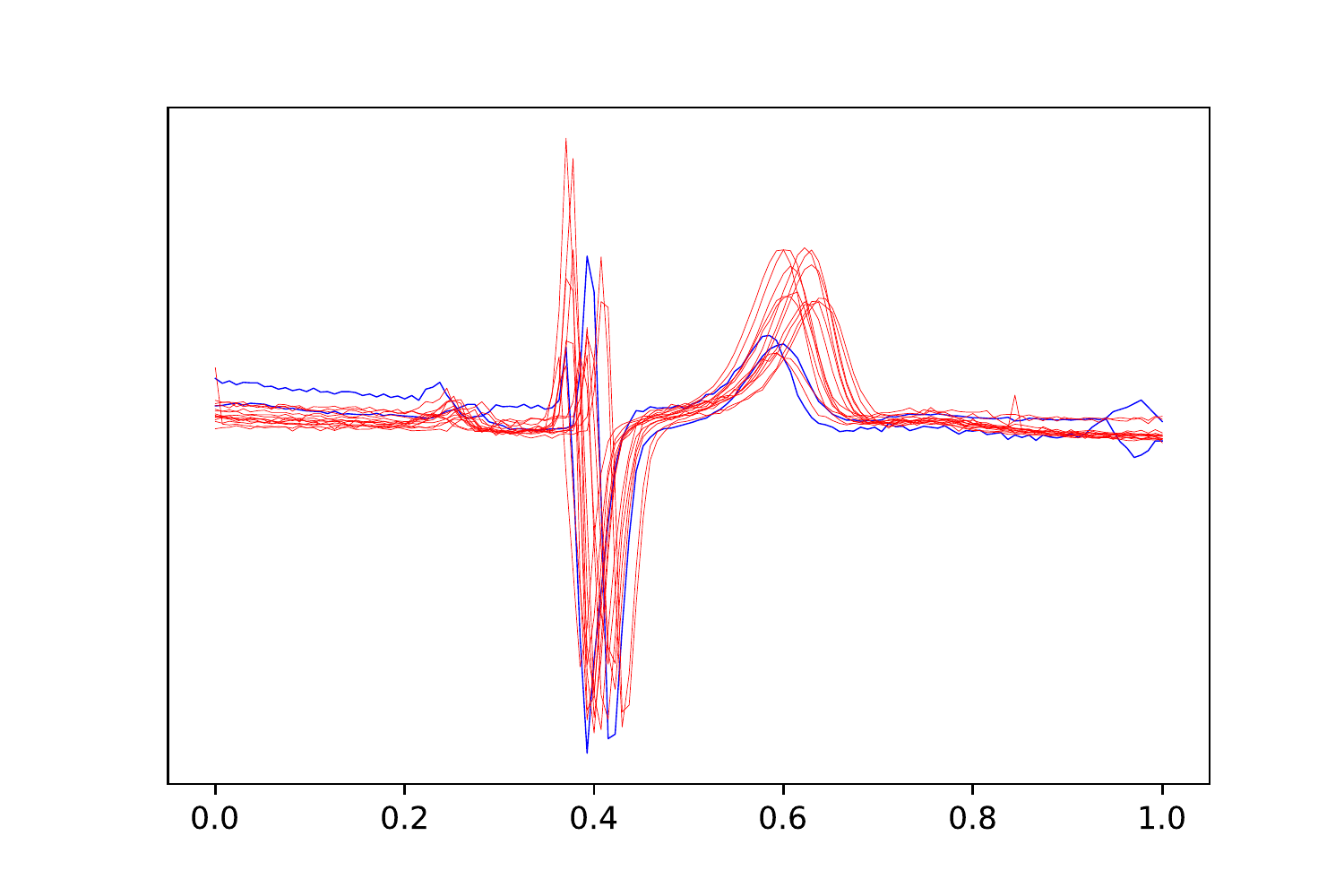}&\includegraphics[width=.2\textwidth, trim=0cm 0cm 1cm 1.2cm,clip=true]{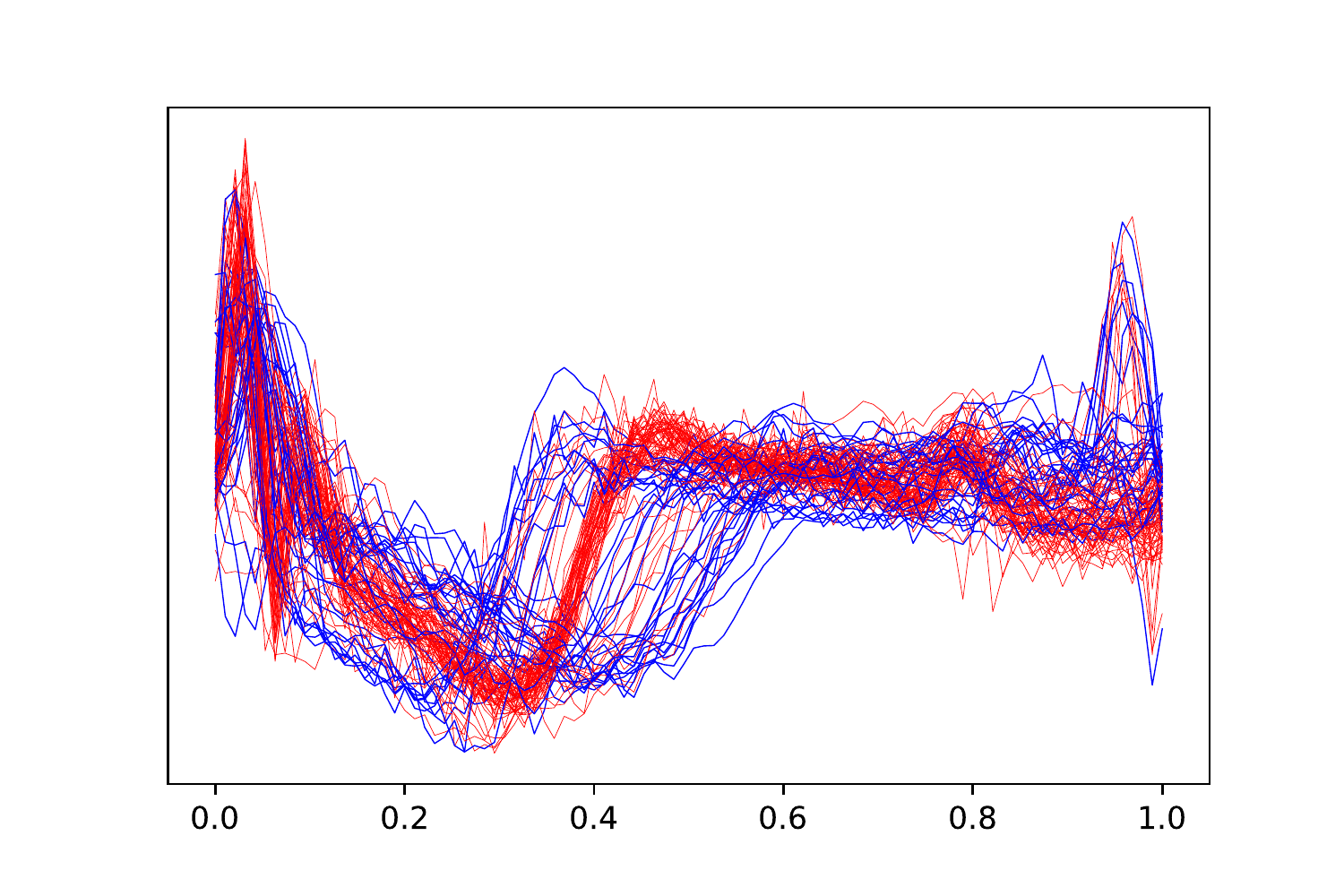}\\
{\tiny Handoutlines}&{\tiny SonyRobotAI1}&{\tiny SonyRobotAI2}&{\tiny StarLightCurves}\\
  \includegraphics[width=.2\textwidth, trim=0cm 0cm 1cm 1.2cm,clip=true]{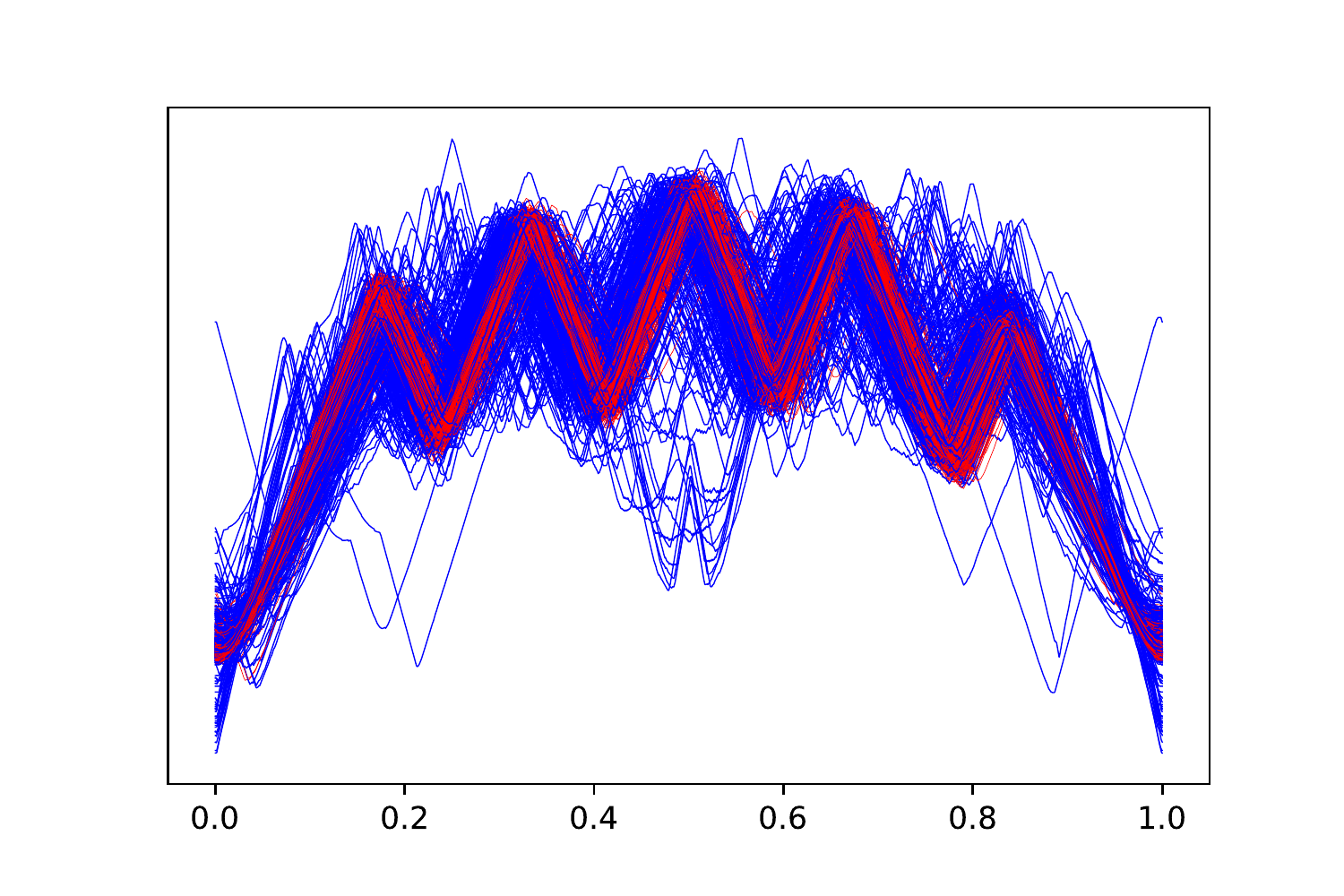}&\includegraphics[width=.2\textwidth, trim=0cm 0cm 1cm 1.2cm,clip=true]{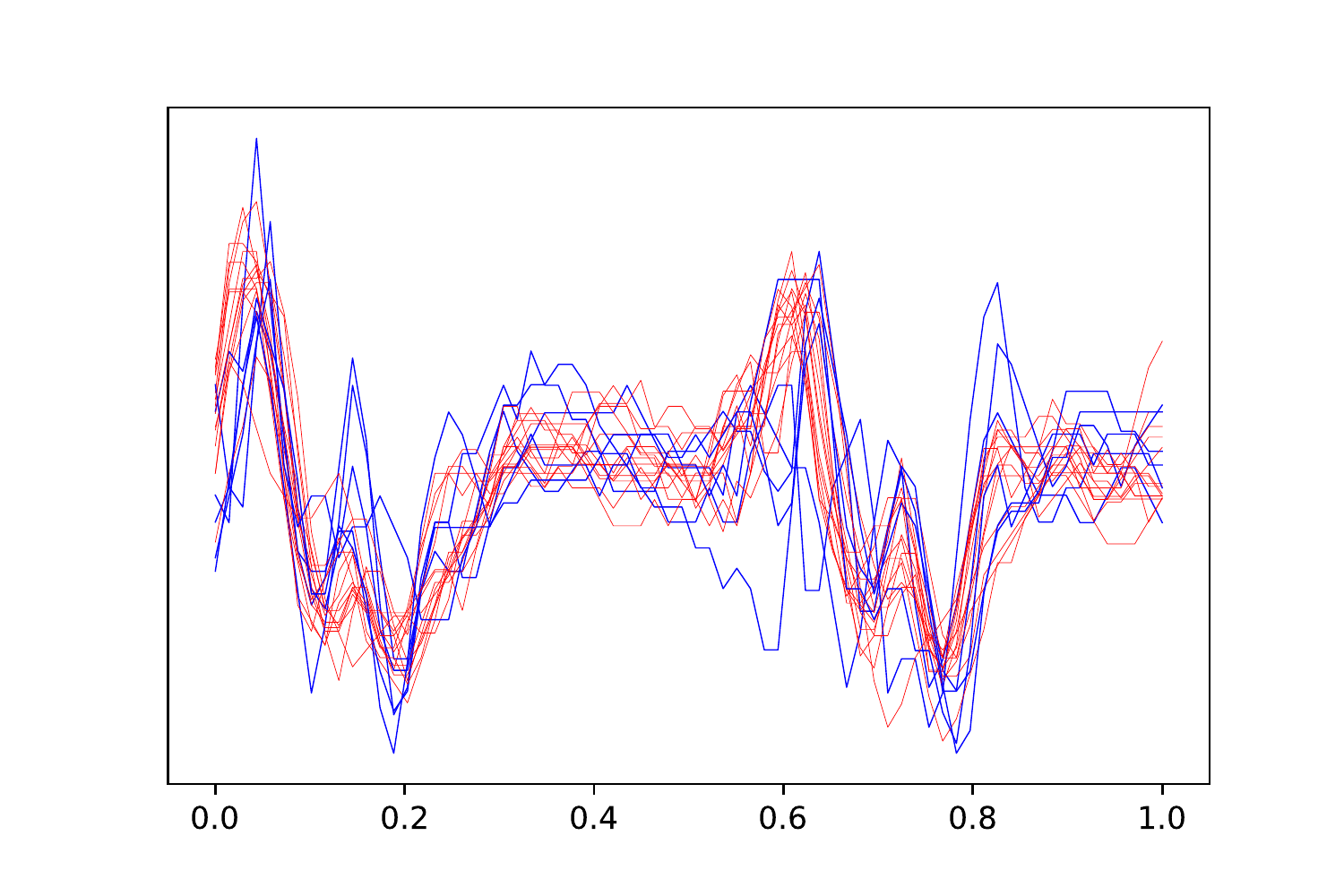}&\includegraphics[width=.2\textwidth, trim=0cm 0cm 1cm 1.2cm,clip=true]{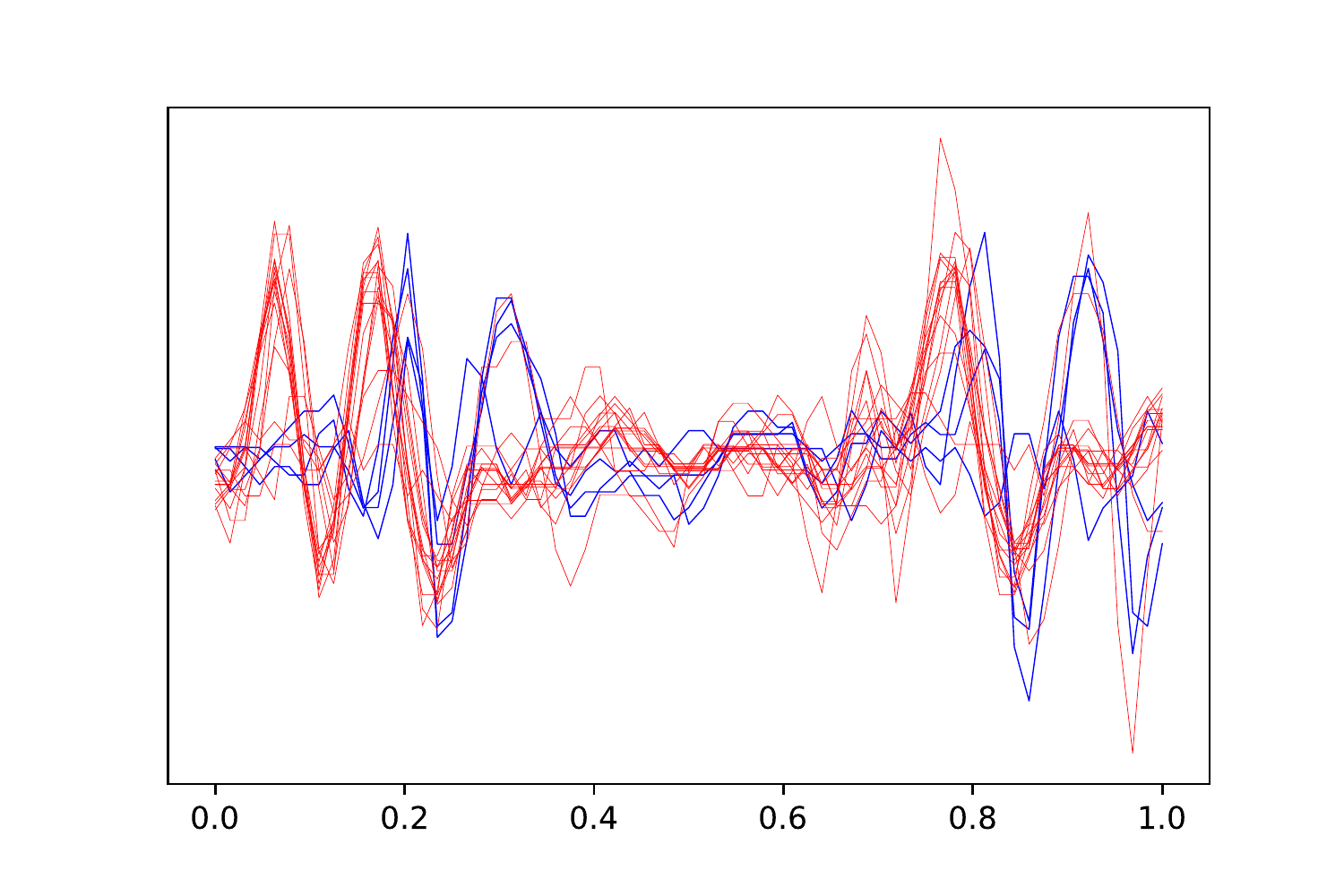}&\includegraphics[width=.2\textwidth, trim=0cm 0cm 1cm 1.2cm,clip=true]{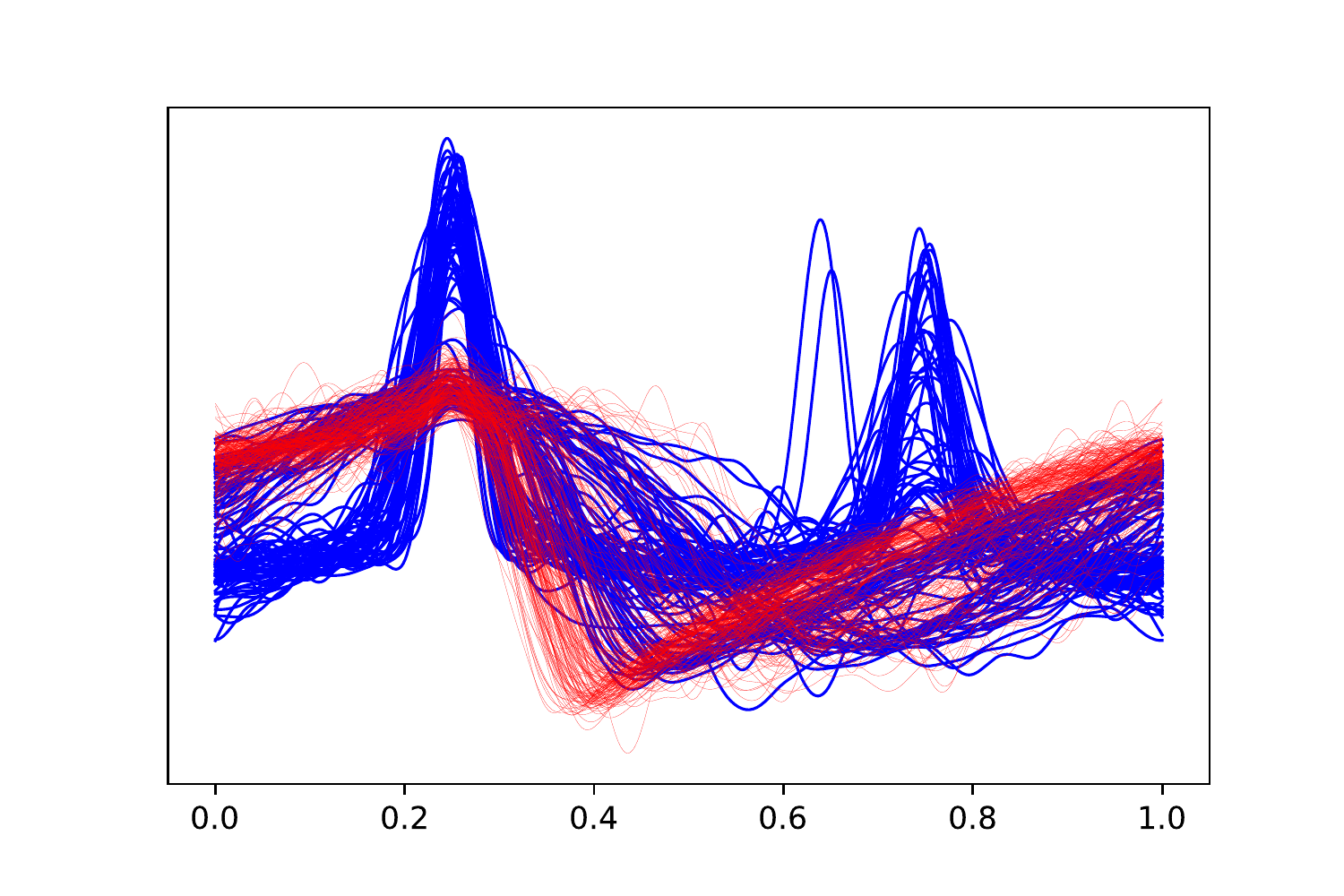}\\
  {\tiny TwoLeadECG}&{\tiny Yoga}&{\tiny EOGHorizontal}&{\tiny CinECGTorso}\\
\includegraphics[width=.2\textwidth, trim=0cm 0cm 1cm 1.2cm,clip=true]{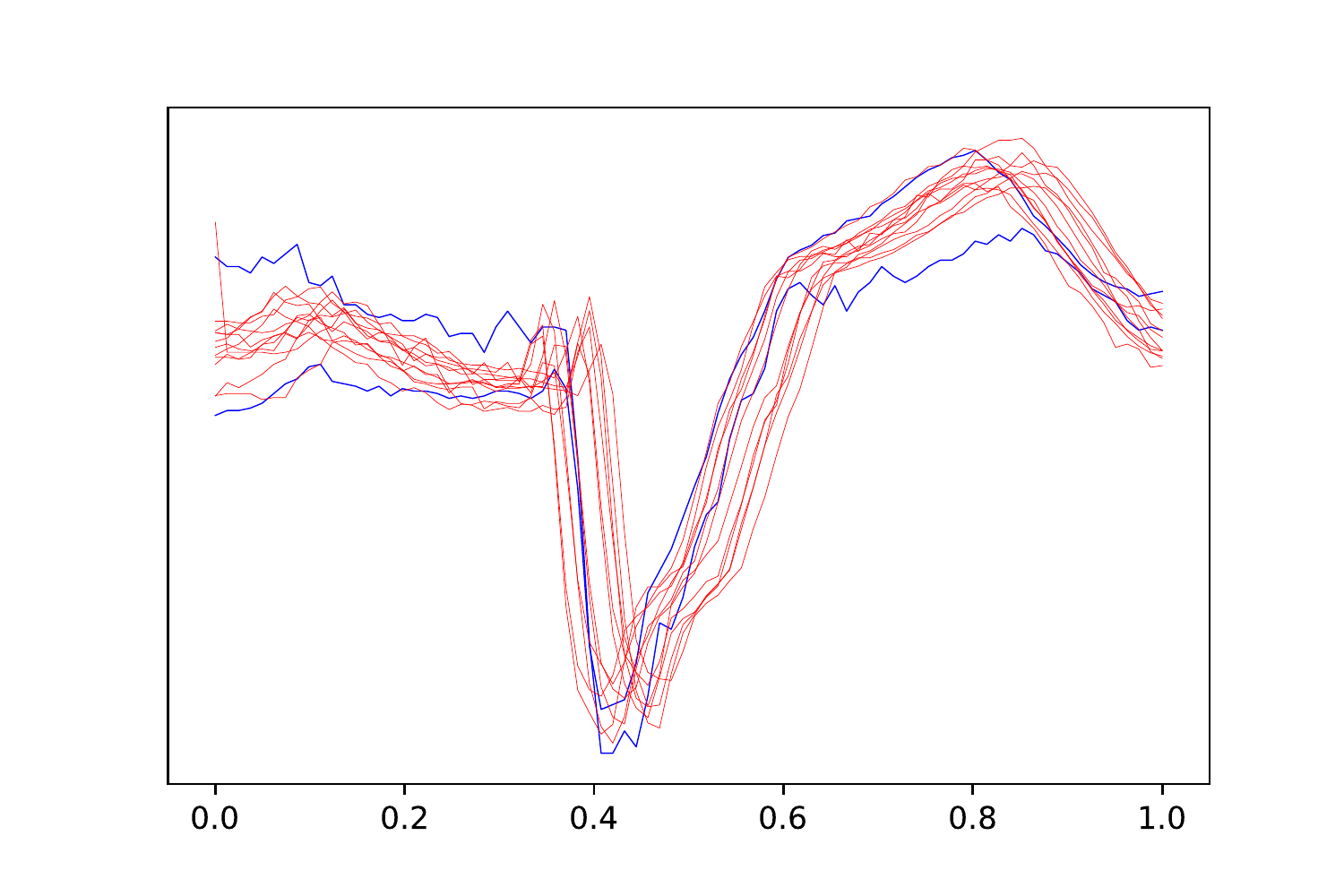}&\includegraphics[width=.2\textwidth, trim=0cm 0cm 1cm 1.2cm,clip=true]{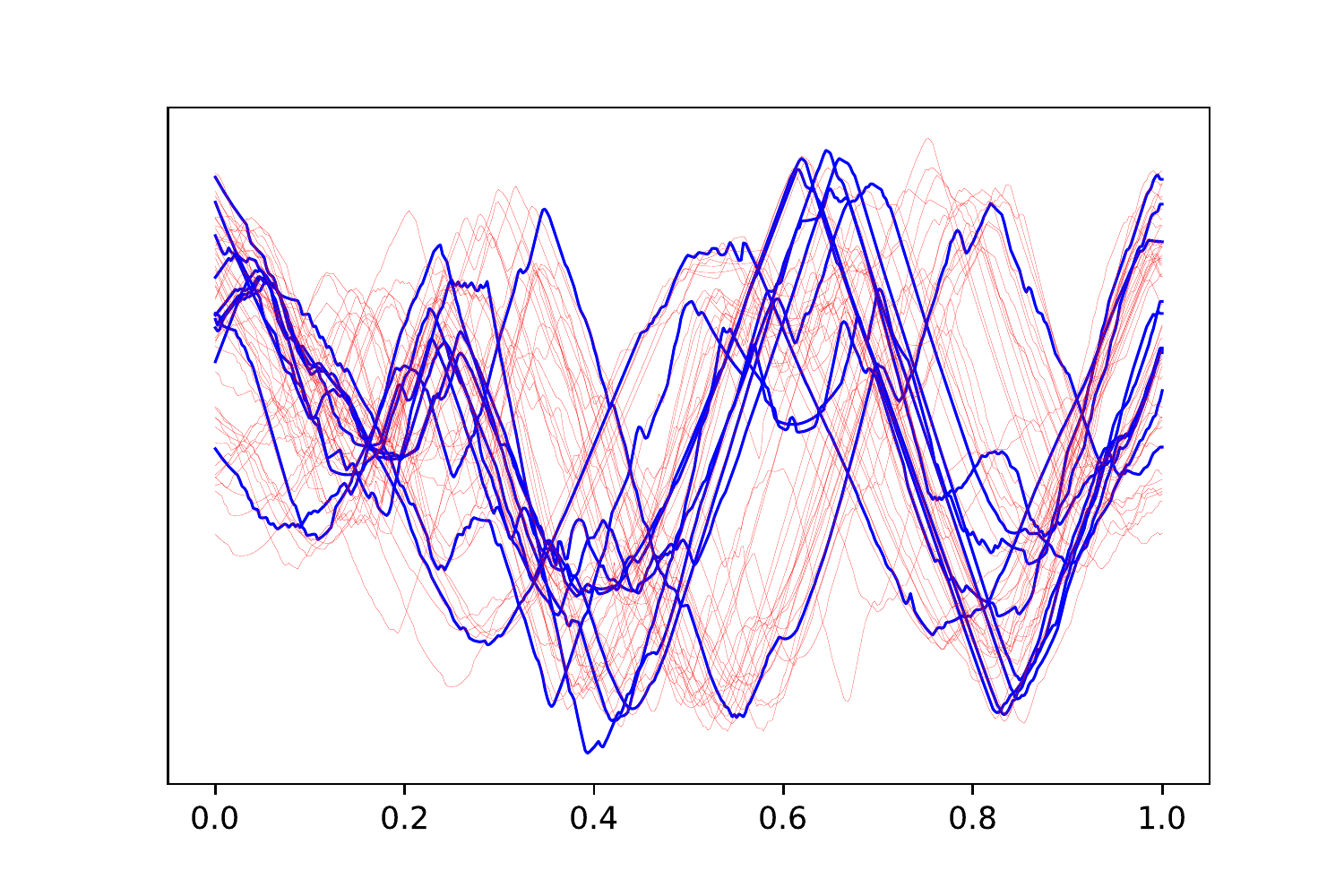}&\includegraphics[width=.2\textwidth, trim=0cm 0cm 1cm 1.2cm,clip=true]{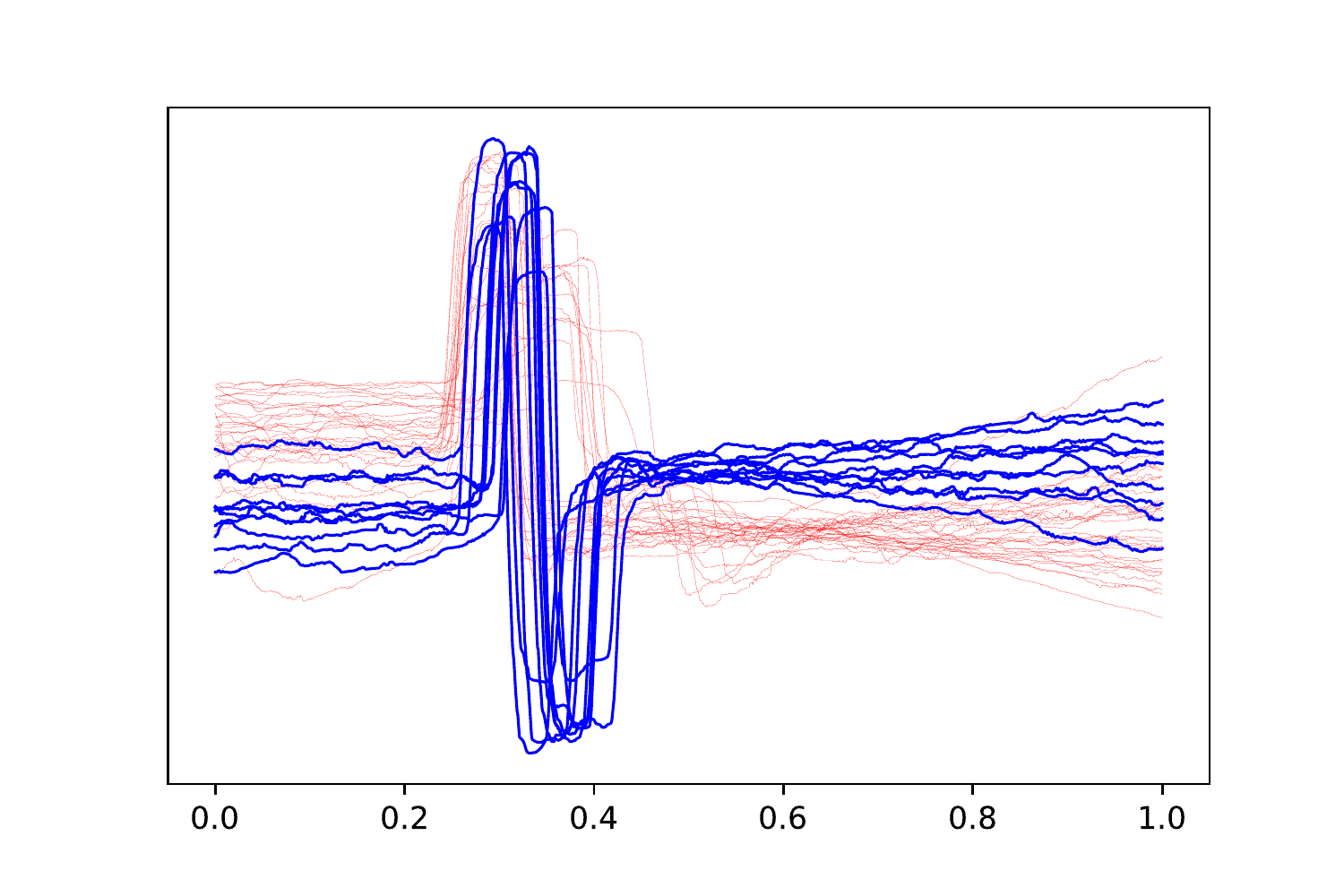}&\includegraphics[width=.2\textwidth, trim=0cm 0cm 1cm 1.2cm,clip=true]{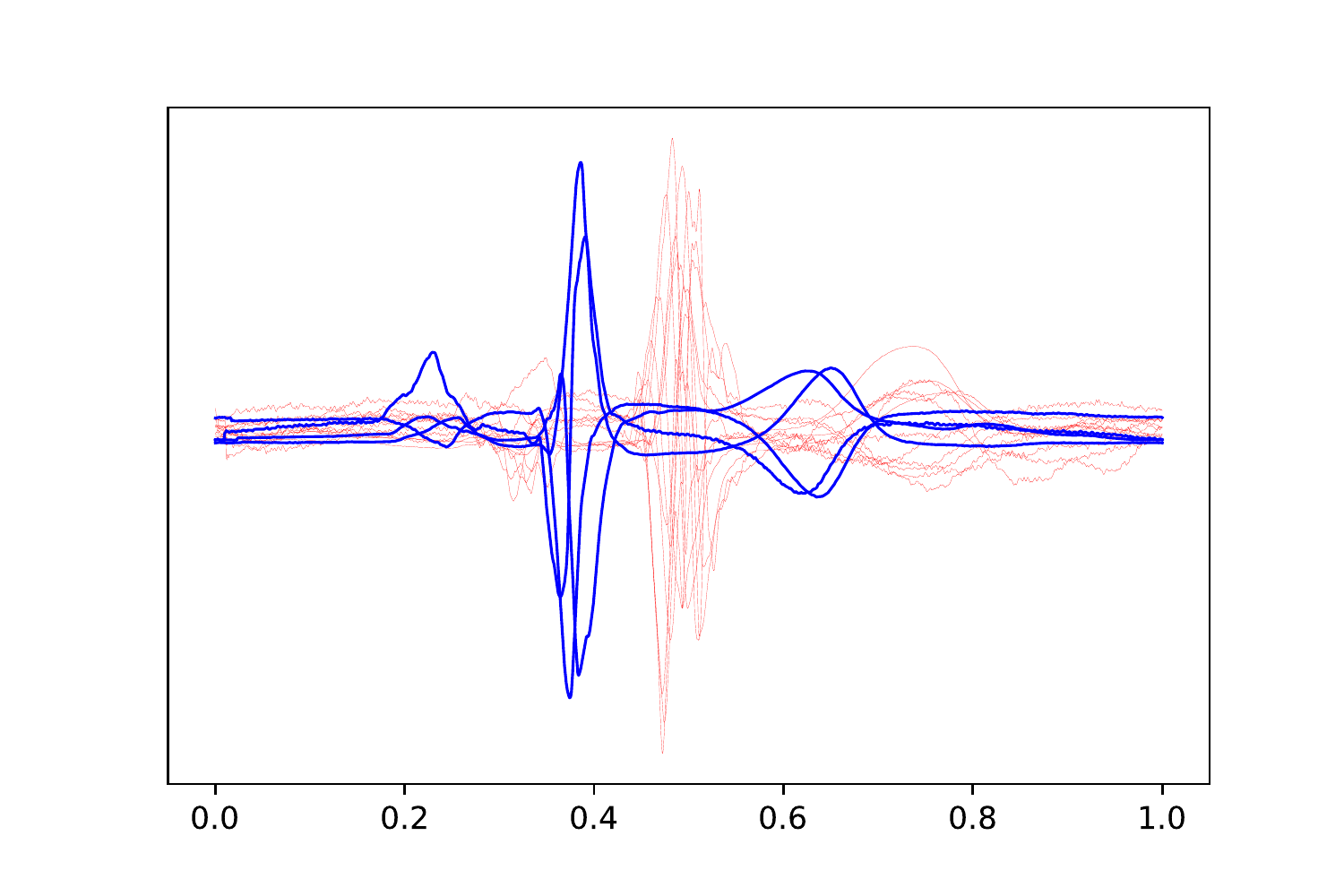}\\
{\tiny ECG5000}\\
\includegraphics[width=.2\textwidth, trim=0cm 0cm 1cm 1.2cm,clip=true]{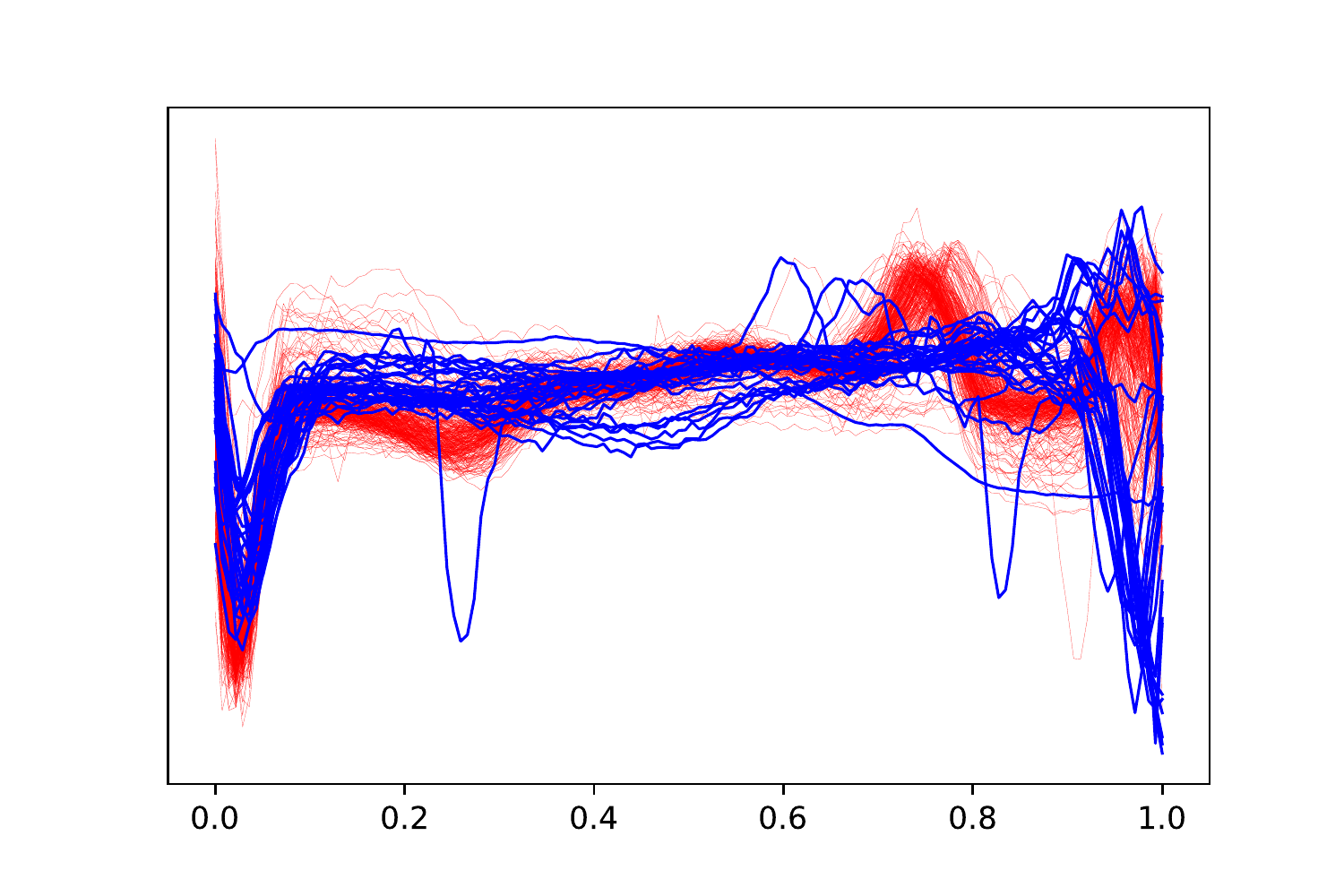}

\end{tabular}

\subsection{Functional depth}

In this part, we present further $8$ functional depths which are outperformed (on an average) by the two depth functions presented in Section~4 on the $13$ real-world datasets and we display their AUC performance. SFD corresponds to \textit{simplicial integrated depth}, HFD to \textit{Halfspace integrated depth}, RP-SD to the \textit{random projection method with simplicial depth}, RP-RHD to the \textit{random projection method with random halfspace depth},  fAO to the \textit{functional adjusted outlyingness}, fDO to the \textit{functional directional outlyingness} to and fbd to \textit{functional bagdistance}. The reader is referred to \cite{cuevas,Fraiman2001,rousseeuw} for the bibliography on employed functional data depth notions.
\begin{center}
\resizebox{9cm}{!} {
{\renewcommand{\arraystretch}{1.5} 
{\setlength{\tabcolsep}{0.4cm}
\begin{tabular}{l  cccccccc}
\toprule
        Datasets &  S-FD &  H-FD  &Modal &RP-SD&RP-RHD&fAO&fDO&fbd \\
\midrule
		Chinatown & 0.74&0.77&0.75&0.77&0.73&0.70&0.83&0.83\\
		Coffee & 0.60&0.59&0.69&0.70&0.51&0.53&0.59&0.60\\
		ECGFiveDays & 0.65&0.64&0.60&0.64&0.56&0.72&0.76&0.80\\
		ECG200 & 0.82&0.82&0.84&0.85&0.74&0.78&0.82&0.82\\
		Handoutlines & 0.70&0.70&0.75&0.72&0.63&0.60&0.71&0.73\\
		SonyRobotAI1 & 0.89&0.89&0.94&0.83&0.62&0.90&0.93&0.93\\
		SonyRobotAI2 & 0.82&0.82&0.92&0.86&0.71&0.80&0.82&0.80\\
		StarLightCurves & 0.80&0.80&0.85&0.78&0.68&0.80&0.82&0.83\\
		TwoLeadECG & 0.68&0.67&0.68&0.66&0.60&0.67&0.69&0.69\\
		Yoga & 0.55&0.53&0.57&0.57&0.54&0.54&0.55&0.56\\
		EOGHorizontal & 0.59&0.52&0.84&0.74&0.64&0.53&0.59&0.66\\
		CinECGTorso & 0.69&0.69&0.73&0.62&0.66&0.85&0.83&0.79\\
		ECG5000 & 0.90&0.90&0.92&0.92&0.84&0.87&0.92&0.92\\\bottomrule
\end{tabular}
}}
}
\end{center}
\subsection{Isolation forest after dimension reduction by filtering methods on the benchmark datasets}
Here, we show the results of the filtering approach using 106 bases from the \texttt{PyWavelets} python library and the Fourier basis. Afterwards, we apply (multivariate) Isolation Forest on the coefficients of the projections and display the AUC performance.
\begin{center}
\resizebox{15cm}{!} {
\begin{tabular}{lrrrrrrrrrrrrrrrrrrrrrr}
\toprule
        Datasets &  Fourier &  bior1.1 &  bior1.3 &  bior1.5 &  bior2.2 &  bior2.4 &  bior2.6 &  bior2.8 &  bior3.1 &  bior3.3 &  bior3.5 &  bior3.7 &  bior3.9 &  bior4.4 &  bior5.5 &  bior6.8 &  coif1 &  coif2 &  coif3 &  coif4 &  coif5 &  coif6 \\
\midrule
        Chinatow &     0.77 &     0.92 &     0.87 &     0.89 &     0.93 &     0.90 &     0.92 &     0.95 &     0.64 &     0.94 &     0.90 &     0.95 &     0.93 &     0.93 &     0.97 &     0.93 &   0.91 &   0.94 &   0.97 &   0.96 &   0.97 &   0.93 \\
          Coffee &     0.49 &     0.56 &     0.67 &     0.67 &     0.60 &     0.56 &     0.69 &     0.60 &     0.51 &     0.53 &     0.65 &     0.53 &     0.47 &     0.47 &     0.62 &     0.65 &   0.51 &   0.76 &   0.50 &   0.71 &   0.54 &   0.69 \\
     ECGFiveDays &     0.58 &     0.78 &     0.78 &     0.80 &     0.73 &     0.72 &     0.73 &     0.80 &     0.58 &     0.68 &     0.67 &     0.82 &     0.70 &     0.67 &     0.75 &     0.75 &   0.75 &   0.75 &   0.82 &   0.69 &   0.75 &   0.69 \\
          ECG200 &     0.46 &     0.70 &     0.72 &     0.68 &     0.57 &     0.53 &     0.59 &     0.66 &     0.44 &     0.46 &     0.50 &     0.54 &     0.53 &     0.52 &     0.59 &     0.68 &   0.61 &   0.60 &   0.63 &   0.54 &   0.69 &   0.68 \\
    Handoutlines &     0.50 &     0.74 &     0.77 &     0.77 &     0.56 &     0.59 &     0.55 &     0.56 &     0.57 &     0.51 &     0.52 &     0.54 &     0.51 &     0.58 &     0.47 &     0.49 &   0.57 &   0.52 &   0.53 &   0.56 &   0.56 &   0.55 \\
    SonyRobotAI1 &     0.98 &     0.95 &     0.95 &     0.97 &     0.98 &     0.97 &     0.98 &     0.97 &     0.96 &     0.96 &     0.97 &     0.96 &     0.97 &     0.97 &     0.94 &     0.96 &   0.97 &   0.98 &   0.98 &   0.95 &   0.96 &   0.97 \\
    SonyRobotAI2 &     0.89 &     0.81 &     0.80 &     0.78 &     0.81 &     0.84 &     0.81 &     0.84 &     0.83 &     0.81 &     0.81 &     0.80 &     0.84 &     0.83 &     0.83 &     0.84 &   0.82 &   0.83 &   0.87 &   0.83 &   0.85 &   0.78 \\
 StarLightCurves &     0.46 &     0.68 &     0.69 &     0.70 &     0.58 &     0.55 &     0.58 &     0.58 &     0.53 &     0.54 &     0.55 &     0.54 &     0.62 &     0.57 &     0.54 &     0.60 &   0.55 &   0.57 &   0.57 &   0.62 &   0.63 &   0.69 \\
      TwoLeadECG &     0.52 &     0.64 &     0.63 &     0.66 &     0.57 &     0.56 &     0.58 &     0.59 &     0.55 &     0.57 &     0.58 &     0.56 &     0.59 &     0.59 &     0.56 &     0.62 &   0.54 &   0.65 &   0.58 &   0.62 &   0.59 &   0.60 \\
            Yoga &     0.63 &     0.59 &     0.59 &     0.61 &     0.58 &     0.59 &     0.58 &     0.60 &     0.61 &     0.61 &     0.61 &     0.61 &     0.62 &     0.59 &     0.59 &     0.60 &   0.58 &   0.57 &   0.60 &   0.61 &   0.61 &   0.61 \\
   EOGHorizontal &     0.44 &     0.61 &     0.60 &     0.56 &     0.61 &     0.59 &     0.60 &     0.61 &     0.62 &     0.63 &     0.63 &     0.62 &     0.60 &     0.66 &     0.63 &     0.64 &   0.63 &   0.64 &   0.62 &   0.62 &   0.63 &   0.63 \\
     CinECGTorso &     0.28 &     0.25 &     0.24 &     0.20 &     0.17 &     0.16 &     0.15 &     0.17 &     0.15 &     0.17 &     0.16 &     0.15 &     0.14 &     0.14 &     0.15 &     0.14 &   0.14 &   0.16 &   0.17 &   0.16 &   0.15 &   0.15 \\
         ECG5000 &     0.65 &     0.85 &     0.87 &     0.89 &     0.72 &     0.76 &     0.77 &     0.82 &     0.65 &     0.66 &     0.71 &     0.73 &     0.81 &     0.77 &     0.75 &     0.78 &   0.78 &   0.75 &   0.79 &   0.84 &   0.80 &   0.82 \\
\bottomrule
\end{tabular}

}
\end{center}
\begin{center}
\resizebox{15cm}{!} {
\begin{tabular}{lrrrrrrrrrrrrrrrrrrrrrr}
\toprule
        Datasets &  coif7 &  coif8 &  coif9 &  coif10 &  coif11 &  coif12 &  coif13 &  coif14 &  coif15 &  coif16 &  coif17 &   db1 &   db2 &   db3 &   db4 &   db5 &   db6 &   db7 &   db8 &   db9 &  db10 &  db11 \\
\midrule
        Chinatow &   0.95 &   0.95 &   0.94 &    0.93 &    0.93 &    0.95 &    0.98 &    0.95 &    0.90 &    0.93 &    0.91 &  0.94 &  0.90 &  0.92 &  0.97 &  0.89 &  0.92 &  0.94 &  0.95 &  0.97 &  0.94 &  0.95 \\
          Coffee &   0.68 &   0.59 &   0.49 &    0.69 &    0.65 &    0.59 &    0.73 &    0.63 &    0.65 &    0.56 &    0.51 &  0.65 &  0.51 &  0.67 &  0.60 &  0.37 &  0.56 &  0.47 &  0.55 &  0.56 &  0.63 &  0.63 \\
     ECGFiveDays &   0.67 &   0.71 &   0.68 &    0.71 &    0.70 &    0.77 &    0.83 &    0.77 &    0.79 &    0.79 &    0.80 &  0.70 &  0.78 &  0.80 &  0.82 &  0.70 &  0.79 &  0.73 &  0.74 &  0.74 &  0.73 &  0.66 \\
          ECG200 &   0.72 &   0.71 &   0.84 &    0.86 &    0.85 &    0.85 &    0.85 &    0.84 &    0.85 &    0.85 &    0.86 &  0.63 &  0.59 &  0.57 &  0.55 &  0.65 &  0.57 &  0.63 &  0.64 &  0.62 &  0.69 &  0.70 \\
    Handoutlines &   0.47 &   0.56 &   0.55 &    0.57 &    0.57 &    0.55 &    0.60 &    0.57 &    0.59 &    0.51 &    0.61 &  0.74 &  0.55 &  0.58 &  0.58 &  0.54 &  0.52 &  0.57 &  0.52 &  0.51 &  0.53 &  0.60 \\
    SonyRobotAI1 &   0.92 &   0.94 &   0.96 &    0.91 &    0.91 &    0.92 &    0.90 &    0.93 &    0.89 &    0.94 &    0.90 &  0.97 &  0.98 &  0.98 &  0.97 &  0.98 &  0.96 &  0.97 &  0.97 &  0.95 &  0.96 &  0.97 \\
    SonyRobotAI2 &   0.79 &   0.80 &   0.82 &    0.79 &    0.79 &    0.79 &    0.80 &    0.78 &    0.83 &    0.78 &    0.83 &  0.79 &  0.82 &  0.81 &  0.87 &  0.81 &  0.86 &  0.83 &  0.87 &  0.87 &  0.85 &  0.90 \\
 StarLightCurves &   0.68 &   0.71 &   0.64 &    0.72 &    0.76 &    0.77 &    0.76 &    0.75 &    0.75 &    0.77 &    0.75 &  0.70 &  0.57 &  0.53 &  0.53 &  0.60 &  0.52 &  0.54 &  0.61 &  0.61 &  0.65 &  0.57 \\
      TwoLeadECG &   0.65 &   0.67 &   0.68 &    0.69 &    0.64 &    0.76 &    0.71 &    0.73 &    0.73 &    0.67 &    0.69 &  0.58 &  0.58 &  0.57 &  0.57 &  0.57 &  0.62 &  0.57 &  0.53 &  0.60 &  0.62 &  0.63 \\
            Yoga &   0.62 &   0.59 &   0.59 &    0.59 &    0.62 &    0.62 &    0.61 &    0.61 &    0.61 &    0.60 &    0.59 &  0.58 &  0.57 &  0.58 &  0.58 &  0.60 &  0.60 &  0.60 &  0.59 &  0.60 &  0.61 &  0.61 \\
   EOGHorizontal &   0.60 &   0.61 &   0.65 &    0.56 &    0.68 &    0.59 &    0.60 &    0.64 &    0.62 &    0.60 &    0.59 &  0.63 &  0.61 &  0.62 &  0.64 &  0.65 &  0.62 &  0.61 &  0.60 &  0.63 &  0.64 &  0.57 \\
     CinECGTorso &   0.15 &   0.17 &   0.18 &    0.17 &    0.17 &    0.16 &    0.17 &    0.17 &    0.19 &    0.17 &    0.16 &  0.23 &  0.14 &  0.16 &  0.15 &  0.18 &  0.16 &  0.17 &  0.14 &  0.17 &  0.16 &  0.18 \\
         ECG5000 &   0.86 &   0.85 &   0.85 &    0.83 &    0.87 &    0.92 &    0.91 &    0.93 &    0.92 &    0.91 &    0.93 &  0.85 &  0.79 &  0.75 &  0.71 &  0.69 &  0.78 &  0.74 &  0.73 &  0.73 &  0.74 &  0.72 \\
\bottomrule
\end{tabular}

}
\end{center}
\begin{center}
\resizebox{15cm}{!} {
\begin{tabular}{lrrrrrrrrrrrrrrrrrrrrrr}
\toprule
        Datasets &  db12 &  db13 &  db14 &  db15 &  db16 &  db17 &  db18 &  db19 &  db20 &  db21 &  db22 &  db23 &  db24 &  db25 &  db26 &  db27 &  db28 &  db29 &  db30 &  db31 &  db32 &  db33 \\
\midrule
        Chinatow &  0.95 &  0.93 &  0.95 &  0.92 &  0.96 &  0.92 &  0.96 &  0.95 &  0.92 &  0.96 &  0.94 &  0.94 &  0.95 &  0.97 &  0.97 &  0.94 &  0.94 &  0.97 &  0.97 &  0.96 &  0.96 &  0.93 \\
          Coffee &  0.53 &  0.65 &  0.62 &  0.71 &  0.55 &  0.63 &  0.58 &  0.51 &  0.51 &  0.64 &  0.51 &  0.63 &  0.53 &  0.45 &  0.73 &  0.55 &  0.47 &  0.62 &  0.68 &  0.50 &  0.54 &  0.56 \\
     ECGFiveDays &  0.69 &  0.65 &  0.66 &  0.70 &  0.70 &  0.68 &  0.73 &  0.68 &  0.64 &  0.67 &  0.62 &  0.66 &  0.68 &  0.70 &  0.72 &  0.68 &  0.66 &  0.67 &  0.66 &  0.64 &  0.67 &  0.65 \\
          ECG200 &  0.65 &  0.72 &  0.72 &  0.72 &  0.75 &  0.72 &  0.71 &  0.73 &  0.74 &  0.70 &  0.74 &  0.73 &  0.71 &  0.84 &  0.84 &  0.84 &  0.84 &  0.84 &  0.85 &  0.84 &  0.84 &  0.82 \\
    Handoutlines &  0.55 &  0.55 &  0.52 &  0.53 &  0.55 &  0.57 &  0.55 &  0.54 &  0.56 &  0.50 &  0.61 &  0.57 &  0.55 &  0.51 &  0.53 &  0.62 &  0.51 &  0.54 &  0.51 &  0.55 &  0.58 &  0.56 \\
    SonyRobotAI1 &  0.97 &  0.99 &  0.98 &  0.98 &  0.97 &  0.98 &  0.98 &  0.93 &  0.90 &  0.92 &  0.94 &  0.94 &  0.96 &  0.92 &  0.92 &  0.91 &  0.97 &  0.94 &  0.92 &  0.96 &  0.96 &  0.91 \\
    SonyRobotAI2 &  0.94 &  0.90 &  0.88 &  0.90 &  0.89 &  0.76 &  0.86 &  0.80 &  0.79 &  0.82 &  0.78 &  0.80 &  0.79 &  0.81 &  0.80 &  0.75 &  0.78 &  0.77 &  0.80 &  0.82 &  0.83 &  0.80 \\
 StarLightCurves &  0.54 &  0.66 &  0.67 &  0.66 &  0.67 &  0.68 &  0.66 &  0.69 &  0.68 &  0.67 &  0.70 &  0.70 &  0.65 &  0.68 &  0.69 &  0.64 &  0.66 &  0.68 &  0.70 &  0.72 &  0.68 &  0.76 \\
      TwoLeadECG &  0.62 &  0.66 &  0.63 &  0.62 &  0.62 &  0.62 &  0.62 &  0.66 &  0.62 &  0.59 &  0.65 &  0.66 &  0.67 &  0.65 &  0.66 &  0.67 &  0.70 &  0.61 &  0.74 &  0.74 &  0.68 &  0.68 \\
            Yoga &  0.58 &  0.59 &  0.61 &  0.61 &  0.59 &  0.61 &  0.59 &  0.59 &  0.60 &  0.59 &  0.61 &  0.59 &  0.60 &  0.59 &  0.59 &  0.60 &  0.61 &  0.60 &  0.63 &  0.60 &  0.61 &  0.60 \\
   EOGHorizontal &  0.63 &  0.63 &  0.65 &  0.67 &  0.57 &  0.59 &  0.61 &  0.68 &  0.63 &  0.60 &  0.61 &  0.63 &  0.64 &  0.65 &  0.65 &  0.65 &  0.65 &  0.61 &  0.61 &  0.66 &  0.63 &  0.59 \\
     CinECGTorso &  0.16 &  0.16 &  0.19 &  0.17 &  0.18 &  0.16 &  0.14 &  0.17 &  0.16 &  0.17 &  0.17 &  0.15 &  0.17 &  0.16 &  0.16 &  0.18 &  0.18 &  0.16 &  0.18 &  0.17 &  0.17 &  0.18 \\
         ECG5000 &  0.77 &  0.73 &  0.79 &  0.76 &  0.77 &  0.79 &  0.78 &  0.86 &  0.81 &  0.85 &  0.83 &  0.77 &  0.84 &  0.84 &  0.85 &  0.82 &  0.84 &  0.82 &  0.83 &  0.81 &  0.81 &  0.81 \\
\bottomrule
\end{tabular}

}
\end{center}
\begin{center}
\resizebox{15cm}{!} {
\begin{tabular}{lrrrrrrrrrrrrrrrrrrrrrr}
\toprule
        Datasets &  db34 &  db35 &  db36 &  db37 &  db38 &  dmey &  haar &  rbio1.1 &  rbio1.3 &  rbio1.5 &  rbio2.2 &  rbio2.4 &  rbio2.6 &  rbio2.8 &  rbio3.1 &  rbio3.3 &  rbio3.5 &  rbio3.7 &  rbio3.9 &  rbio4.4 &  rbio5.5 &  rbio6.8 \\
\midrule
        Chinatow &  0.98 &  0.95 &  0.93 &  0.92 &  0.95 &  0.94 &  0.89 &     0.92 &     0.91 &     0.93 &     0.93 &     0.90 &     0.96 &     0.97 &     0.97 &     0.96 &     0.95 &     0.96 &     0.97 &     0.88 &     0.95 &     0.95 \\
          Coffee &  0.55 &  0.62 &  0.55 &  0.62 &  0.51 &  0.55 &  0.56 &     0.74 &     0.64 &     0.47 &     0.64 &     0.56 &     0.56 &     0.50 &     0.72 &     0.59 &     0.60 &     0.53 &     0.67 &     0.41 &     0.55 &     0.60 \\
     ECGFiveDays &  0.67 &  0.76 &  0.79 &  0.82 &  0.74 &  0.72 &  0.74 &     0.76 &     0.81 &     0.73 &     0.80 &     0.79 &     0.75 &     0.76 &     0.89 &     0.85 &     0.83 &     0.84 &     0.79 &     0.73 &     0.72 &     0.83 \\
          ECG200 &  0.84 &  0.85 &  0.83 &  0.82 &  0.86 &  0.84 &  0.64 &     0.71 &     0.55 &     0.60 &     0.67 &     0.58 &     0.66 &     0.62 &     0.75 &     0.73 &     0.62 &     0.64 &     0.65 &     0.63 &     0.60 &     0.57 \\
    Handoutlines &  0.55 &  0.60 &  0.57 &  0.50 &  0.55 &  0.58 &  0.73 &     0.71 &     0.55 &     0.55 &     0.57 &     0.54 &     0.55 &     0.58 &     0.73 &     0.56 &     0.56 &     0.56 &     0.54 &     0.56 &     0.53 &     0.53 \\
    SonyRobotAI1 &  0.93 &  0.95 &  0.96 &  0.92 &  0.95 &  0.94 &  0.97 &     0.96 &     0.96 &     0.97 &     0.95 &     0.97 &     0.97 &     0.96 &     0.90 &     0.97 &     0.97 &     0.97 &     0.96 &     0.95 &     0.95 &     0.97 \\
    SonyRobotAI2 &  0.77 &  0.79 &  0.81 &  0.85 &  0.80 &  0.77 &  0.82 &     0.82 &     0.83 &     0.84 &     0.81 &     0.81 &     0.78 &     0.83 &     0.80 &     0.79 &     0.86 &     0.80 &     0.88 &     0.85 &     0.82 &     0.80 \\
 StarLightCurves &  0.75 &  0.73 &  0.76 &  0.74 &  0.76 &  0.81 &  0.70 &     0.70 &     0.54 &     0.56 &     0.56 &     0.56 &     0.58 &     0.62 &     0.71 &     0.53 &     0.55 &     0.60 &     0.60 &     0.56 &     0.57 &     0.51 \\
      TwoLeadECG &  0.70 &  0.73 &  0.69 &  0.69 &  0.69 &  0.67 &  0.53 &     0.59 &     0.61 &     0.59 &     0.64 &     0.58 &     0.58 &     0.58 &     0.68 &     0.63 &     0.59 &     0.57 &     0.58 &     0.60 &     0.60 &     0.60 \\
            Yoga &  0.60 &  0.60 &  0.62 &  0.61 &  0.62 &  0.60 &  0.59 &     0.60 &     0.59 &     0.59 &     0.58 &     0.57 &     0.59 &     0.59 &     0.61 &     0.58 &     0.58 &     0.57 &     0.58 &     0.60 &     0.60 &     0.58 \\
   EOGHorizontal &  0.63 &  0.62 &  0.59 &  0.64 &  0.62 &  0.67 &  0.60 &     0.55 &     0.63 &     0.58 &     0.59 &     0.64 &     0.64 &     0.63 &     0.63 &     0.64 &     0.56 &     0.63 &     0.64 &     0.57 &     0.61 &     0.59 \\
     CinECGTorso &  0.19 &  0.17 &  0.18 &  0.16 &  0.19 &  0.18 &  0.25 &     0.23 &     0.16 &     0.18 &     0.15 &     0.16 &     0.15 &     0.15 &     0.28 &     0.14 &     0.16 &     0.17 &     0.19 &     0.15 &     0.15 &     0.17 \\
         ECG5000 &  0.83 &  0.79 &  0.93 &  0.92 &  0.93 &  0.84 &  0.86 &     0.85 &     0.75 &     0.73 &     0.82 &     0.81 &     0.79 &     0.83 &     0.90 &     0.82 &     0.83 &     0.79 &     0.83 &     0.78 &     0.83 &     0.83 \\
\bottomrule
\end{tabular}

}
\end{center}
\begin{center}
\resizebox{15cm}{!} {
\begin{tabular}{lrrrrrrrrrrrrrrrrrrr}
\toprule
        Datasets &  sym2 &  sym3 &  sym4 &  sym5 &  sym6 &  sym7 &  sym8 &  sym9 &  sym10 &  sym11 &  sym12 &  sym13 &  sym14 &  sym15 &  sym16 &  sym17 &  sym18 &  sym19 &  sym20 \\
\midrule
        Chinatow &  0.86 &  0.90 &  0.95 &  0.89 &  0.83 &  0.97 &  0.96 &  0.97 &   0.88 &   0.94 &   0.95 &   0.89 &   0.94 &   0.96 &   0.94 &   0.95 &   0.94 &   0.94 &   0.95 \\
          Coffee &  0.50 &  0.55 &  0.46 &  0.63 &  0.51 &  0.46 &  0.59 &  0.54 &   0.59 &   0.41 &   0.50 &   0.46 &   0.58 &   0.51 &   0.62 &   0.64 &   0.56 &   0.46 &   0.47 \\
     ECGFiveDays &  0.80 &  0.75 &  0.75 &  0.76 &  0.77 &  0.76 &  0.80 &  0.87 &   0.74 &   0.68 &   0.75 &   0.74 &   0.72 &   0.77 &   0.72 &   0.75 &   0.72 &   0.71 &   0.69 \\
          ECG200 &  0.58 &  0.52 &  0.54 &  0.54 &  0.49 &  0.58 &  0.62 &  0.58 &   0.57 &   0.59 &   0.63 &   0.65 &   0.71 &   0.72 &   0.74 &   0.74 &   0.67 &   0.64 &   0.70 \\
    Handoutlines &  0.63 &  0.59 &  0.49 &  0.55 &  0.56 &  0.57 &  0.57 &  0.53 &   0.53 &   0.55 &   0.56 &   0.56 &   0.51 &   0.60 &   0.53 &   0.51 &   0.51 &   0.54 &   0.54 \\
    SonyRobotAI1 &  0.97 &  0.98 &  0.94 &  0.98 &  0.97 &  0.97 &  0.96 &  0.95 &   0.96 &   0.97 &   0.98 &   0.95 &   0.96 &   0.96 &   0.94 &   0.96 &   0.96 &   0.94 &   0.96 \\
    SonyRobotAI2 &  0.86 &  0.86 &  0.81 &  0.87 &  0.86 &  0.84 &  0.86 &  0.86 &   0.83 &   0.86 &   0.89 &   0.89 &   0.87 &   0.86 &   0.89 &   0.81 &   0.79 &   0.83 &   0.80 \\
 StarLightCurves &  0.57 &  0.53 &  0.55 &  0.58 &  0.59 &  0.53 &  0.58 &  0.66 &   0.62 &   0.64 &   0.64 &   0.63 &   0.59 &   0.58 &   0.59 &   0.71 &   0.67 &   0.71 &   0.65 \\
      TwoLeadECG &  0.58 &  0.57 &  0.55 &  0.56 &  0.63 &  0.62 &  0.55 &  0.57 &   0.58 &   0.65 &   0.59 &   0.64 &   0.62 &   0.68 &   0.62 &   0.61 &   0.60 &   0.66 &   0.57 \\
            Yoga &  0.58 &  0.58 &  0.59 &  0.60 &  0.60 &  0.59 &  0.59 &  0.58 &   0.60 &   0.60 &   0.61 &   0.59 &   0.60 &   0.59 &   0.60 &   0.62 &   0.62 &   0.61 &   0.61 \\
   EOGHorizontal &  0.60 &  0.66 &  0.62 &  0.55 &  0.61 &  0.63 &  0.63 &  0.61 &   0.61 &   0.65 &   0.67 &   0.68 &   0.58 &   0.67 &   0.64 &   0.67 &   0.71 &   0.64 &   0.60 \\
     CinECGTorso &  0.16 &  0.16 &  0.15 &  0.15 &  0.17 &  0.15 &  0.16 &  0.16 &   0.17 &   0.15 &   0.17 &   0.16 &   0.16 &   0.15 &   0.16 &   0.15 &   0.17 &   0.17 &   0.17 \\
         ECG5000 &  0.79 &  0.75 &  0.81 &  0.77 &  0.81 &  0.72 &  0.79 &  0.78 &   0.77 &   0.79 &   0.84 &   0.83 &   0.82 &   0.80 &   0.84 &   0.84 &   0.86 &   0.81 &   0.83 \\
\bottomrule
\end{tabular}

}
\end{center}
\subsection{IF$_{FPCA}$ with different filtering preliminary step on the benchmark datasets}
Here, we show the results of the FPCA approach using the Fourier basis and $106$ further bases from the \textit{PyWavelets} python library as preliminary filtering stage. Afterwards, we apply (multivariate) Isolation Forest on the coefficients of the projections and display the AUC performance.
\begin{center}
\resizebox{15cm}{!} {
\begin{tabular}{lrrrrrrrrrrrrrrrrrrrrrr}
\toprule
        Datasets &  bior1.1 &  bior1.3 &  bior1.5 &  bior2.2 &  bior2.4 &  bior2.6 &  bior2.8 &  bior3.1 &  bior3.3 &  bior3.5 &  bior3.7 &  bior3.9 &  bior4.4 &  bior5.5 &  bior6.8 &  coif1 &  coif2 &  coif3 &  coif4 &  coif5 &  coif6 &  coif7 \\
\midrule
        Chinatow &     0.69 &     0.69 &     0.68 &     0.78 &     0.73 &     0.75 &     0.66 &     0.74 &     0.69 &     0.77 &     0.65 &     0.68 &     0.69 &     0.62 &     0.70 &   0.66 &   0.68 &   0.71 &   0.80 &   0.73 &   0.68 &   0.71 \\
          Coffee &     0.60 &     0.54 &     0.47 &     0.53 &     0.45 &     0.62 &     0.44 &     0.38 &     0.49 &     0.55 &     0.64 &     0.49 &     0.58 &     0.56 &     0.45 &   0.53 &   0.51 &   0.53 &   0.45 &   0.42 &   0.59 &   0.58 \\
     ECGFiveDays &     0.81 &     0.79 &     0.83 &     0.89 &     0.85 &     0.87 &     0.90 &     0.85 &     0.82 &     0.88 &     0.89 &     0.86 &     0.85 &     0.83 &     0.87 &   0.91 &   0.88 &   0.83 &   0.80 &   0.80 &   0.84 &   0.79 \\
          ECG200 &     0.80 &     0.81 &     0.78 &     0.76 &     0.75 &     0.71 &     0.78 &     0.78 &     0.73 &     0.70 &     0.75 &     0.77 &     0.73 &     0.71 &     0.71 &   0.81 &   0.78 &   0.80 &   0.76 &   0.78 &   0.77 &   0.81 \\
     Handoutline &     0.68 &     0.68 &     0.68 &     0.72 &     0.66 &     0.70 &     0.72 &     0.69 &     0.78 &     0.70 &     0.68 &     0.69 &     0.70 &     0.68 &     0.70 &   0.70 &   0.72 &   0.73 &   0.68 &   0.71 &   0.71 &   0.70 \\
    SonyRobotAI1 &     0.79 &     0.71 &     0.81 &     0.77 &     0.80 &     0.76 &     0.76 &     0.77 &     0.82 &     0.84 &     0.85 &     0.85 &     0.82 &     0.81 &     0.82 &   0.82 &   0.87 &   0.79 &   0.85 &   0.80 &   0.79 &   0.88 \\
    SonyRobotAI2 &     0.86 &     0.78 &     0.89 &     0.80 &     0.80 &     0.85 &     0.89 &     0.80 &     0.82 &     0.81 &     0.84 &     0.87 &     0.84 &     0.87 &     0.89 &   0.83 &   0.85 &   0.84 &   0.85 &   0.87 &   0.85 &   0.88 \\
 StarLightCurves &     0.76 &     0.78 &     0.76 &     0.74 &     0.73 &     0.77 &     0.76 &     0.62 &     0.69 &     0.78 &     0.75 &     0.76 &     0.77 &     0.74 &     0.75 &   0.75 &   0.74 &   0.76 &   0.77 &   0.75 &   0.74 &   0.70 \\
      TwoLeadECG &     0.71 &     0.62 &     0.69 &     0.65 &     0.60 &     0.72 &     0.68 &     0.62 &     0.61 &     0.71 &     0.66 &     0.78 &     0.65 &     0.66 &     0.66 &   0.72 &   0.65 &   0.72 &   0.61 &   0.55 &   0.67 &   0.73 \\
            Yoga &     0.57 &     0.55 &     0.57 &     0.57 &     0.56 &     0.57 &     0.58 &     0.57 &     0.55 &     0.54 &     0.56 &     0.59 &     0.58 &     0.58 &     0.57 &   0.57 &   0.56 &   0.59 &   0.59 &   0.57 &   0.57 &   0.58 \\
   EOGHorizontal &     0.70 &     0.72 &     0.71 &     0.76 &     0.74 &     0.66 &     0.72 &     0.67 &     0.83 &     0.74 &     0.68 &     0.76 &     0.71 &     0.65 &     0.70 &   0.66 &   0.63 &   0.69 &   0.66 &   0.74 &   0.72 &   0.69 \\
     CinECGTorso &     0.51 &     0.46 &     0.46 &     0.32 &     0.42 &     0.28 &     0.49 &     0.32 &     0.48 &     0.43 &     0.39 &     0.35 &     0.43 &     0.38 &     0.39 &   0.53 &   0.50 &   0.39 &   0.50 &   0.67 &   0.44 &   0.49 \\
         ECG5000 &     0.96 &     0.94 &     0.94 &     0.94 &     0.95 &     0.95 &     0.94 &     0.93 &     0.95 &     0.95 &     0.93 &     0.94 &     0.94 &     0.95 &     0.95 &   0.94 &   0.95 &   0.96 &   0.95 &   0.95 &   0.95 &   0.95 \\
\bottomrule
\end{tabular}

}
\end{center}
\begin{center}
\resizebox{15cm}{!} {
\begin{tabular}{lrrrrrrrrrrrrrrrrrrrrrr}
\toprule
        Datasets &  coif8 &  coif9 &  coif10 &  coif11 &  coif12 &  coif13 &  coif14 &  coif15 &  coif16 &  coif17 &   db1 &   db2 &   db3 &   db4 &   db5 &   db6 &   db7 &   db8 &   db9 &  db10 &  db11 &  db12 \\
\midrule
        Chinatow &   0.68 &   0.67 &    0.64 &    0.62 &    0.71 &    0.73 &    0.76 &    0.63 &    0.76 &    0.71 &  0.72 &  0.69 &  0.71 &  0.72 &  0.75 &  0.69 &  0.73 &  0.75 &  0.71 &  0.69 &  0.73 &  0.66 \\
          Coffee &   0.74 &   0.58 &    0.46 &    0.46 &    0.54 &    0.49 &    0.38 &    0.41 &    0.44 &    0.53 &  0.54 &  0.44 &  0.58 &  0.47 &  0.41 &  0.56 &  0.51 &  0.58 &  0.53 &  0.55 &  0.49 &  0.44 \\
     ECGFiveDays &   0.84 &   0.87 &    0.85 &    0.85 &    0.88 &    0.83 &    0.83 &    0.87 &    0.82 &    0.78 &  0.83 &  0.84 &  0.84 &  0.85 &  0.85 &  0.83 &  0.87 &  0.87 &  0.80 &  0.89 &  0.86 &  0.88 \\
          ECG200 &   0.79 &   0.79 &    0.84 &    0.82 &    0.82 &    0.84 &    0.80 &    0.81 &    0.82 &    0.82 &  0.83 &  0.80 &  0.81 &  0.77 &  0.84 &  0.81 &  0.75 &  0.75 &  0.75 &  0.76 &  0.81 &  0.76 \\
    Handoutlines &   0.68 &   0.72 &    0.77 &    0.74 &    0.76 &    0.71 &    0.72 &    0.70 &    0.64 &    0.69 &  0.66 &  0.78 &  0.70 &  0.72 &  0.68 &  0.72 &  0.72 &  0.66 &  0.72 &  0.75 &  0.73 &  0.71 \\
    SonyRobotAI1 &   0.85 &   0.85 &    0.87 &    0.89 &    0.81 &    0.88 &    0.83 &    0.86 &    0.85 &    0.79 &  0.79 &  0.78 &  0.78 &  0.78 &  0.80 &  0.83 &  0.80 &  0.88 &  0.87 &  0.88 &  0.90 &  0.81 \\
    SonyRobotAI2 &   0.84 &   0.86 &    0.85 &    0.86 &    0.86 &    0.84 &    0.86 &    0.82 &    0.86 &    0.84 &  0.88 &  0.79 &  0.86 &  0.79 &  0.85 &  0.84 &  0.88 &  0.84 &  0.85 &  0.78 &  0.89 &  0.86 \\
 StarLightCurves &   0.74 &   0.73 &    0.73 &    0.72 &    0.73 &    0.74 &    0.72 &    0.71 &    0.70 &    0.71 &  0.75 &  0.75 &  0.77 &  0.78 &  0.75 &  0.75 &  0.79 &  0.77 &  0.75 &  0.78 &  0.77 &  0.75 \\
      TwoLeadECG &   0.60 &   0.66 &    0.52 &    0.63 &    0.60 &    0.59 &    0.62 &    0.65 &    0.60 &    0.55 &  0.67 &  0.68 &  0.67 &  0.66 &  0.67 &  0.72 &  0.66 &  0.64 &  0.63 &  0.55 &  0.57 &  0.66 \\
            Yoga &   0.60 &   0.58 &    0.58 &    0.58 &    0.60 &    0.59 &    0.60 &    0.59 &    0.59 &    0.58 &  0.56 &  0.56 &  0.57 &  0.56 &  0.57 &  0.56 &  0.58 &  0.56 &  0.56 &  0.57 &  0.59 &  0.60 \\
   EOGHorizontal &   0.71 &   0.68 &    0.70 &    0.70 &    0.70 &    0.68 &    0.66 &    0.69 &    0.67 &    0.68 &  0.69 &  0.78 &  0.72 &  0.71 &  0.69 &  0.72 &  0.67 &  0.72 &  0.72 &  0.75 &  0.75 &  0.70 \\
     CinECGTorso &   0.52 &   0.55 &    0.46 &    0.34 &    0.54 &    0.45 &    0.55 &    0.46 &    0.48 &    0.46 &  0.47 &  0.43 &  0.50 &  0.62 &  0.39 &  0.49 &  0.44 &  0.40 &  0.46 &  0.40 &  0.46 &  0.39 \\
         ECG5000 &   0.94 &   0.94 &    0.94 &    0.95 &    0.95 &    0.95 &    0.96 &    0.95 &    0.95 &    0.95 &  0.95 &  0.95 &  0.94 &  0.94 &  0.95 &  0.95 &  0.95 &  0.94 &  0.95 &  0.95 &  0.95 &  0.94 \\
\bottomrule
\end{tabular}

}
\end{center}
\begin{center}
\resizebox{15cm}{!} {
\begin{tabular}{lrrrrrrrrrrrrrrrrrrrrrr}
\toprule
        Datasets &  db13 &  db14 &  db15 &  db16 &  db17 &  db18 &  db19 &  db20 &  db21 &  db22 &  db23 &  db24 &  db25 &  db26 &  db27 &  db28 &  db29 &  db30 &  db31 &  db32 &  db33 &  db34 \\
\midrule
        Chinatow &  0.72 &  0.62 &  0.74 &  0.71 &  0.76 &  0.75 &  0.68 &  0.76 &  0.69 &  0.76 &  0.78 &  0.70 &  0.74 &  0.65 &  0.80 &  0.69 &  0.65 &  0.74 &  0.72 &  0.74 &  0.73 &  0.76 \\
          Coffee &  0.51 &  0.42 &  0.38 &  0.42 &  0.47 &  0.49 &  0.54 &  0.46 &  0.37 &  0.46 &  0.40 &  0.58 &  0.62 &  0.45 &  0.53 &  0.33 &  0.40 &  0.53 &  0.55 &  0.42 &  0.49 &  0.41 \\
     ECGFiveDays &  0.83 &  0.86 &  0.83 &  0.89 &  0.80 &  0.87 &  0.81 &  0.77 &  0.81 &  0.79 &  0.82 &  0.79 &  0.84 &  0.85 &  0.86 &  0.85 &  0.72 &  0.84 &  0.86 &  0.80 &  0.81 &  0.79 \\
          ECG200 &  0.81 &  0.75 &  0.80 &  0.79 &  0.76 &  0.80 &  0.81 &  0.83 &  0.86 &  0.81 &  0.78 &  0.78 &  0.82 &  0.77 &  0.81 &  0.81 &  0.82 &  0.77 &  0.83 &  0.80 &  0.81 &  0.78 \\
    Handoutlines &  0.74 &  0.76 &  0.73 &  0.77 &  0.73 &  0.74 &  0.71 &  0.72 &  0.75 &  0.72 &  0.74 &  0.74 &  0.74 &  0.78 &  0.75 &  0.70 &  0.69 &  0.73 &  0.74 &  0.74 &  0.72 &  0.72 \\
    SonyRobotAI1 &  0.82 &  0.83 &  0.79 &  0.82 &  0.84 &  0.84 &  0.87 &  0.85 &  0.84 &  0.82 &  0.84 &  0.84 &  0.84 &  0.85 &  0.83 &  0.86 &  0.87 &  0.83 &  0.81 &  0.79 &  0.82 &  0.82 \\
    SonyRobotAI2 &  0.86 &  0.88 &  0.87 &  0.88 &  0.82 &  0.83 &  0.85 &  0.85 &  0.85 &  0.83 &  0.83 &  0.87 &  0.87 &  0.86 &  0.84 &  0.86 &  0.86 &  0.83 &  0.82 &  0.85 &  0.87 &  0.86 \\
 StarLightCurves &  0.78 &  0.77 &  0.73 &  0.77 &  0.74 &  0.75 &  0.74 &  0.75 &  0.73 &  0.72 &  0.73 &  0.72 &  0.72 &  0.73 &  0.73 &  0.70 &  0.71 &  0.73 &  0.74 &  0.70 &  0.71 &  0.72 \\
      TwoLeadECG &  0.71 &  0.63 &  0.70 &  0.72 &  0.62 &  0.58 &  0.65 &  0.70 &  0.48 &  0.64 &  0.73 &  0.62 &  0.71 &  0.64 &  0.61 &  0.57 &  0.63 &  0.62 &  0.65 &  0.64 &  0.62 &  0.58 \\
            Yoga &  0.59 &  0.57 &  0.59 &  0.59 &  0.58 &  0.59 &  0.59 &  0.58 &  0.59 &  0.59 &  0.59 &  0.59 &  0.61 &  0.58 &  0.59 &  0.59 &  0.58 &  0.58 &  0.57 &  0.60 &  0.58 &  0.59 \\
   EOGHorizontal &  0.73 &  0.66 &  0.75 &  0.75 &  0.73 &  0.70 &  0.77 &  0.69 &  0.72 &  0.74 &  0.71 &  0.72 &  0.67 &  0.74 &  0.73 &  0.68 &  0.75 &  0.71 &  0.73 &  0.68 &  0.72 &  0.72 \\
     CinECGTorso &  0.40 &  0.50 &  0.52 &  0.52 &  0.52 &  0.47 &  0.50 &  0.45 &  0.41 &  0.54 &  0.52 &  0.52 &  0.37 &  0.47 &  0.51 &  0.44 &  0.51 &  0.48 &  0.45 &  0.53 &  0.44 &  0.38 \\
         ECG5000 &  0.95 &  0.95 &  0.95 &  0.94 &  0.94 &  0.95 &  0.95 &  0.94 &  0.95 &  0.95 &  0.95 &  0.95 &  0.95 &  0.95 &  0.94 &  0.94 &  0.94 &  0.94 &  0.95 &  0.95 &  0.95 &  0.95 \\
\bottomrule
\end{tabular}

}
\end{center}
\begin{center}
\resizebox{15cm}{!} {
\begin{tabular}{lrrrrrrrrrrrrrrrrrrrrrr}
\toprule
        Datasets &  db35 &  db36 &  db37 &  db38 &  dmey &  haar &  rbio1.1 &  rbio1.3 &  rbio1.5 &  rbio2.2 &  rbio2.4 &  rbio2.6 &  rbio2.8 &  rbio3.1 &  rbio3.3 &  rbio3.5 &  rbio3.7 &  rbio3.9 &  rbio4.4 &  rbio5.5 &  rbio6.8 &  sym2 \\
\midrule
        Chinatow &  0.71 &  0.69 &  0.77 &  0.74 &  0.74 &  0.71 &     0.71 &     0.69 &     0.72 &     0.78 &     0.74 &     0.72 &     0.81 &     0.75 &     0.73 &     0.74 &     0.61 &     0.72 &     0.79 &     0.67 &     0.81 &  0.66 \\
          Coffee &  0.50 &  0.38 &  0.41 &  0.50 &  0.54 &  0.53 &     0.58 &     0.46 &     0.50 &     0.45 &     0.67 &     0.49 &     0.63 &     0.33 &     0.42 &     0.49 &     0.56 &     0.49 &     0.54 &     0.53 &     0.54 &  0.47 \\
     ECGFiveDays &  0.84 &  0.85 &  0.83 &  0.84 &  0.82 &  0.88 &     0.86 &     0.88 &     0.81 &     0.85 &     0.65 &     0.90 &     0.82 &     0.92 &     0.82 &     0.79 &     0.87 &     0.84 &     0.88 &     0.77 &     0.78 &  0.81 \\
          ECG200 &  0.78 &  0.80 &  0.78 &  0.83 &  0.79 &  0.78 &     0.80 &     0.76 &     0.78 &     0.72 &     0.68 &     0.70 &     0.72 &     0.81 &     0.71 &     0.76 &     0.75 &     0.72 &     0.75 &     0.73 &     0.82 &  0.79 \\
    Handoutlines &  0.72 &  0.72 &  0.70 &  0.74 &  0.75 &  0.70 &     0.65 &     0.70 &     0.71 &     0.75 &     0.69 &     0.71 &     0.73 &     0.74 &     0.71 &     0.73 &     0.73 &     0.67 &     0.72 &     0.70 &     0.68 &  0.75 \\
    SonyRobotAI1 &  0.86 &  0.87 &  0.83 &  0.84 &  0.87 &  0.74 &     0.80 &     0.87 &     0.80 &     0.86 &     0.76 &     0.81 &     0.79 &     0.81 &     0.77 &     0.85 &     0.79 &     0.84 &     0.79 &     0.80 &     0.86 &  0.80 \\
    SonyRobotAI2 &  0.82 &  0.85 &  0.86 &  0.81 &  0.84 &  0.88 &     0.86 &     0.86 &     0.85 &     0.83 &     0.85 &     0.86 &     0.85 &     0.81 &     0.88 &     0.92 &     0.87 &     0.88 &     0.83 &     0.85 &     0.81 &  0.86 \\
 StarLightCurves &  0.75 &  0.70 &  0.73 &  0.73 &  0.73 &  0.77 &     0.76 &     0.79 &     0.77 &     0.77 &     0.75 &     0.74 &     0.78 &     0.72 &     0.76 &     0.73 &     0.74 &     0.75 &     0.77 &     0.76 &     0.73 &  0.74 \\
      TwoLeadECG &  0.69 &  0.58 &  0.62 &  0.67 &  0.57 &  0.70 &     0.66 &     0.68 &     0.71 &     0.63 &     0.63 &     0.67 &     0.66 &     0.66 &     0.61 &     0.73 &     0.71 &     0.62 &     0.62 &     0.64 &     0.66 &  0.63 \\
            Yoga &  0.57 &  0.60 &  0.59 &  0.56 &  0.59 &  0.58 &     0.59 &     0.59 &     0.59 &     0.58 &     0.58 &     0.58 &     0.59 &     0.56 &     0.60 &     0.58 &     0.60 &     0.57 &     0.55 &     0.56 &     0.59 &  0.56 \\
   EOGHorizontal &  0.74 &  0.71 &  0.71 &  0.70 &  0.68 &  0.69 &     0.70 &     0.64 &     0.74 &     0.68 &     0.65 &     0.66 &     0.68 &     0.71 &     0.75 &     0.72 &     0.68 &     0.67 &     0.69 &     0.76 &     0.72 &  0.74 \\
     CinECGTorso &  0.42 &  0.48 &  0.38 &  0.51 &  0.47 &  0.48 &     0.53 &     0.42 &     0.44 &     0.41 &     0.36 &     0.35 &     0.41 &     0.30 &     0.39 &     0.30 &     0.49 &     0.36 &     0.41 &     0.42 &     0.42 &  0.44 \\
         ECG5000 &  0.95 &  0.95 &  0.96 &  0.95 &  0.95 &  0.95 &     0.94 &     0.94 &     0.95 &     0.96 &     0.94 &     0.95 &     0.95 &     0.95 &     0.94 &     0.93 &     0.94 &     0.95 &     0.95 &     0.95 &     0.95 &  0.95 \\
\bottomrule
\end{tabular}

}
\end{center}
\begin{center}
\resizebox{15cm}{!} {
\begin{tabular}{lrrrrrrrrrrrrrrrrrr}
\toprule
        Datasets &  sym3 &  sym4 &  sym5 &  sym6 &  sym7 &  sym8 &  sym9 &  sym10 &  sym11 &  sym12 &  sym13 &  sym14 &  sym15 &  sym16 &  sym17 &  sym18 &  sym19 &  sym20 \\
\midrule
        Chinatow &  0.67 &  0.70 &  0.74 &  0.67 &  0.80 &  0.72 &  0.70 &   0.74 &   0.76 &   0.65 &   0.79 &   0.74 &   0.71 &   0.83 &   0.69 &   0.62 &   0.77 &   0.66 \\
          Coffee &  0.65 &  0.62 &  0.64 &  0.63 &  0.49 &  0.50 &  0.41 &   0.38 &   0.60 &   0.46 &   0.45 &   0.56 &   0.55 &   0.56 &   0.46 &   0.46 &   0.56 &   0.55 \\
     ECGFiveDays &  0.90 &  0.81 &  0.81 &  0.81 &  0.90 &  0.89 &  0.81 &   0.75 &   0.78 &   0.79 &   0.80 &   0.88 &   0.83 &   0.82 &   0.81 &   0.83 &   0.78 &   0.85 \\
          ECG200 &  0.76 &  0.76 &  0.81 &  0.72 &  0.76 &  0.77 &  0.72 &   0.78 &   0.77 &   0.78 &   0.82 &   0.79 &   0.81 &   0.78 &   0.77 &   0.81 &   0.83 &   0.80 \\
    Handoutlines &  0.75 &  0.73 &  0.70 &  0.74 &  0.71 &  0.74 &  0.69 &   0.71 &   0.73 &   0.72 &   0.73 &   0.70 &   0.69 &   0.74 &   0.72 &   0.75 &   0.72 &   0.72 \\
    SonyRobotAI1 &  0.75 &  0.85 &  0.82 &  0.76 &  0.82 &  0.76 &  0.86 &   0.86 &   0.83 &   0.83 &   0.90 &   0.86 &   0.83 &   0.88 &   0.90 &   0.88 &   0.84 &   0.83 \\
    SonyRobotAI2 &  0.87 &  0.86 &  0.84 &  0.85 &  0.86 &  0.86 &  0.84 &   0.85 &   0.81 &   0.87 &   0.87 &   0.91 &   0.88 &   0.85 &   0.85 &   0.87 &   0.85 &   0.87 \\
 StarLightCurves &  0.76 &  0.77 &  0.77 &  0.76 &  0.76 &  0.75 &  0.77 &   0.76 &   0.77 &   0.76 &   0.75 &   0.74 &   0.70 &   0.74 &   0.76 &   0.73 &   0.78 &   0.71 \\
      TwoLeadECG &  0.66 &  0.68 &  0.58 &  0.62 &  0.71 &  0.69 &  0.69 &   0.64 &   0.66 &   0.69 &   0.64 &   0.67 &   0.65 &   0.64 &   0.68 &   0.63 &   0.67 &   0.70 \\
            Yoga &  0.58 &  0.57 &  0.59 &  0.60 &  0.58 &  0.59 &  0.58 &   0.58 &   0.58 &   0.59 &   0.59 &   0.57 &   0.57 &   0.58 &   0.59 &   0.58 &   0.58 &   0.59 \\
   EOGHorizontal &  0.72 &  0.68 &  0.68 &  0.70 &  0.70 &  0.71 &  0.67 &   0.70 &   0.69 &   0.71 &   0.70 &   0.70 &   0.69 &   0.65 &   0.69 &   0.68 &   0.72 &   0.65 \\
     CinECGTorso &  0.45 &  0.44 &  0.51 &  0.41 &  0.49 &  0.56 &  0.39 &   0.48 &   0.49 &   0.37 &   0.46 &   0.48 &   0.43 &   0.58 &   0.36 &   0.41 &   0.48 &   0.40 \\
         ECG5000 &  0.95 &  0.95 &  0.94 &  0.95 &  0.95 &  0.95 &  0.95 &   0.94 &   0.95 &   0.95 &   0.94 &   0.95 &   0.94 &   0.95 &   0.95 &   0.95 &   0.95 &   0.95 \\
\bottomrule
\end{tabular}

}
\end{center}

\clearpage

\section{Multivariate Functional Isolation Forest and depth mapping}

\begin{figure}[h]
\begin{center}
\begin{tabular}{cc}
\hspace*{1cm}\includegraphics[width=.3\textwidth, trim=0cm -1cm 2cm 2cm ]{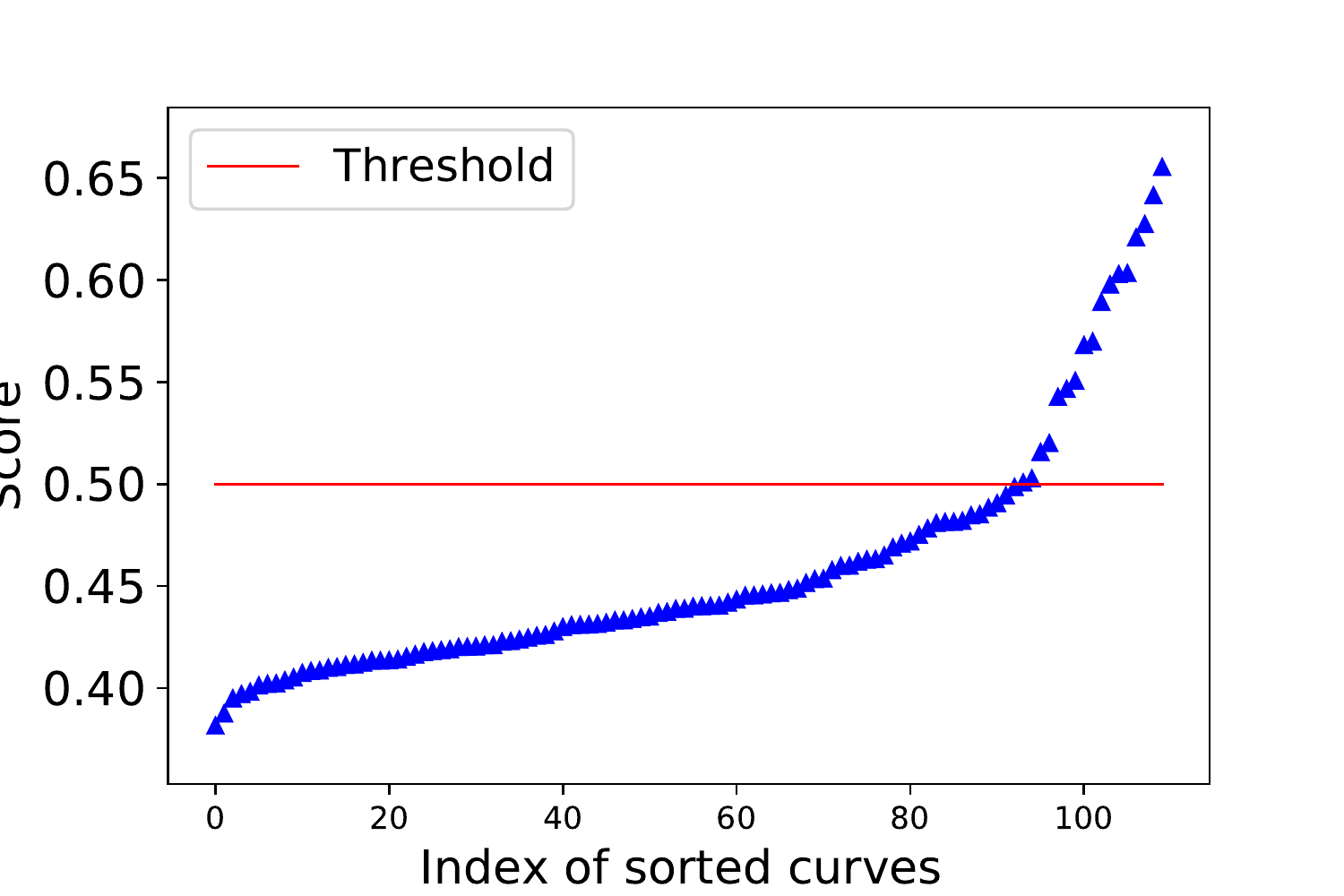} & \includegraphics[width=.6\textwidth, trim=0cm 1.5cm 2cm 2cm,clip=true]{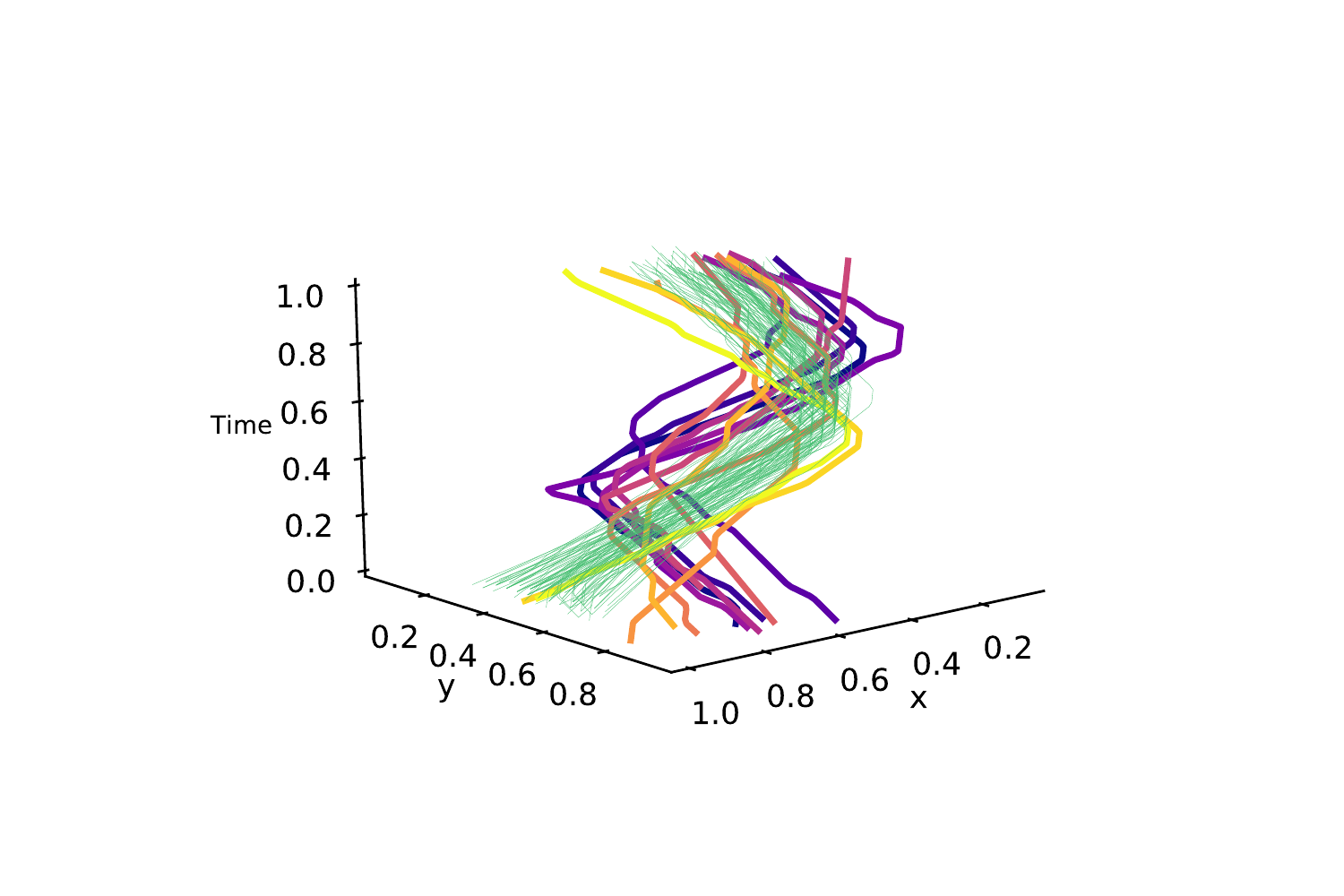} 
\end{tabular}
\vspace*{0.2cm}
\label{mnist}
\end{center}
\begin{center}
{\setlength{\tabcolsep}{0cm}
\begin{tabular}{ccccccccc}
\includegraphics[width=.1\textwidth,]{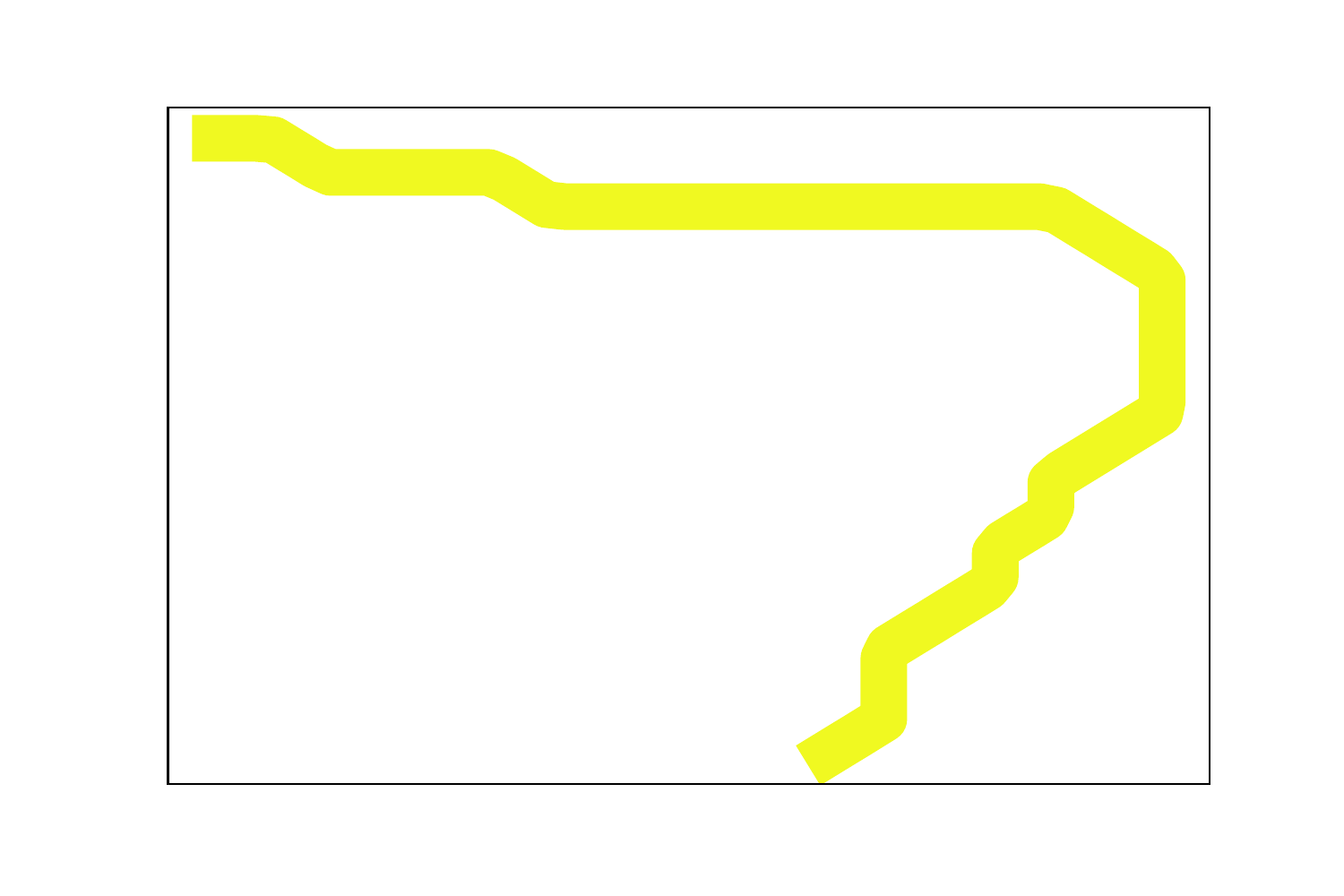} &
 \includegraphics[width=.1\textwidth, ]{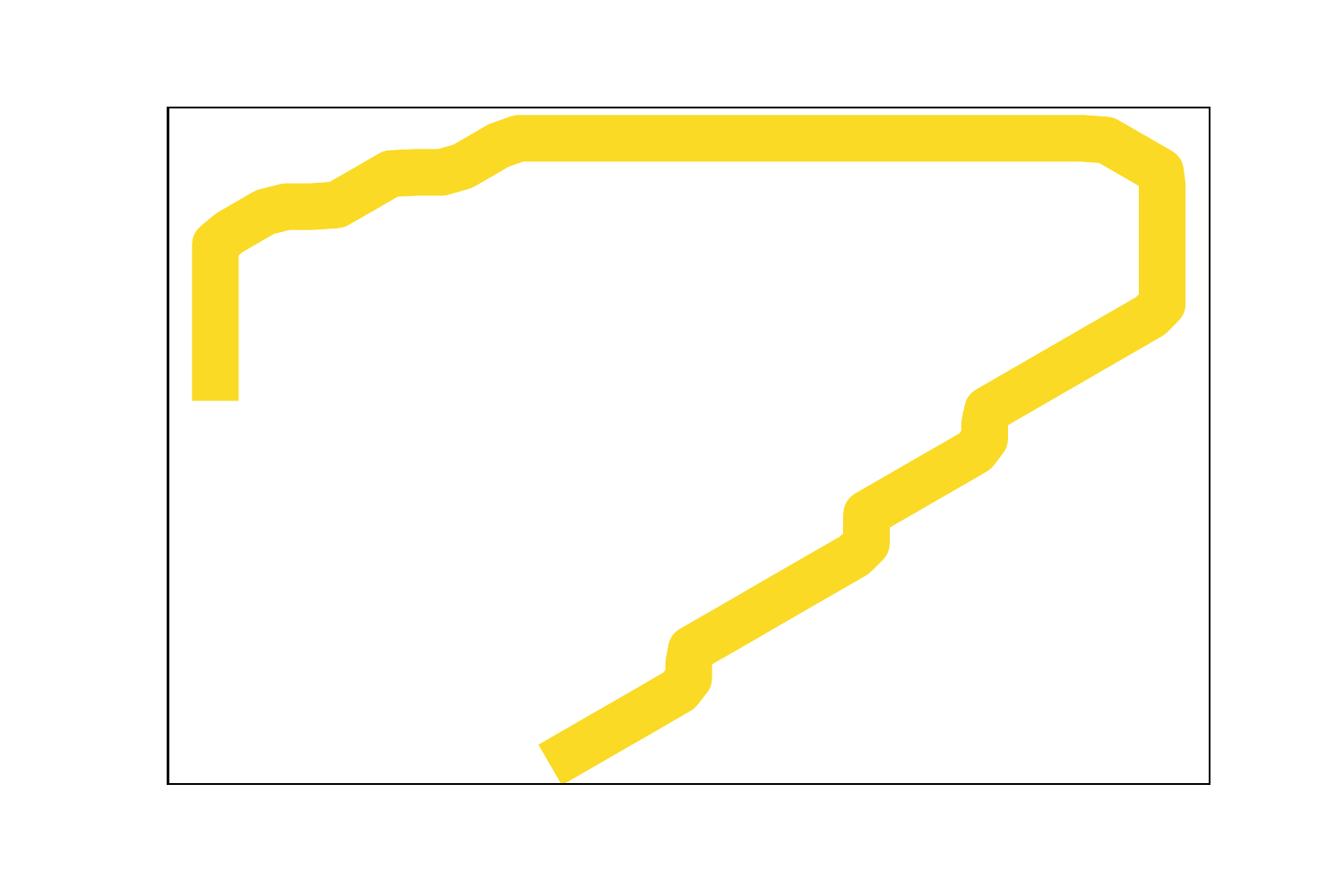} &
\includegraphics[width=.1\textwidth, ]{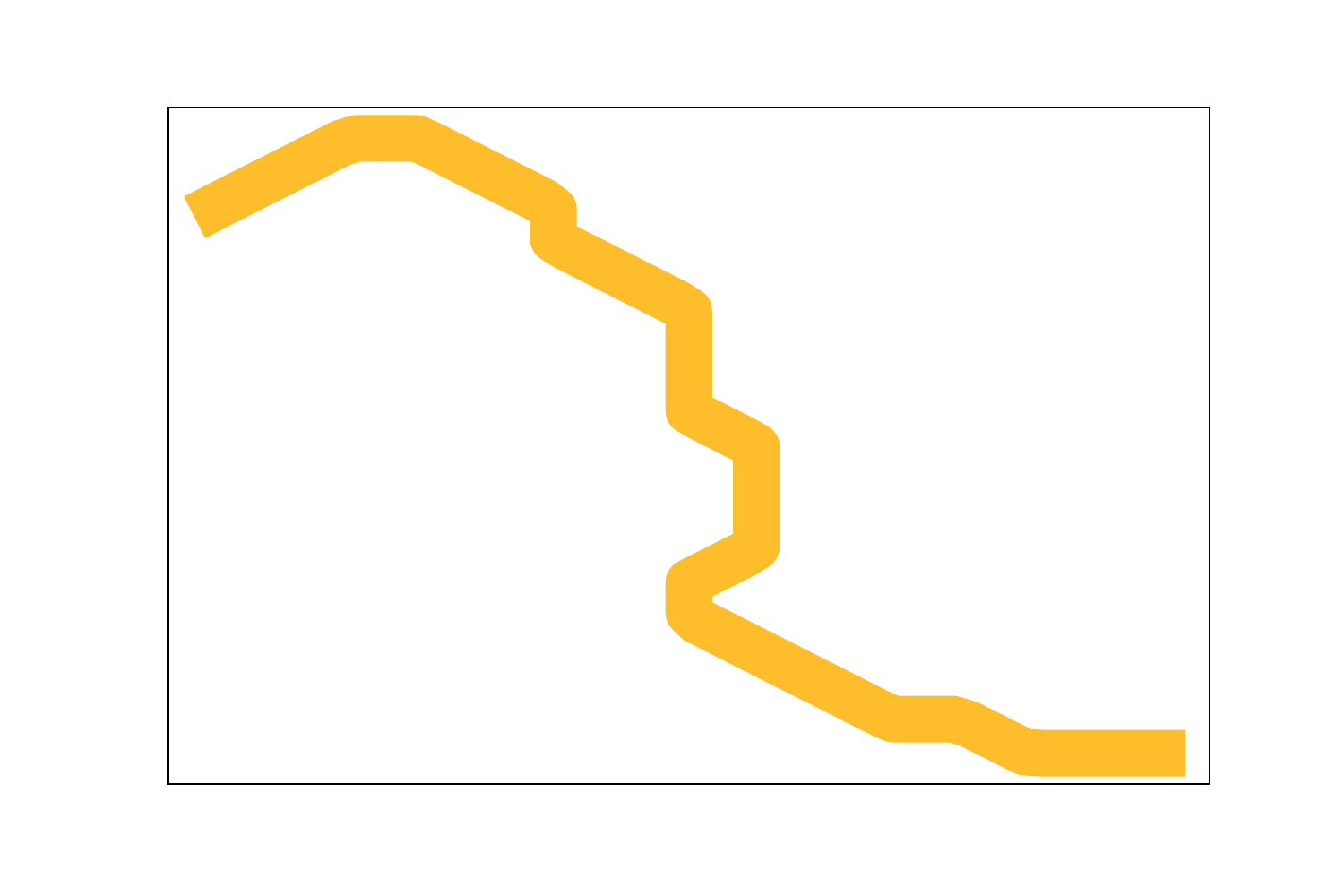} &
\includegraphics[width=.1\textwidth, ]{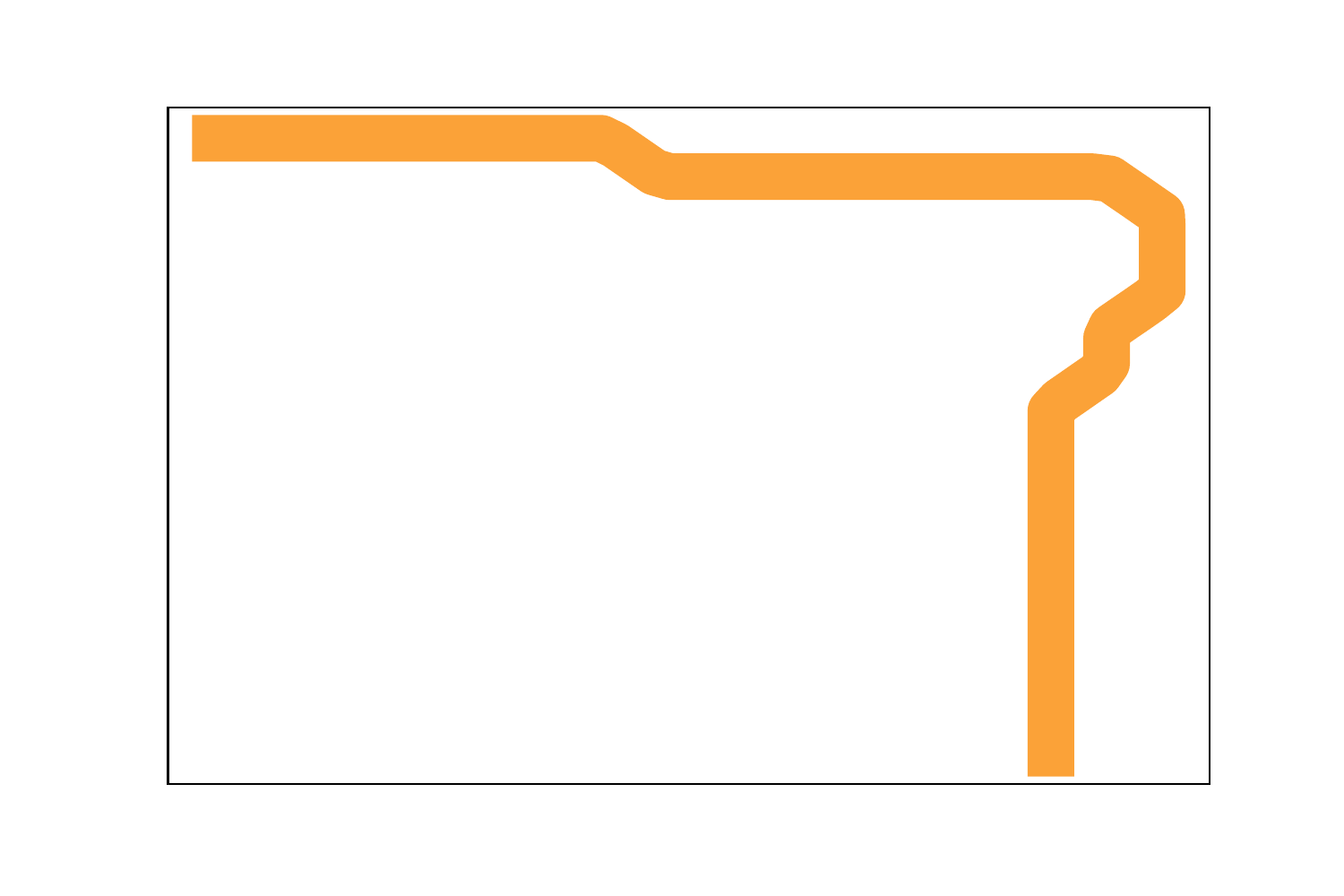} &
\includegraphics[width=.1\textwidth, ]{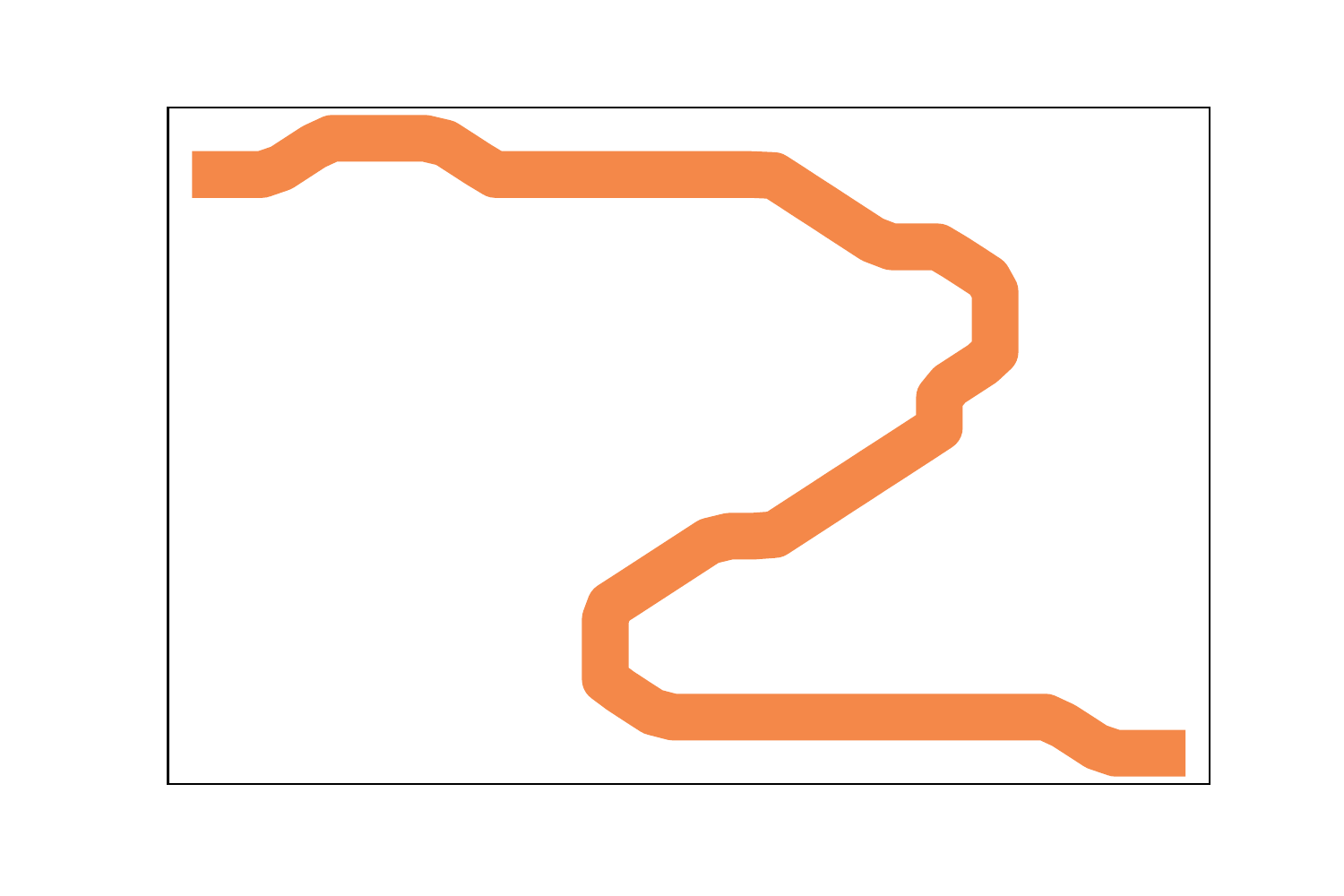} &
\includegraphics[width=.1\textwidth, ]{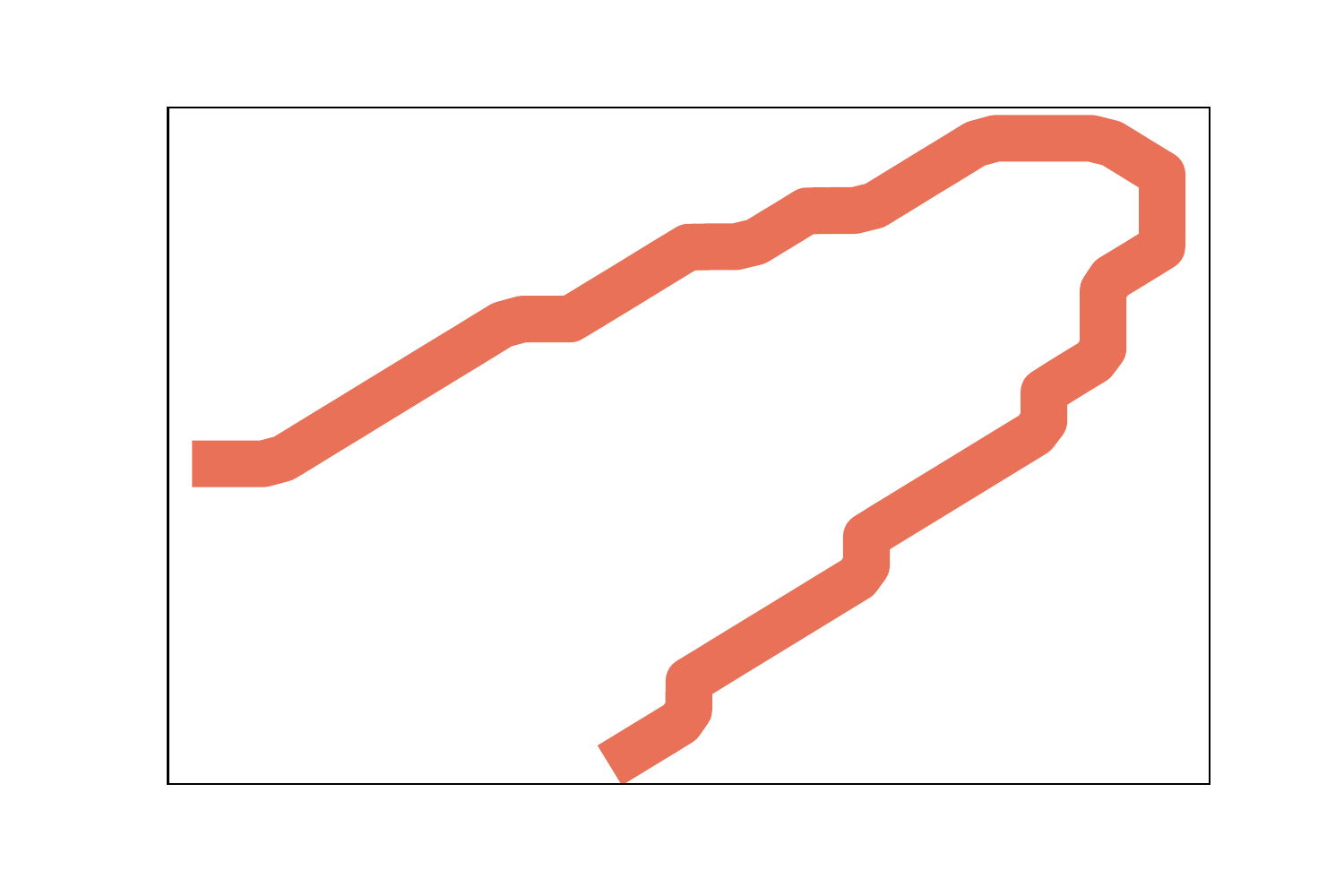} &
\includegraphics[width=.1\textwidth, ]{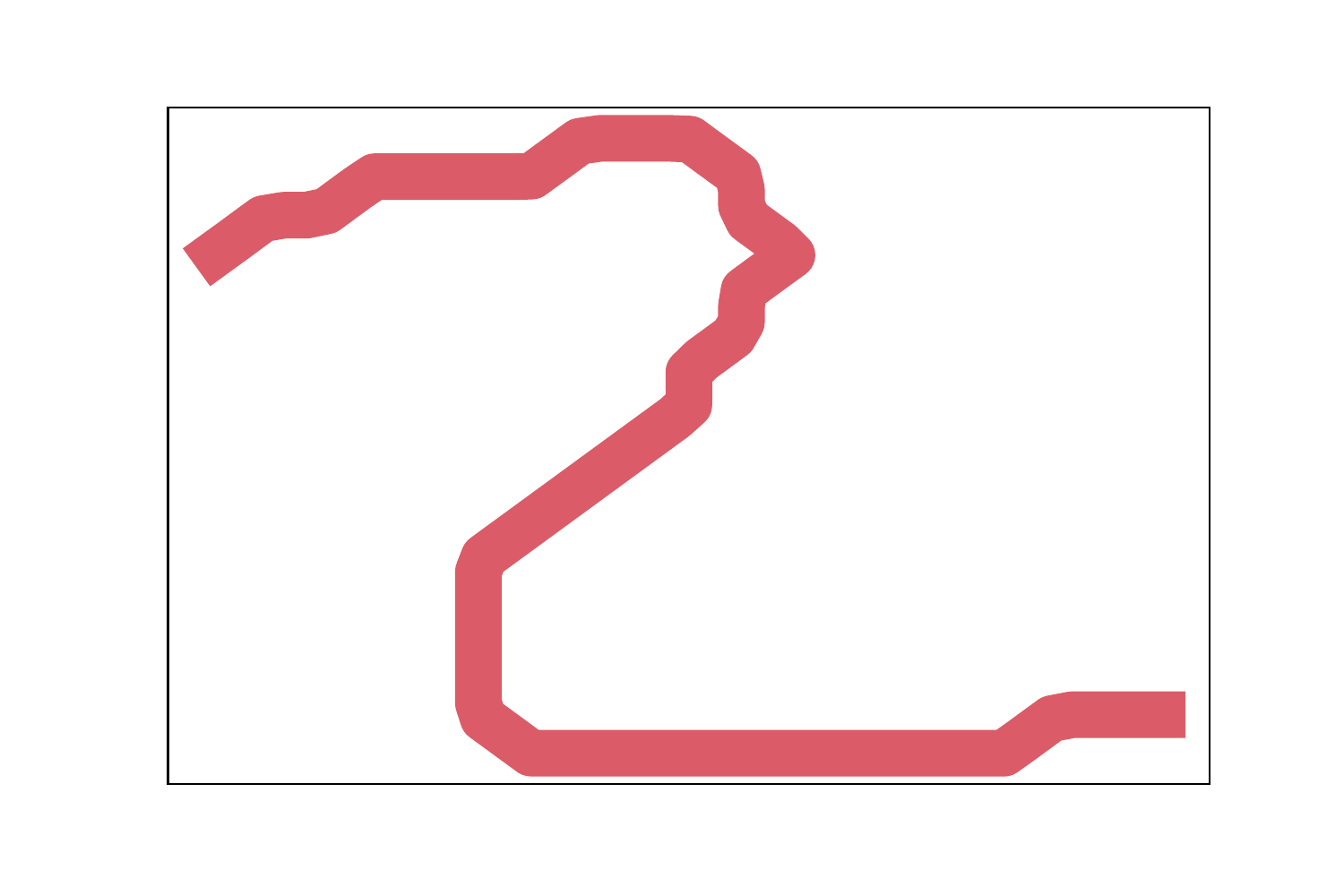} &
\includegraphics[width=.1\textwidth, ]{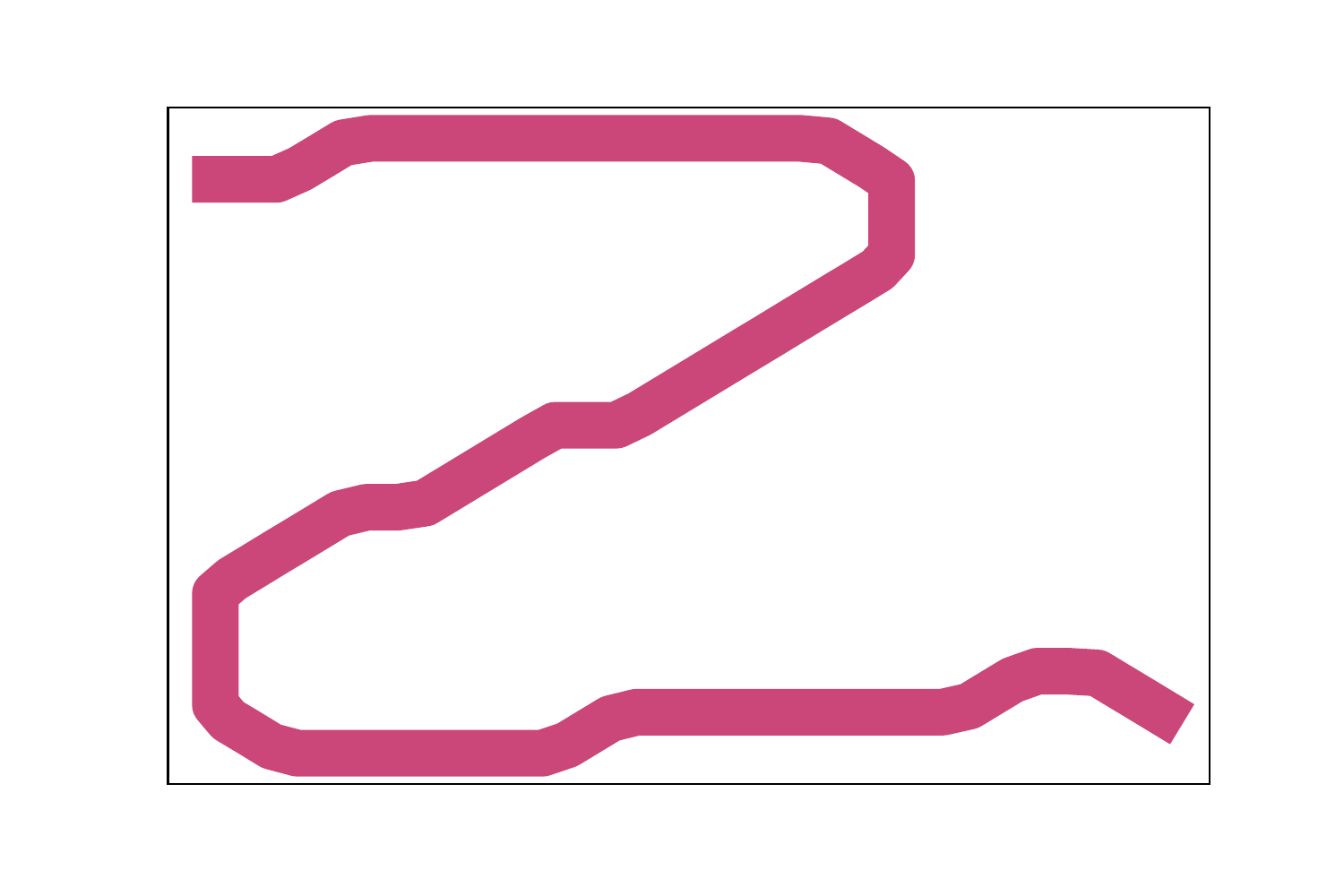} &\\
\includegraphics[width=.1\textwidth, ]{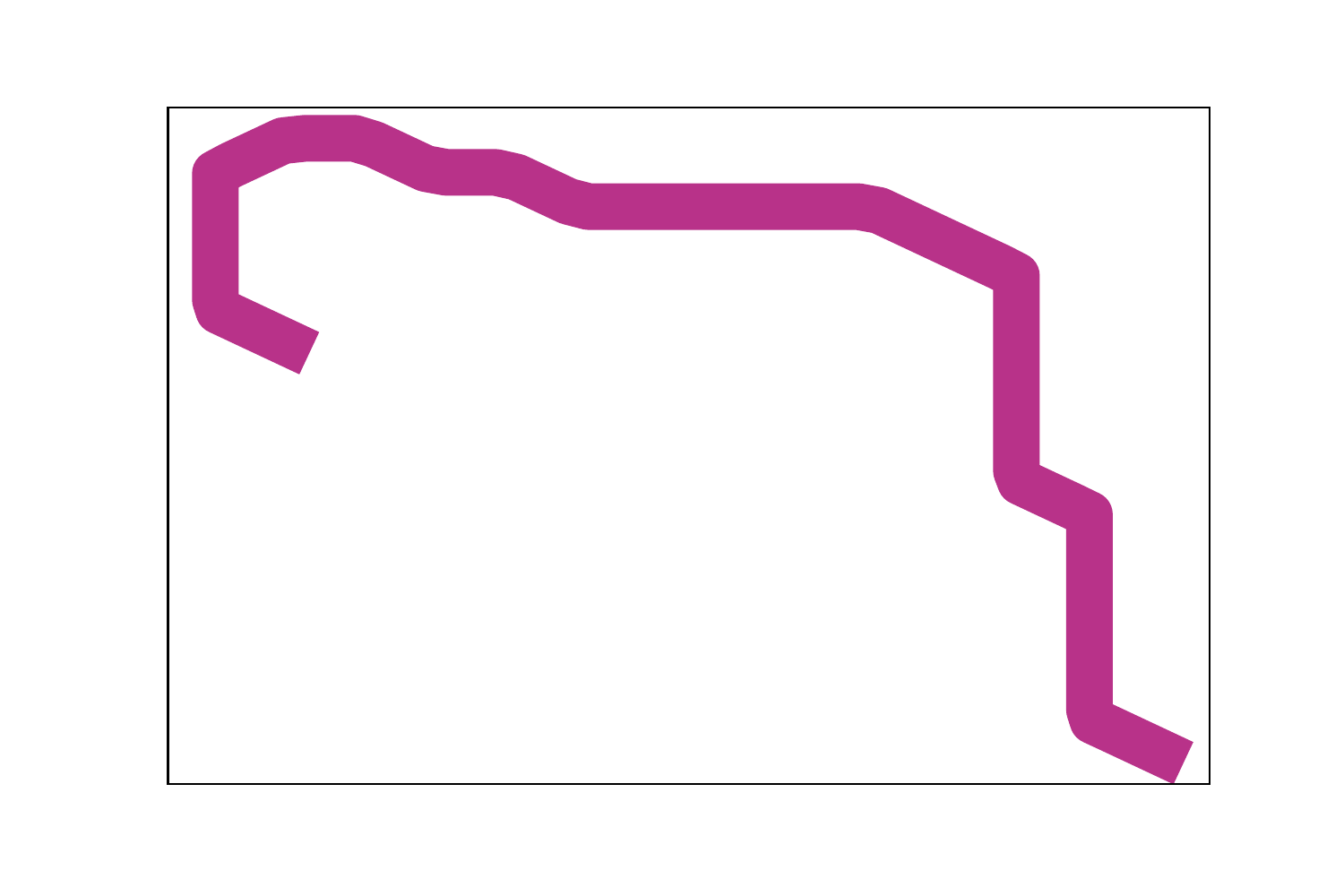} &
\includegraphics[width=.1\textwidth, ]{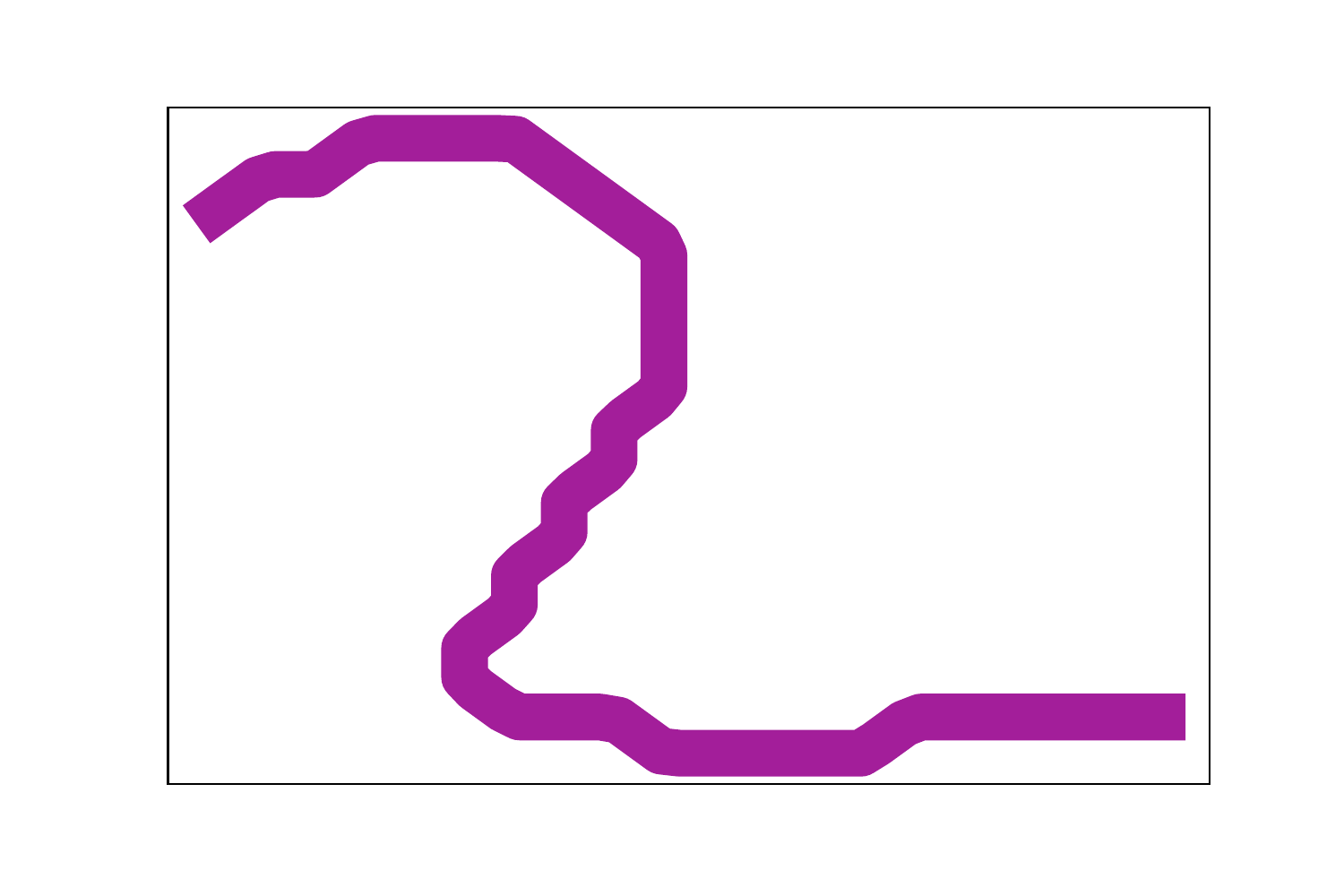} &
\includegraphics[width=.1\textwidth, ]{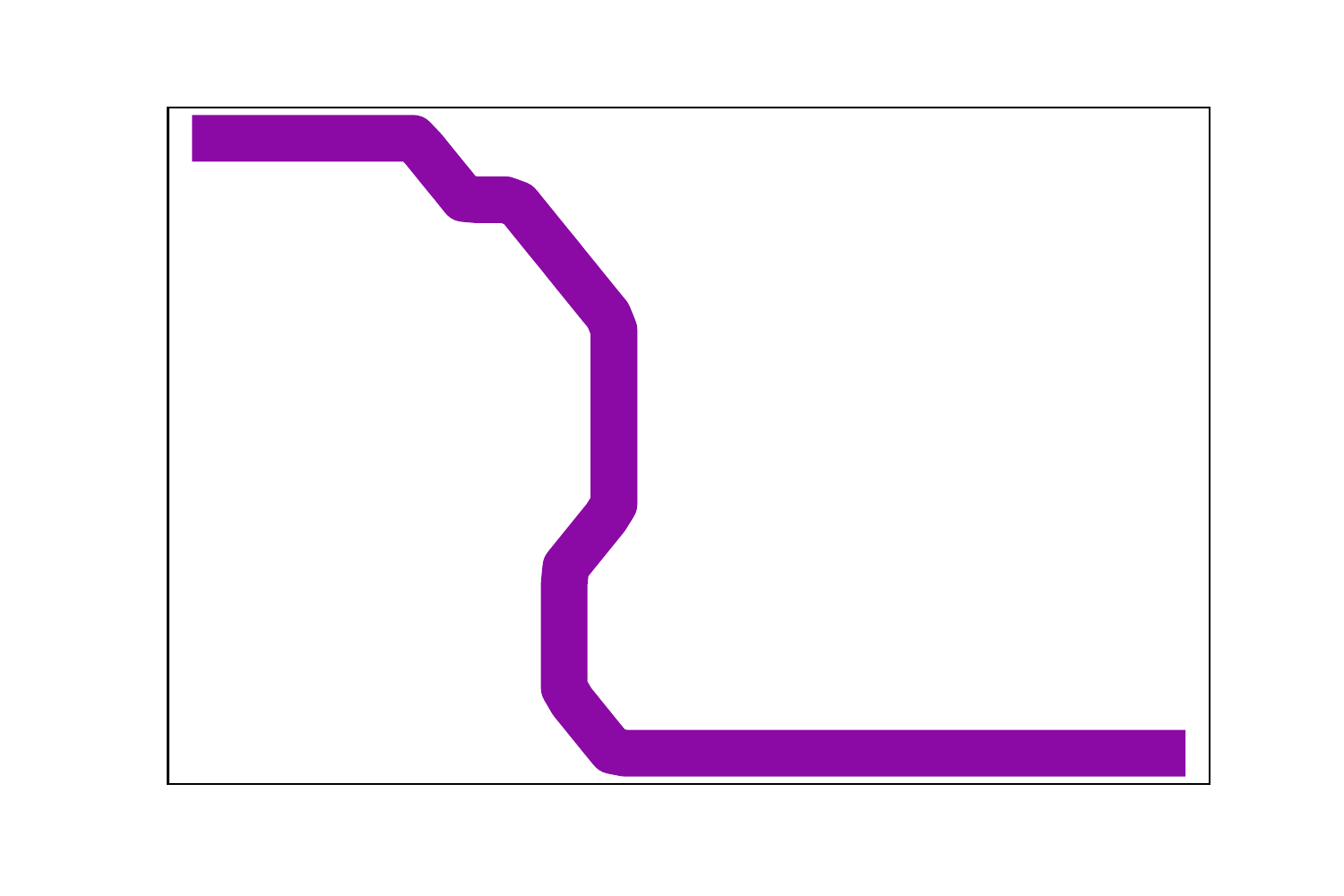} &
\includegraphics[width=.1\textwidth, ]{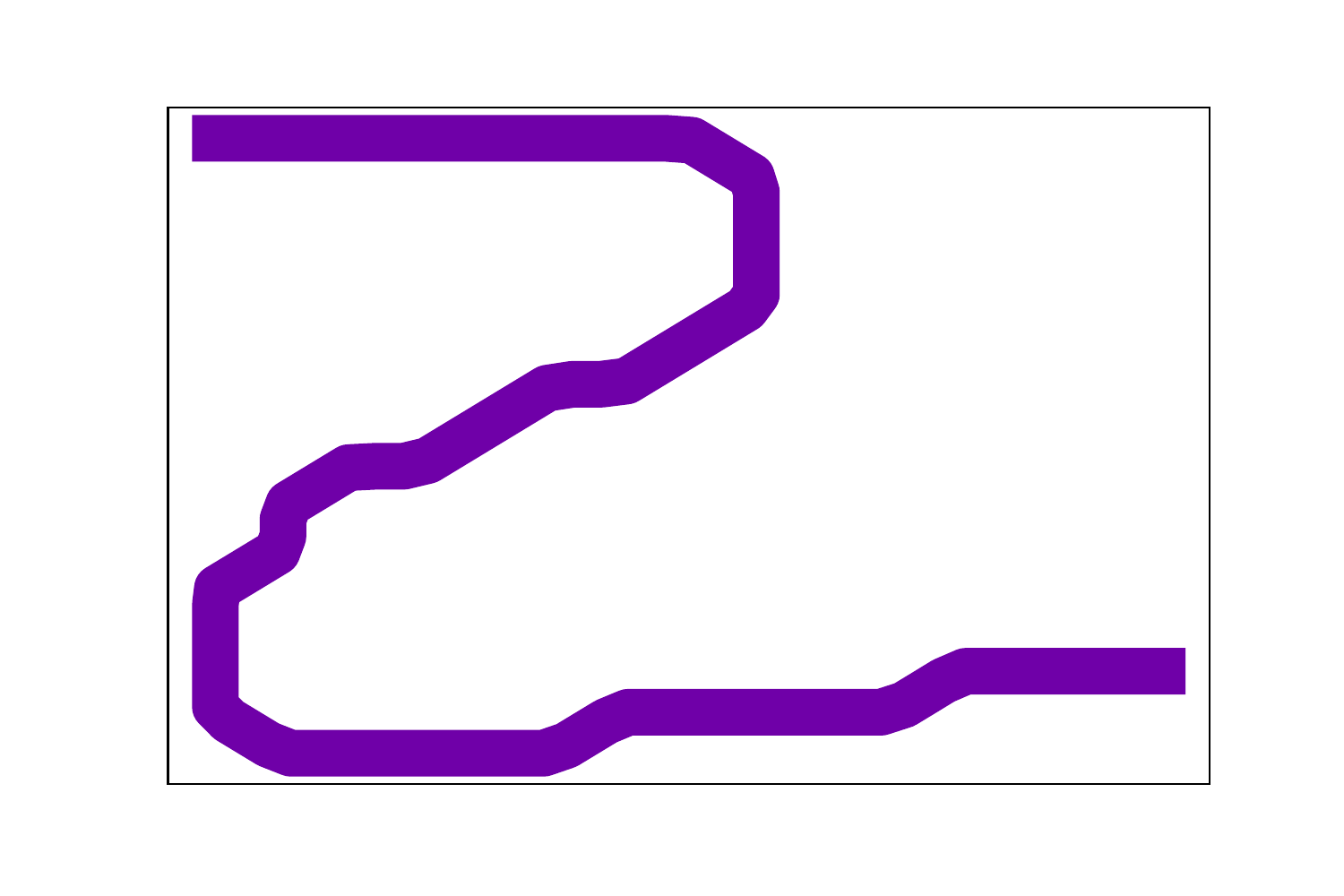} &
\includegraphics[width=.1\textwidth, ]{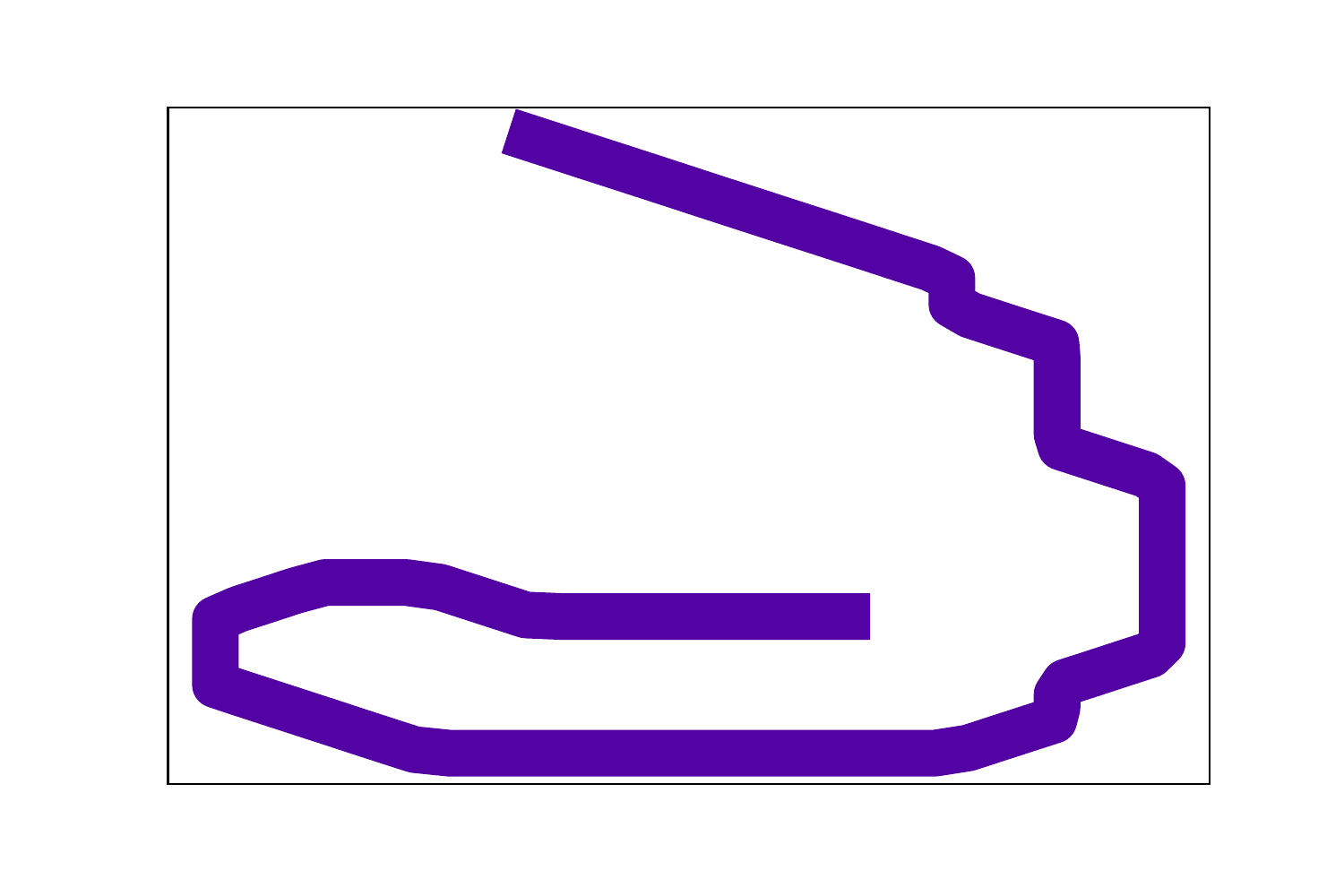} &
\includegraphics[width=.1\textwidth, ]{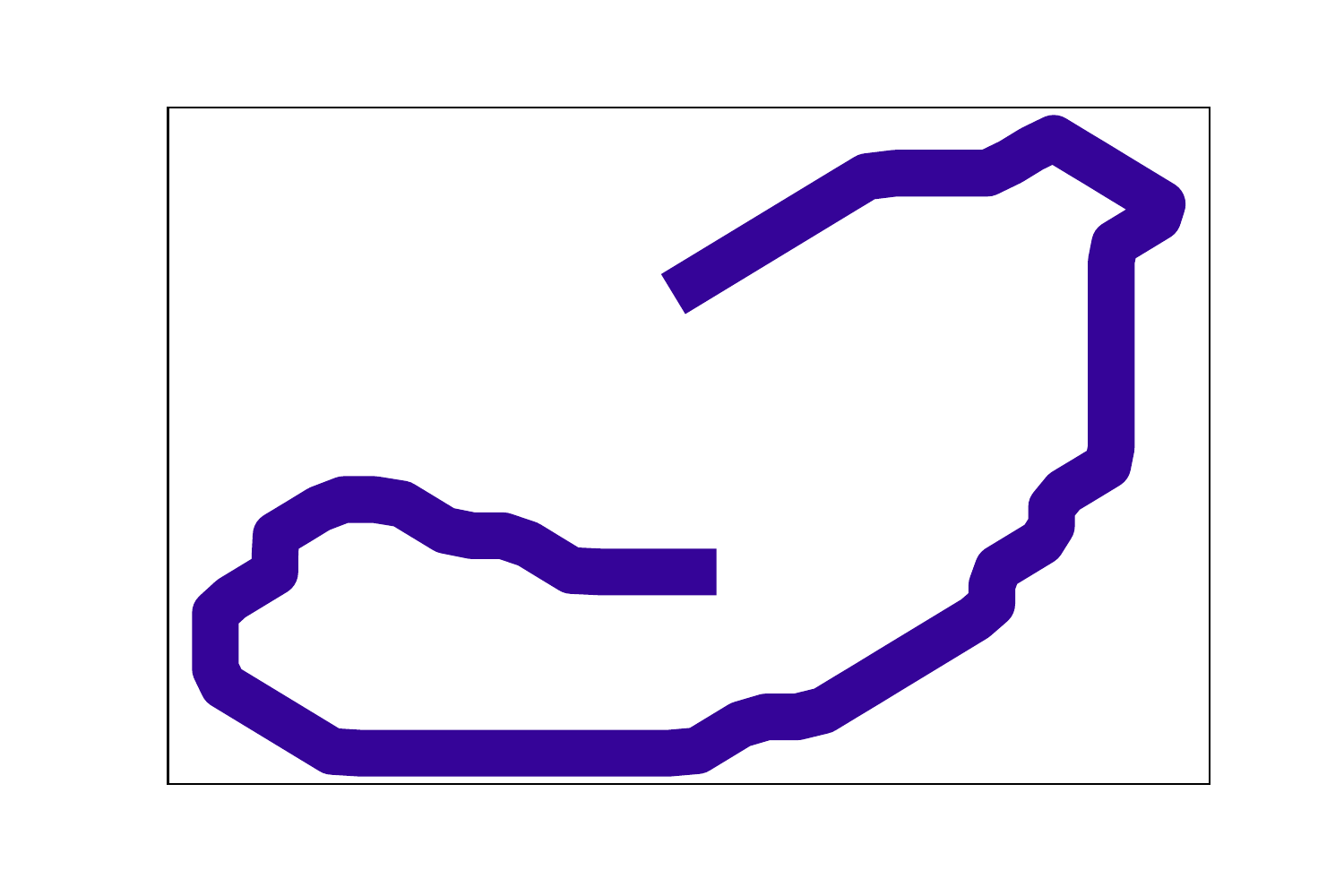} &
\includegraphics[width=.1\textwidth, ]{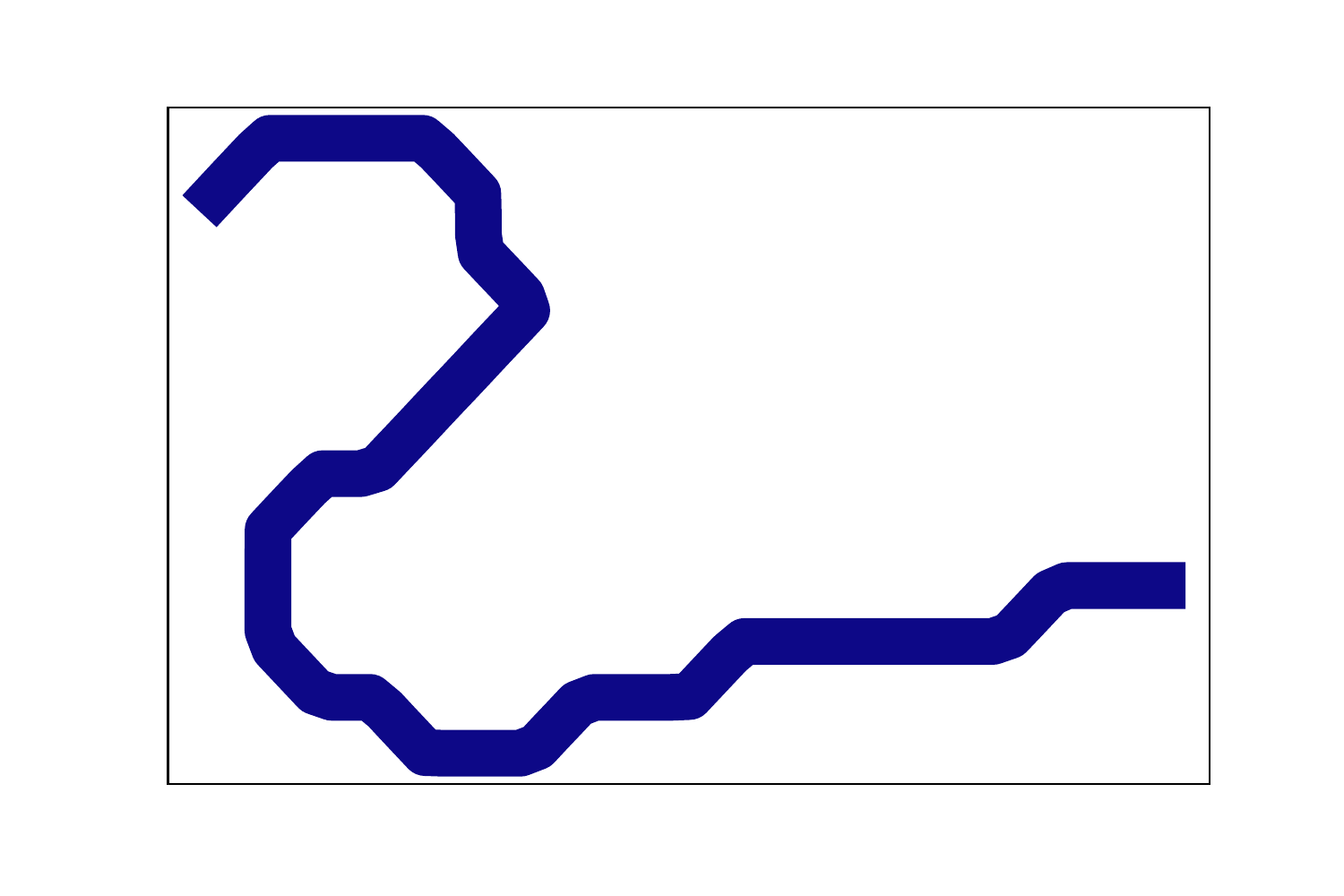} &
\end{tabular}
}
\end{center}
\caption{FIF anomaly scores for a sample of $110$ digits ($100$ seven and $10$ two). Left plot corresponds to the sorted score of these curves. Right plot represents the digits in three dimensions, green ones correspond to normal data, anomalies score increases from orange to dark red. Bottom plot shows the fifteen detected anomalies.}
\label{fig:multMnist}
\end{figure}

FIF can be easily extended to the multivariate functional data, \textit{i.e.} when the quantity of interest lies in $\mathbb{R}^d$ for each moment of time: 
\begin{align*}
X : &\,\Omega \longrightarrow (\mathcal{H}([0,1]))^{\otimes d} \\
&\,\omega \longmapsto \left( (X_1(\omega))_{t\in [0,1]},\; \ldots,\;(X_d(\omega))_{t\in [0,1]}\right)
\end{align*}
For this, the coordinate-wise sum of the $d$ corresponding scalar products is used to project the data onto a chosen dictionary element: $$\langle \mathbf{f},\mathbf{g} \rangle_{\mathcal{H}^{\otimes d}} := \sum_{i=1}^{d} \langle f^{(i)},g^{(i)} \rangle_{\mathcal{H}}.$$
Further, a dictionary should be defined in $(\mathcal{H}([0,1]))^{\otimes d}$. This can be done, \textit{e.g.}, by either componentwise application of one or several univariate dictionaries from Sections~3 and~B, or by constructing of special $d$-variate ones. For illustration purposes, regard the following example constructed based on the MNIST~\cite{Lecun} dataset. First, we extract the digits' contours (skeletons) using skimage python library \cite{skimage}. Then each observation is transformed into a curve in $(L_{2}([0,1])\times L_{2}([0,1]) )$ (one vertical and one horizontal coordinates) using length parametrization on $[0,1]$. We construct the problem by taking $100$ curves from class 7 and adding $10$ observations from class 2. We apply FIF with two-dimensional \textit{sinuscosine} dictionary and the following scalar product : $\langle \mathbf{f},\mathbf{g} \rangle_{(L_{2})^{\otimes d}}$. \textit{sinuscosine} is constructed as a direct extension of \textit{cosine} dictionary introduced for FIF by selecting randomly cosine or sinus function on each coordinates.
Figure~\ref{fig:multMnist} shows anomaly detection using the visual elbow rule to define the threshold. Among those detected, five digits are indeed 7s, but do not resemble them and thus are identified as anomalies.


\end{document}